\documentclass[10pt,twocolumn,letterpaper]{article}

\usepackage{cvpr}              
\usepackage{overpic}
\usepackage{multirow}
\usepackage{makecell}
\usepackage[accsupp]{axessibility}  

\usepackage{tabularx} 
\usepackage{tikz} 
\usepackage{pgfplots} 
\usepackage{pgfplotstable} 
\usepackage{standalone} 
\newlength{\itemwidth} 
\newcolumntype{Y}{>{\centering\arraybackslash}X} 
\newcolumntype{P}[1]{>{\centering\arraybackslash}p{#1}} 

\usetikzlibrary{calc} 
\usetikzlibrary{tikzmark} 
\usetikzlibrary{spy} 
\usetikzlibrary{shapes.misc} 
\usetikzlibrary{fadings} 
\pgfplotsset{compat=newest} 

\definecolor{EE7F0E}{RGB}{238,127,14}
\definecolor{3787CF}{RGB}{55,135,207}
\definecolor{619D47}{RGB}{97,157,71}
\definecolor{DBDC4A}{RGB}{219,220,74}
\definecolor{F1C232}{RGB}{241,194,50}

\pgfplotscreateplotcyclelist{custompalette}{
    {EE7F0E!90!black,fill=EE7F0E},
    {3787CF!90!black,fill=3787CF},
    {619D47!90!black,fill=619D47},
    {DBDC4A!90!black,fill=DBDC4A}
}
\pgfplotscreateplotcyclelist{anothercustompalette}{
    {EE7F0E!90!black,line width=1.5pt},
    {3787CF!90!black,line width=1.5pt},
    {619D47!90!black,line width=1.5pt},
    {DBDC4A!90!black,line width=1.5pt}
}
\pgfplotscreateplotcyclelist{anothercustompalette_rev}{
    {619D47!90!black,line width=1.5pt},
    {3787CF!90!black,line width=1.5pt},
    {EE7F0E!90!black,line width=1.5pt}
}

\newcommand{\usolid}[1]{%
    \tikz[remember picture, baseline=(tosolid.base)]{
        \node[inner sep=0pt, outer sep=0pt] (tosolid) {#1};
    }%
    \tikz[remember picture, overlay]{
        \draw[] ([yshift=-1.5pt]tosolid.south west) -- ([yshift=-1.5pt]tosolid.south east);
    }%
}%

\definecolor{cvprblue}{rgb}{0.21,0.49,0.74}
\usepackage[pagebackref,breaklinks,colorlinks,allcolors=cvprblue]{hyperref}

\usepackage[capitalize]{cleveref}
\crefname{section}{Sec.}{Secs.}
\Crefname{section}{Section}{Sections}
\Crefname{table}{Table}{Tables}
\crefname{table}{Tab.}{Tabs.}

\newcommand{\figref}[1]{Fig.~\ref{#1}}
\newcommand{\tabref}[1]{Tab.~\ref{#1}}

\newcommand{\smallsec}[1]{\vspace{3pt}\noindent\textbf{#1}}

\def\paperID{****} 
\def\confName{CVPR}
\def\confYear{2025}

\title{Classic Video Denoising in a Machine Learning World: \\ Robust, Fast, and Controllable}

\makeatletter
\g@addto@macro\@maketitle{
    \vspace*{-21pt}
    \begin{center}\centering
        \setlength{\tabcolsep}{0.05cm}
        \setlength{\itemwidth}{2.85cm}
        \hspace*{-\tabcolsep}\begin{tabular}{cccccc}
                \begin{tikzpicture}[spy using outlines={3787CF, magnification=10, width={\itemwidth - 0.06cm}, height=2.2cm, connect spies,
        every spy in node/.append style={line width=0.06cm}}]
                    \node [inner sep=0.0cm] {\includegraphics[width=\itemwidth]{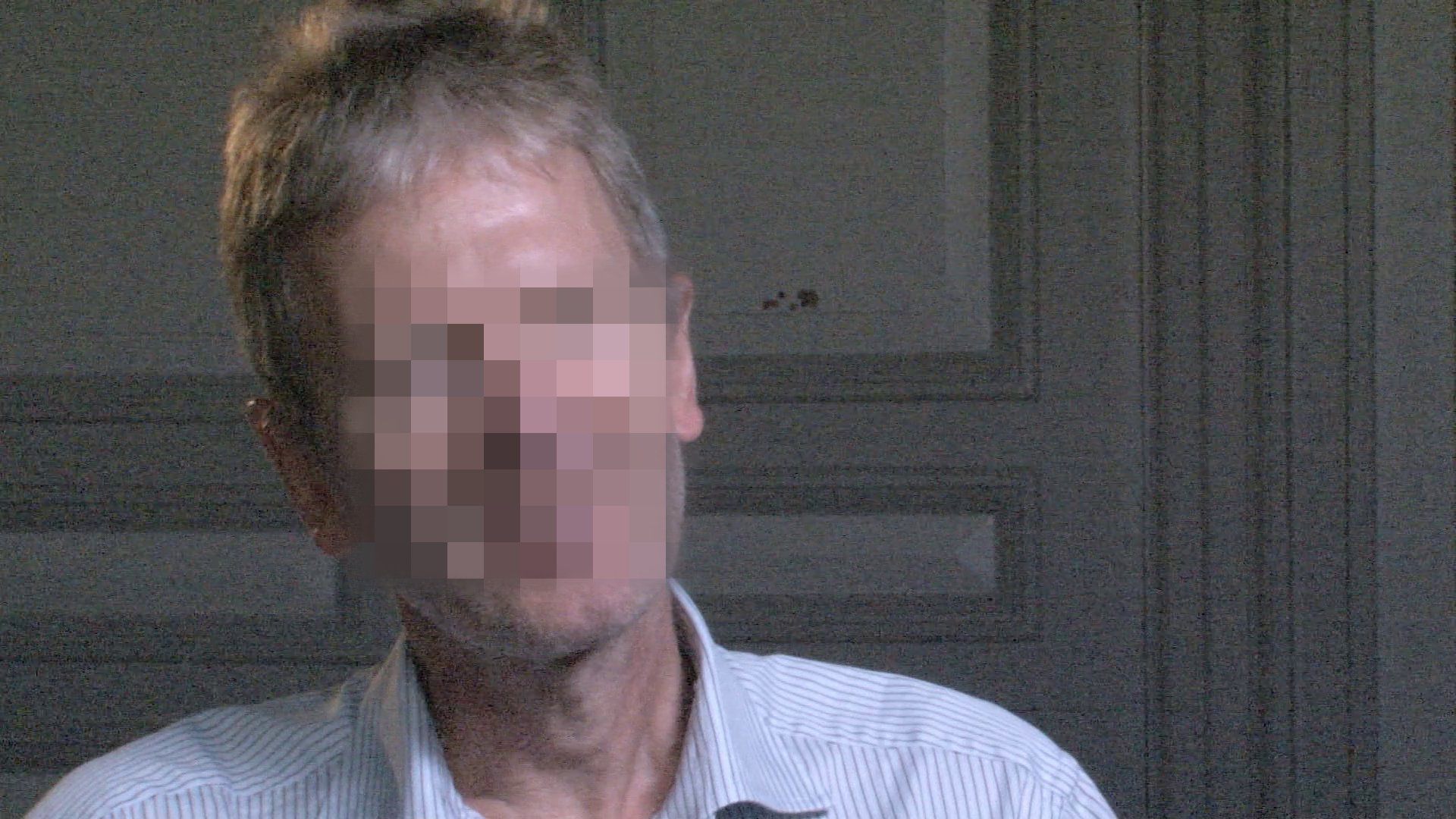}};
                    \spy [every spy on node/.append style={line width=0.06cm}, spy connection path={\draw[line width=0.06cm] (tikzspyonnode) -- (tikzspyinnode);}] on (-0.92,-0.17) in node at (0.0,-2.04);
                \end{tikzpicture}
            &
                \begin{tikzpicture}[spy using outlines={3787CF, magnification=10, width={\itemwidth - 0.06cm}, height=2.2cm, connect spies,
        every spy in node/.append style={line width=0.06cm}}]
                    \node [inner sep=0.0cm] {\includegraphics[width=\itemwidth]{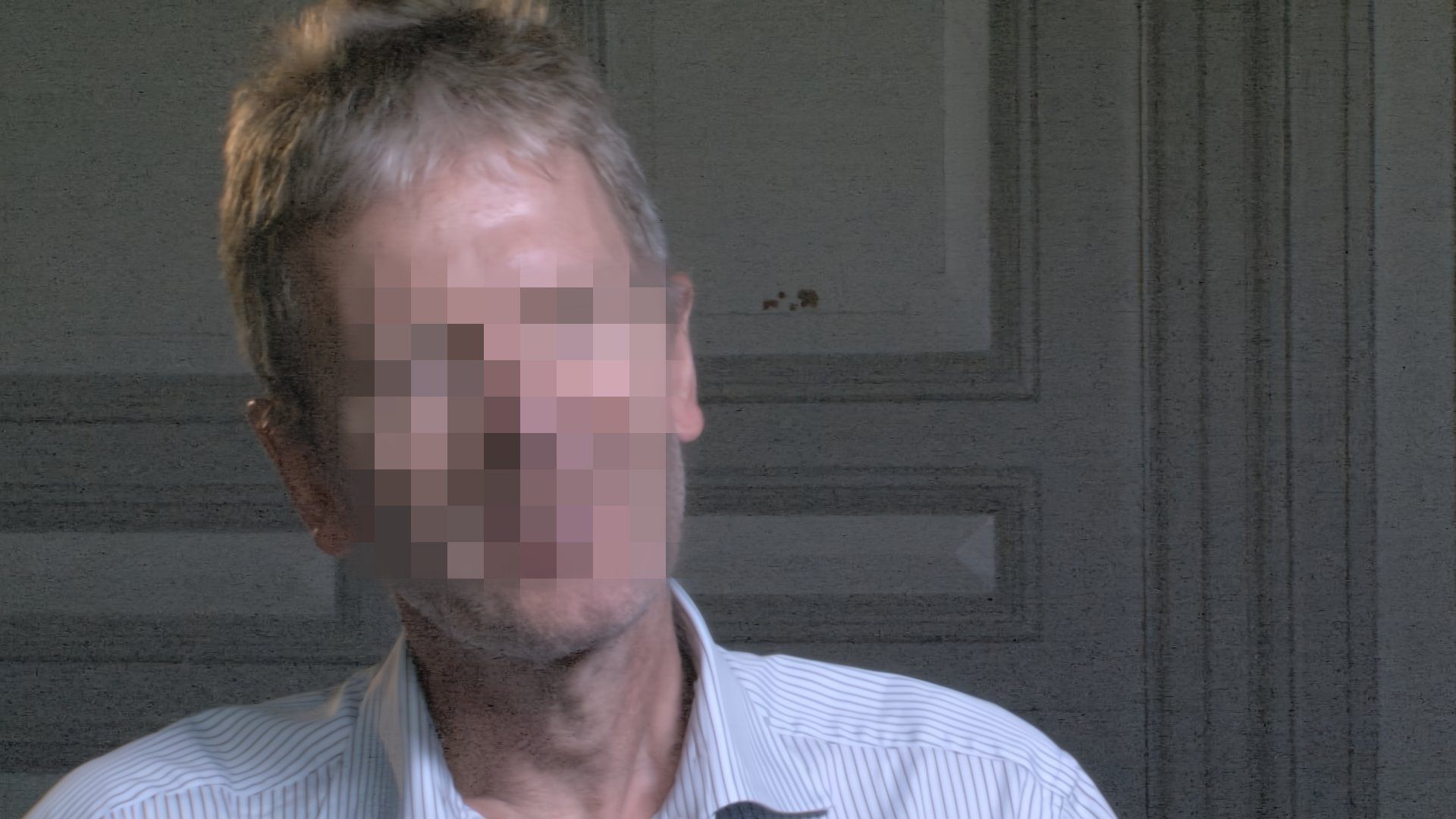}};
                    \spy [every spy on node/.append style={line width=0.06cm}, spy connection path={\draw[line width=0.06cm] (tikzspyonnode) -- (tikzspyinnode);}] on (-0.92,-0.17) in node at (0.0,-2.04);
                \end{tikzpicture}
            &
                \begin{tikzpicture}[spy using outlines={3787CF, magnification=10, width={\itemwidth - 0.06cm}, height=2.2cm, connect spies,
        every spy in node/.append style={line width=0.06cm}}]
                    \node [inner sep=0.0cm] {\includegraphics[width=\itemwidth]{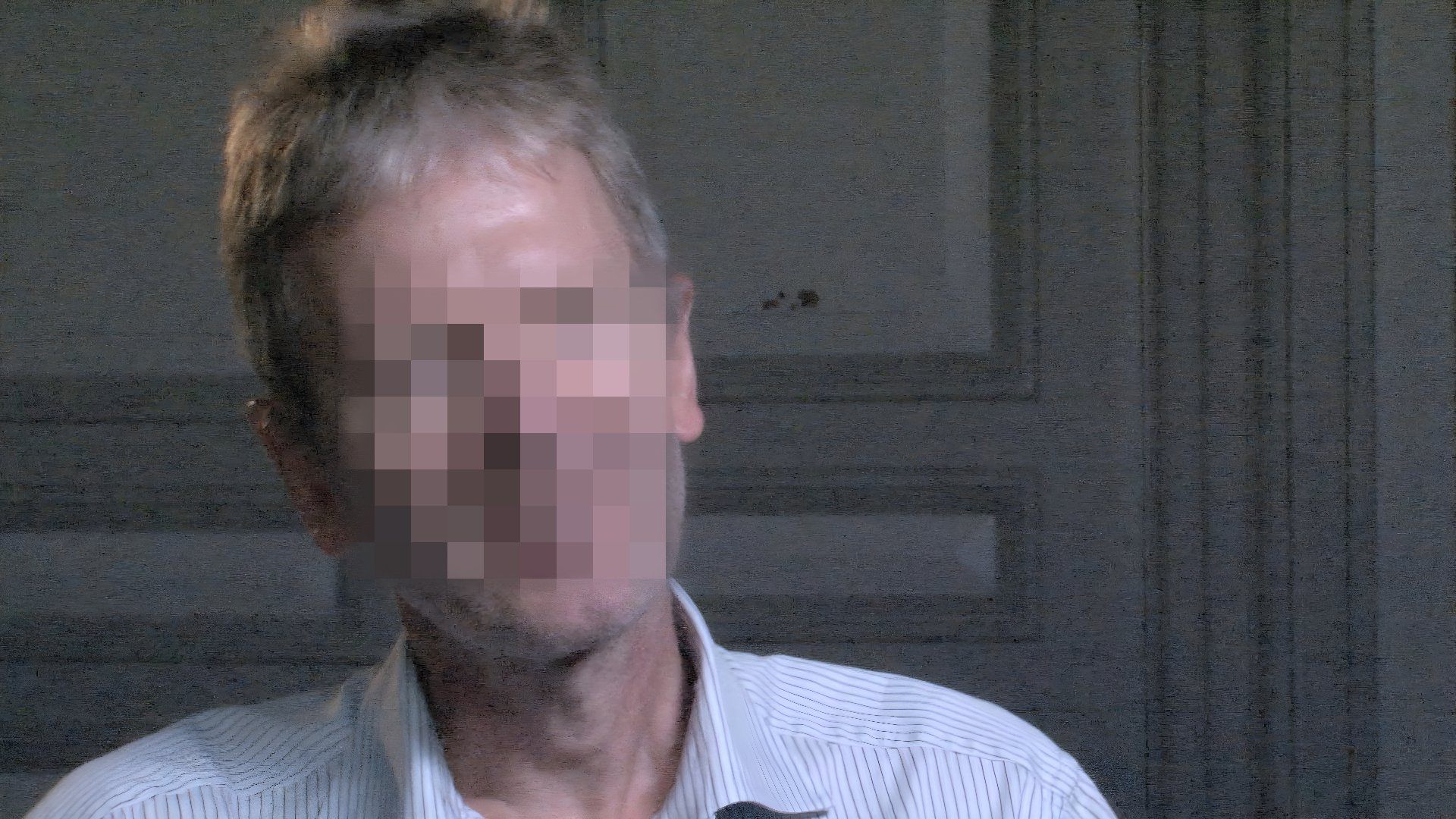}};
                    \spy [every spy on node/.append style={line width=0.06cm}, spy connection path={\draw[line width=0.06cm] (tikzspyonnode) -- (tikzspyinnode);}] on (-0.92,-0.17) in node at (0.0,-2.04);
                \end{tikzpicture}
            &
                \begin{tikzpicture}[spy using outlines={3787CF, magnification=10, width={\itemwidth - 0.06cm}, height=2.2cm, connect spies,
        every spy in node/.append style={line width=0.06cm}}]
                    \node [inner sep=0.0cm] {\includegraphics[width=\itemwidth]{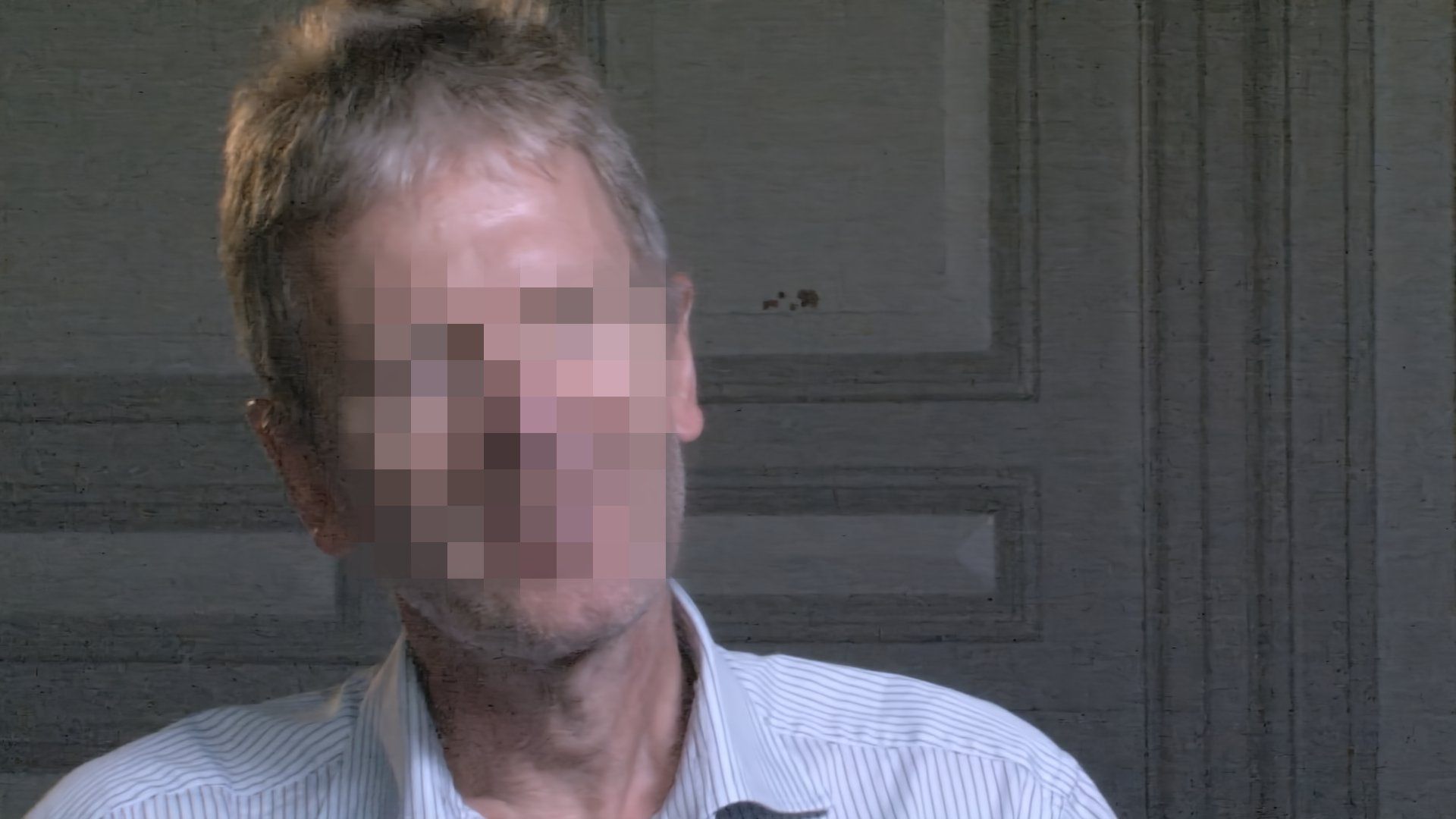}};
                    \spy [every spy on node/.append style={line width=0.06cm}, spy connection path={\draw[line width=0.06cm] (tikzspyonnode) -- (tikzspyinnode);}] on (-0.92,-0.17) in node at (0.0,-2.04);
                \end{tikzpicture}
            &
                \begin{tikzpicture}[spy using outlines={3787CF, magnification=10, width={\itemwidth - 0.06cm}, height=2.2cm, connect spies,
        every spy in node/.append style={line width=0.06cm}}]
                    \node [inner sep=0.0cm] {\includegraphics[width=\itemwidth]{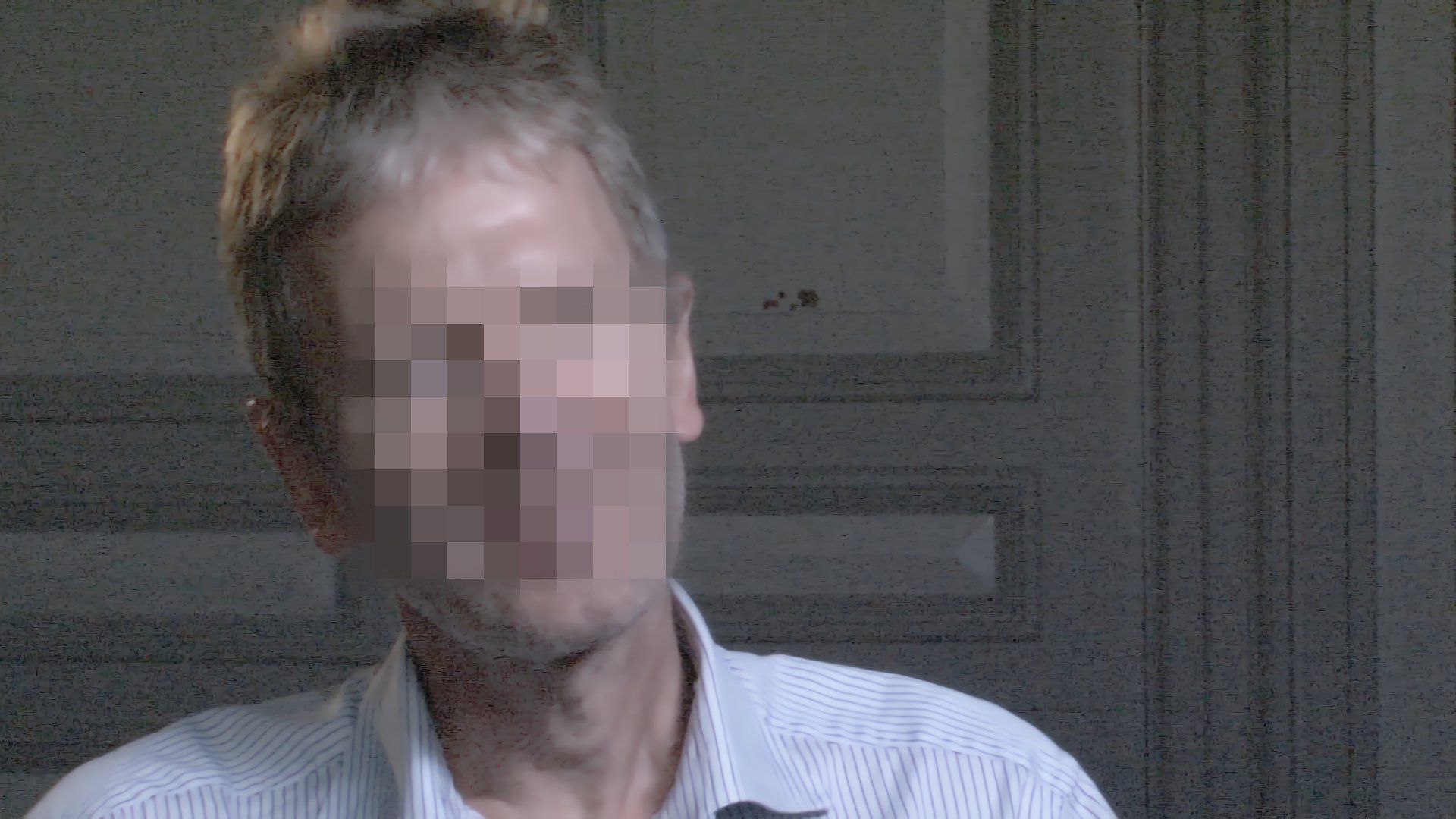}};
                    \spy [every spy on node/.append style={line width=0.06cm}, spy connection path={\draw[line width=0.06cm] (tikzspyonnode) -- (tikzspyinnode);}] on (-0.92,-0.17) in node at (0.0,-2.04);
                \end{tikzpicture}
            &
                \begin{tikzpicture}[spy using outlines={3787CF, magnification=10, width={\itemwidth - 0.06cm}, height=2.2cm, connect spies,
        every spy in node/.append style={line width=0.06cm}}]
                    \node [inner sep=0.0cm] {\includegraphics[width=\itemwidth]{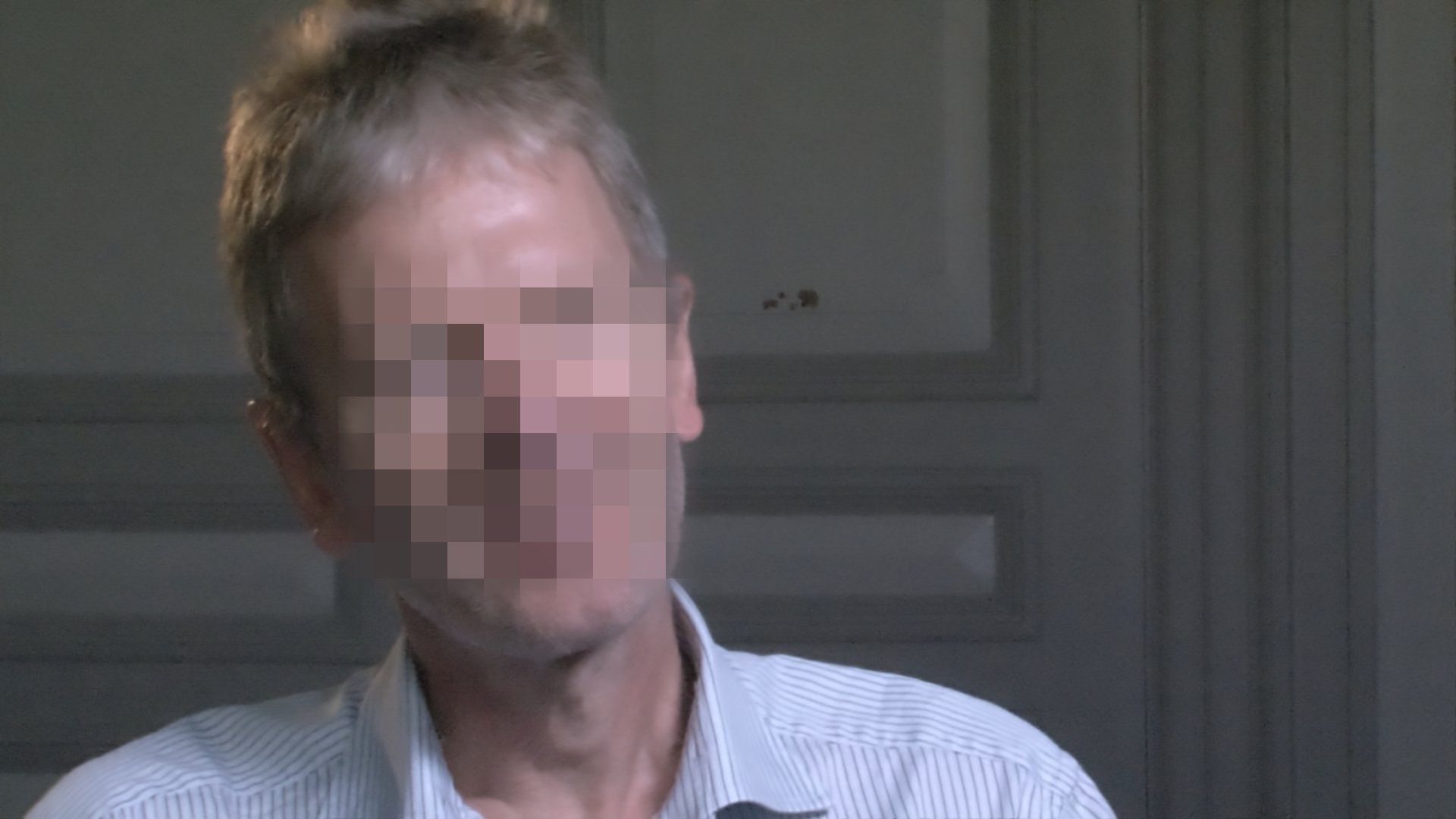}};
                    \spy [every spy on node/.append style={line width=0.06cm}, spy connection path={\draw[line width=0.06cm] (tikzspyonnode) -- (tikzspyinnode);}] on (-0.92,-0.17) in node at (0.0,-2.04);
                \end{tikzpicture}
            \\
                \footnotesize Input
            &
                \footnotesize VRT$^\dag$~\cite{liang2024vrt}
            &
                \footnotesize Real-ESRGAN~\cite{wang2021real}
            &
                \footnotesize MF2F~\cite{dewil2021self}
            &
                \footnotesize UDVD~\cite{sheth2021unsupervised}
            &
                \footnotesize Ours - RFCVD
            \vspace{-0.1cm}\\
                
            &
                \scriptsize (sophisticated model)
            &
                \scriptsize (sophisticated dataset)
            &
                \scriptsize (sophisticated supervision)
            &
                \scriptsize (sophisticated supervision)
            &
                
            \\
        \end{tabular}\vspace{-0.2cm}
    	\captionof{figure}{In real-world videos, noise can manifest in vastly different ways, causing supervised methods to fail when the input is too far outside the training distribution. Footage provided by Robert Kjettrup, face pixelated after inference. We denote retrained models with a $\dagger$.}\vspace{-0.0cm}
    	\label{fig:teaser}
    \end{center}
    \vspace*{4pt}
}
\makeatother

\author{
Xin Jin$^1$\thanks{This project was done during Xin Jin's internship at Adobe Research.}\quad
Simon Niklaus$^2$\thanks{Project lead.\quad$^{\ddagger}$ Corresponding author.}\quad
Zhoutong Zhang$^3$\quad
Zhihao Xia$^3$\quad\\
Chunle Guo$^{1,4\ddagger}$
Yuting Yang$^3$\quad
Jiawen Chen$^3$\quad
Chongyi Li$^{1,4}$\\
{\normalsize{$^1$VCIP, CS, Nankai University}}\quad
{\normalsize{$^2$Adobe Research}}\quad
{\normalsize{$^3$Adobe}}\quad
{\normalsize{$^4$NKIARI, Shenzhen Futian}}
\\
{\tt{\small{xjin@mail.nankai.edu.cn,}}}
{\tt{\small{sniklaus@adobe.com,}}}\\
{\tt{\small{\{guochunle,lichongyi\}@nankai.edu.cn,}}}\\
{\tt{\small{\href{https://srameo.github.io/projects/levd}{https://srameo.github.io/projects/levd}}}\vspace{-3mm}}
}

\begin{document}

\maketitle

\begin{abstract}

Denoising is a crucial step in many video processing pipelines such as in interactive editing, where high quality, speed, and user control are essential. While recent approaches achieve significant improvements in denoising quality by leveraging deep learning, they are prone to unexpected failures due to discrepancies between training data distributions and the wide variety of noise patterns found in real-world videos. These methods also tend to be slow and lack user control. In contrast, traditional denoising methods perform reliably on in-the-wild videos and run relatively quickly on modern hardware. However, they require manually tuning parameters for each input video, which is not only tedious but also requires skill. We bridge the gap between these two paradigms by proposing a differentiable denoising pipeline based on traditional methods. A neural network is then trained to predict the optimal denoising parameters for each specific input, resulting in a robust and efficient approach that also supports user control.

\end{abstract}
    
\section{Introduction}
\label{sec:intro}

Video denoising is a fundamental part of any video editing pipeline, and it is typically applied prior to color grading. After interviewing professional video editors, we have found that they not only want clean results but they also want the denoising to be quick as not to interrupt their workflow, and they want controllability to assert their artistic expression. Specifically, editors often have to decide to leave some noise in the footage in favor of over-smoothing or vice versa, and the answer can differ for each case. For this reason, professional cameras like the Arri Alexa even allow the user to choose between different noise profiles to better support different post-production workflows.

Unfortunately, recent work on video denoising tends to focus only on the first aspect, the denoising quality, which makes these solutions impractical for the typical video editing workflow. At the same time, we have found that such video denoising approaches are subject to failure cases like the one in \figref{fig:teaser} even though they focus on quality. This is not surprising though, considering that noise in real videos can manifest itself in wildly different ways. That is, in addition to the typical degradation that images are subject to, videos also leverage temporal compression. In the H.264 codec, for example, there are P-frames and B-frames that copy information from I-frames which means that the noise is often temporally correlated. Furthermore, there are many different video codecs that affect the noise profile in different ways, and even for the same codec, the noise can vary significantly when using different encoders or settings.

\begin{figure}\centering
    \setlength{\tabcolsep}{0.3cm}
    \hspace*{-\tabcolsep}\begin{tabular}{c|c}
            \includegraphics[height=2.4cm]{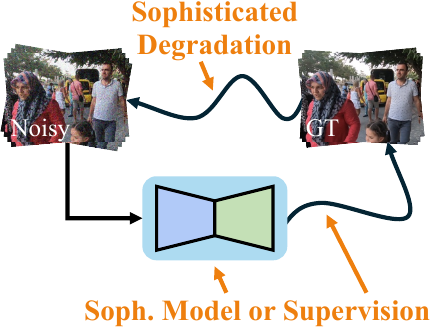}
        &
            \includegraphics[height=2.4cm]{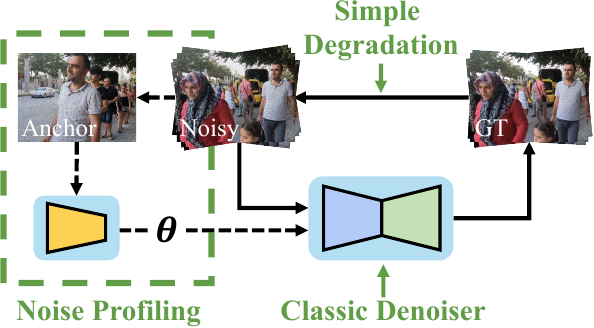}
        \\
            \footnotesize (a) Related Work
        &
            \footnotesize (b) Ours
        \\
    \end{tabular}
    \captionof{figure}{High-level comparison of how related work approaches video denoising (left) and our proposed approach (right).}\vspace{-0.2cm}
    \label{fig:overview}
\end{figure}

To tackle the complexity of video noise in a deep learning context and as summarized in \figref{fig:overview}~(a), we have found that popular approaches either propose a sophisticated way to supervise the model~\cite{dewil2021self, sheth2021unsupervised}, utilize an intricate pipeline to simulate noise~\cite{wang2021real}, leverage an image formation that by design is less prone to overfitting~\cite{davy2019non}, or a combination of these. Yet, the aforementioned example in \figref{fig:teaser} demonstrates that this is not always sufficient. 

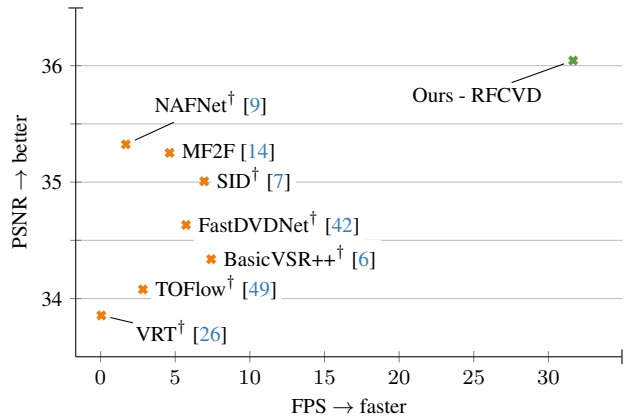
\begin{figure}\centering
    \hspace{-0.15cm}
    \fontsize{8}{10}\selectfont

\begin{tikzpicture}[scale=1.0]

\begin{axis}[
    ymajorgrids=true,
    ticks=both,
    ytick={34,34.5,35,35.5,36},
    yticklabels={34,,35,,36},
    xmin=0.0,
    xmax=33.3,
    ymin=33.5,
    ymax=36.5,
    xlabel={FPS $\rightarrow$ faster},
    ylabel={PSNR $\rightarrow$ better},
    axis x line=bottom,
    axis y line=left,
    x axis line style={-|},
    y axis line style={-|},
    enlarge x limits=0.05,
    enlarge y limits={lower, value=0.0},
    enlarge y limits={upper, value=0.0},
    legend style={at={(0.49, 0.82)}, anchor=south, legend columns=1, /tikz/every even column/.append style={column sep=0.15cm}},
    legend image code/.code={\draw[#1,line width=3pt] (0cm, 0.0cm) -- (0.2cm, 0.0cm);},
    every axis legend/.append style={inner sep=0.1cm},
    legend cell align={left},
    width=8.85cm,
    height=6.23cm,
    cycle list name=anothercustompalette_rev,
    y tick label style={/pgf/number format/.cd, fixed, fixed zerofill, precision=1,/tikz/.cd},
    visualization depends on=\thisrow{angle} \as \angle 
]
    \addplot[
        scatter,
        mark=x,
        mark options={draw=619D47, line width=1.5pt},
        nodes near coords,
        only marks,
        point meta=explicit symbolic,
        nodes near coords=, 
        every node near coord/.append style={anchor=center, pin={[pin edge={thin,black},pin distance=0.3cm]\angle:\pgfplotspointmeta}}, 
        every pin/.append style={fill=white, inner sep=0.04cm},
    ] table [meta=label] {
x y label angle
31.659 36.0437 {Ours - RFCVD} 220
    };

    \addplot[
        scatter,
        mark=x,
        mark options={draw=EE7F0E, line width=1.5pt},
        nodes near coords,
        only marks,
        point meta=explicit symbolic,
        nodes near coords=, 
        every node near coord/.append style={anchor=center, pin={[pin edge={thin,black},pin distance=0.3cm]\angle:\pgfplotspointmeta}}, 
        every pin/.append style={fill=white, inner sep=0.01cm},
    ] table [meta=label] {
x y label angle
1.690 35.3241 {NAFNet$^\dag_{\vphantom{X}}$~\cite{chen2022simple}} 40
0.053 33.8552 {VRT$^\dag_{\vphantom{X}}$~\cite{liang2024vrt}} 357
    };

    \addplot[
        scatter,
        mark=x,
        mark options={draw=EE7F0E, line width=1.5pt},
        nodes near coords,
        only marks,
        point meta=explicit symbolic,
        every node near coord/.append style={fill=white, inner sep=0.06cm, xshift=0.08cm, anchor=west},
    ] table [meta=label] {
x y label angle
4.623 35.2514 {MF2F~\cite{dewil2021self}} 10
6.946 35.0071 {SID$^\dag_{\vphantom{X}}$~\cite{chen2018learning}} 0
5.724 34.6326 {FastDVDNet$^\dag_{\vphantom{X}}$~\cite{tassano2020fastdvdnet}} 0
7.409 34.3381 {BasicVSR++$^\dag_{\vphantom{X}}$~\cite{chan2022basicvsr++}} 0
2.838 34.0784 {TOFlow$^\dag_{\vphantom{X}}$~\cite{xue2019video}} 357
    };
\end{axis}

\end{tikzpicture}\vspace{-0.35cm}
    \captionof{figure}{Video denoising on the CRVD (sRGB) benchmark~\cite{yue2020supervised} using the PSNR across all ISO values with respect to the computational efficiency on an RTX 3090 GPU in FPS (frames per second). We improved some models through retraining, denoted with a $\dagger$.}\vspace{-0.35cm}
    \label{fig:perfplot}
\end{figure}

In contrast, classic denoisers not only work reasonably well but they are also relatively fast, especially on modern hardware. However, they require manually tuning parameters which is not only tedious but also requires a certain amount of skill. We see this as an opportunity. Specifically and as shown in \figref{fig:overview}~(b), by implementing a traditional denoising pipeline in a differentiable manner, we can have a neural network learn to predict what the optimal denoising parameters for a given input should be. As shown in \figref{fig:perfplot}, not only does this end up being robust but it is also comparatively fast while providing support for user control. To be clear on our objectives, if the noise profile is known, such as when developing a denoiser for a specific camera model subject to a constrained video encoder, any reasonable deep learning approach would provide better denoising results. However, such an approach would in turn be limited in terms of robustness and controllability.

In short, our contributions are (1) an approach to leverage a traditional video denoising pipeline in a deep learning setting where we decouple the analysis of the noise from the denoising itself to improve efficiency by avoiding redundant compute, (2) a not only robust and fast but also controllable video denosier (RFCVD), and (3) a training augmentation pipeline based on additive white Gaussian noise together with H.264 transcoding that works surprisingly well.

\begin{figure*}
    \centering
    \begin{overpic}[width=\textwidth]{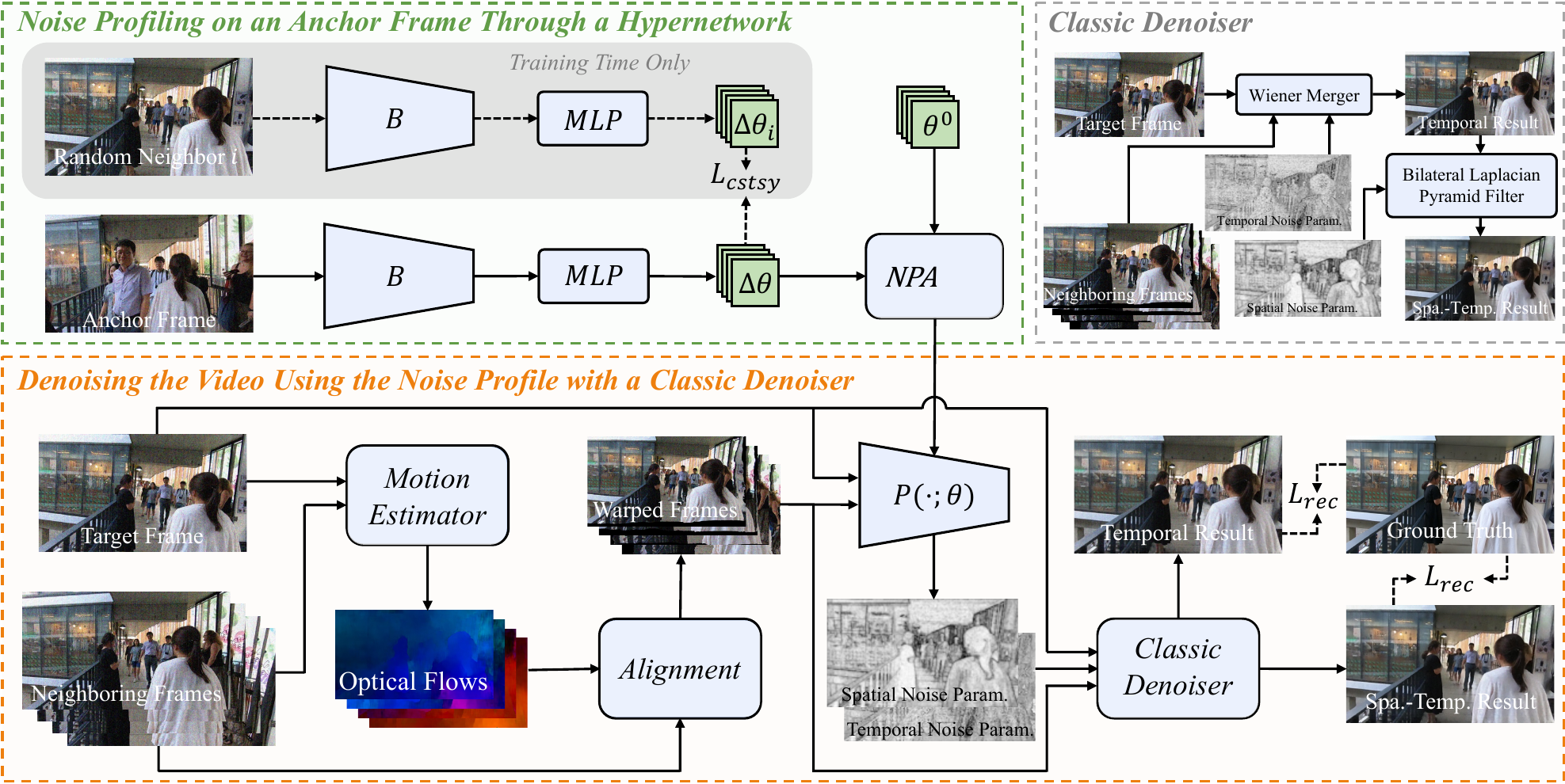}\put(60.3, 31.925){\footnotesize\cite{ortiz2023npa}}\end{overpic}\vspace{-0.25cm}
    \caption{Video denoising fundamentally first needs to analyze the noise and then remove it. We mimic this in our pipeline by first estimating a noise profile on a random anchor frame (top left in green) before using this profile to denoise the video (bottom in yellow). Specifically, we leverage a hypernetwork configuration where the noise profile $\theta$ is essentially the parameters of the subsequent denoiser. That is, our denoiser is a traditional pipeline consisting of (1) a Wiener filter that performs temporal denoising of neighboring frames that were aligned via optical flow and (2) a bilateral Laplacian pyramid filter for spatial denoising of the temporally merged frames, where a small neural network $\mathcal{P}(\cdot;\theta)$ predicts spatially-varying parameters for the Wiener merger and the bilateral filters. This separation of concerns improves the overall efficiency since it avoids having to redundantly analyze the noise over and over again.}\vspace{-0.4cm}
    \label{fig:pipeline}
\end{figure*} 

\section{Related Work}
\label{sec:relwork}

\smallsec{Traditional Video Denoising.} Most traditional video denoising methods can be viewed as a series of linear and non-linear filtering processes, where the filter strength is a function of the estimated noise level. Many such methods are based on block matching~\cite{dabov2006image, dabov2007image, maggioni2012nonlocal, dabov2007video}, where similar patches are first aggregated within and across frames before being filtered and merged to produce the denoised result. There are many variations of this process of course, for example by improving the matching~\cite{liu2010high} or by leveraging bilateral filtering~\cite{tomasi1998bilateral, paris2009bilateral, gavaskar2018fast, chen2007real} for spatial and temporal denoising~\cite{bennett2005video, petreto2019new}. In doing so, image pyramids~\cite{ehmann2018real, burt1987laplacian} can be used to improve the computational efficiency and recent iterations of classic multi-frame denoising methods are even quick to run on commodity smartphones~\cite{hasinoff2016burst, wronski2019handheld, liba2019handheld}.

\smallsec{Machine Learning in Video Denoising.} In recent years, it has become increasingly popular to tackle video denoising with the help of machine learning. This includes methods that, like traditional video denoisers, perform explicit matching~\cite{xue2019video, vaksman2021patch, davy2019non} as well as those where correspondences are only implicit~\cite{maggioni2021efficient, tassano2020fastdvdnet, lee2021restore}. And while most of these methods denoise each frame independently, there are also IIR-like approaches where information is passed along such that previously denoised frames can guide future ones~\cite{chan2021basicvsr, chan2022basicvsr++}. On the model architecture side, the rise of attention in the vision and language community has also led to various attention mechanisms for video restoration~\cite{wang2019edvr, liu2017robust, suin2021gated, cao2021video, liang2024vrt}. But with machine learning there is a catch, if an input has a noise pattern that has not been seen during training then the denoising is unpredictable.

\smallsec{Noise Simulation.} A common approach to facilitate better generalizability on in-the-wild footage is to employ a more sophisticated degradation pipeline when adding synthetic noise to the ground truth. This includes not only doing one but two rounds of degradations~\cite{wang2021real}, shuffling the degradations~\cite{zhang2021designing}, or integrating more advanced operators such as tone mapping~\cite{zhang2023practical}. Although these approaches improve the ability to denoise real-world videos, they can only reduce but not entirely resolve the generalizability gap.

\smallsec{Self-Supervised Learning.} Another strategy to improve the generalizability in video denoising is to adopt self-supervision on real-world videos without requiring access to a ground truth. For example, this can be achieved by leveraging the nature of regression losses~\cite{lehtinen2018noise2noise} or by building on the blind spot idea~\cite{laine2019blindspot, sheth2021unsupervised, ehret2019model, dewil2021self}. While these approaches make it possible to train on noisy in-the-wild videos, it is not necessarily feasible to collect a dataset that contains all possible types of noise that occur in the real world. Therefore, self-supervision often also relies on test-time adaptation which comes with its own challenges.

\section{Method}
\label{sec:method}

Video denoising fundamentally first needs to analyze the noise and then remove it. In contrast, recent approaches perform both tasks at once by utilizing a single neural network that is given the noisy input frames and is tasked with producing the clean output. The noise profile in a video does typically not change over time though, yet these approaches independently denoise one frame after another so they have to analyze the noise over and over again.

To avoid this redundancy, we follow the natural separation of analyzing the noise and then removing it. As shown in \figref{fig:pipeline}, we first estimate the noise profile on an anchor frame and then denoise the video using the estimated profile. Specifically, we leverage a hypernetwork setup where the noise profile $\theta$ provides the parameters of the subsequent denoiser. That is, our denoiser is a traditional pipeline consisting of (1) a Wiener filter that performs temporal denoising of neighboring frames that were aligned via optical flow and (2) a bilateral Laplacian pyramid filter for spatial denoising of the temporally merged frames, where a small neural network $\mathcal{P}(\cdot;\theta)$ predicts spatially-varying parameters for the Wiener merger and the bilateral filters.

We will subsequently discuss these parts in turn before providing more details on user control and the training.

\begin{figure*}
    \centering
    \setlength{\tabcolsep}{0.05cm}
    \setlength{\itemwidth}{2.85cm}
    \renewcommand{\arraystretch}{0.9}
    \hspace*{-\tabcolsep}\begin{tabular}{cccccc}
            \begin{tikzpicture}[spy using outlines={3787CF, magnification=18, width={\itemwidth - 0.06cm}, height=1.8cm, connect spies,
    every spy in node/.append style={line width=0.06cm}}]
                \node [inner sep=0.0cm] {\includegraphics[width=\itemwidth]{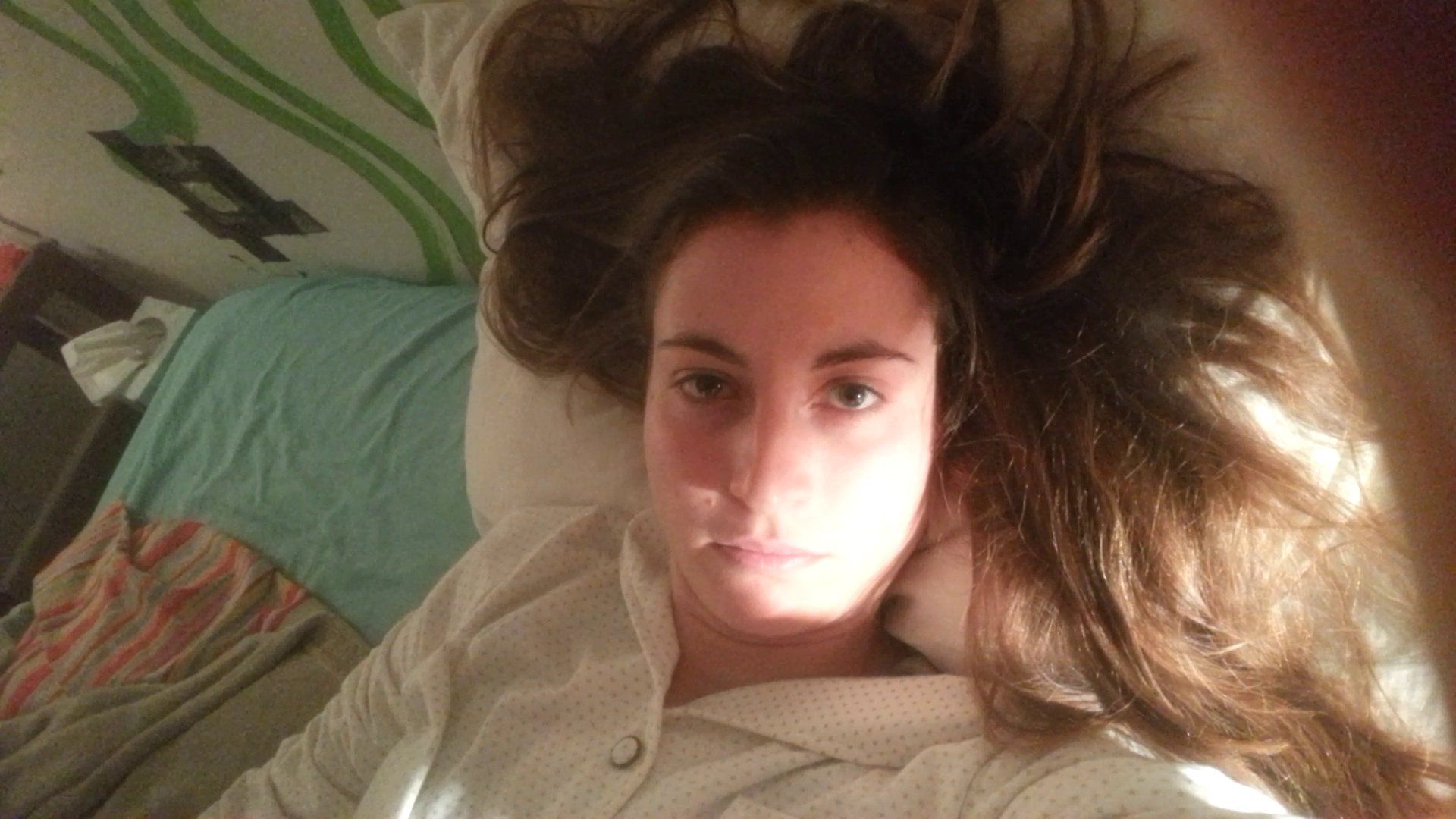}};
                \spy [every spy on node/.append style={line width=0.06cm}, spy connection path={\draw[line width=0.06cm] (tikzspyonnode) -- (tikzspyinnode);}] on (-0.79,0.55) in node at (0.0,-1.85);
            \end{tikzpicture}
        &
            \begin{tikzpicture}[spy using outlines={3787CF, magnification=18, width={\itemwidth - 0.06cm}, height=1.8cm, connect spies,
    every spy in node/.append style={line width=0.06cm}}]
                \node [inner sep=0.0cm] {\includegraphics[width=\itemwidth]{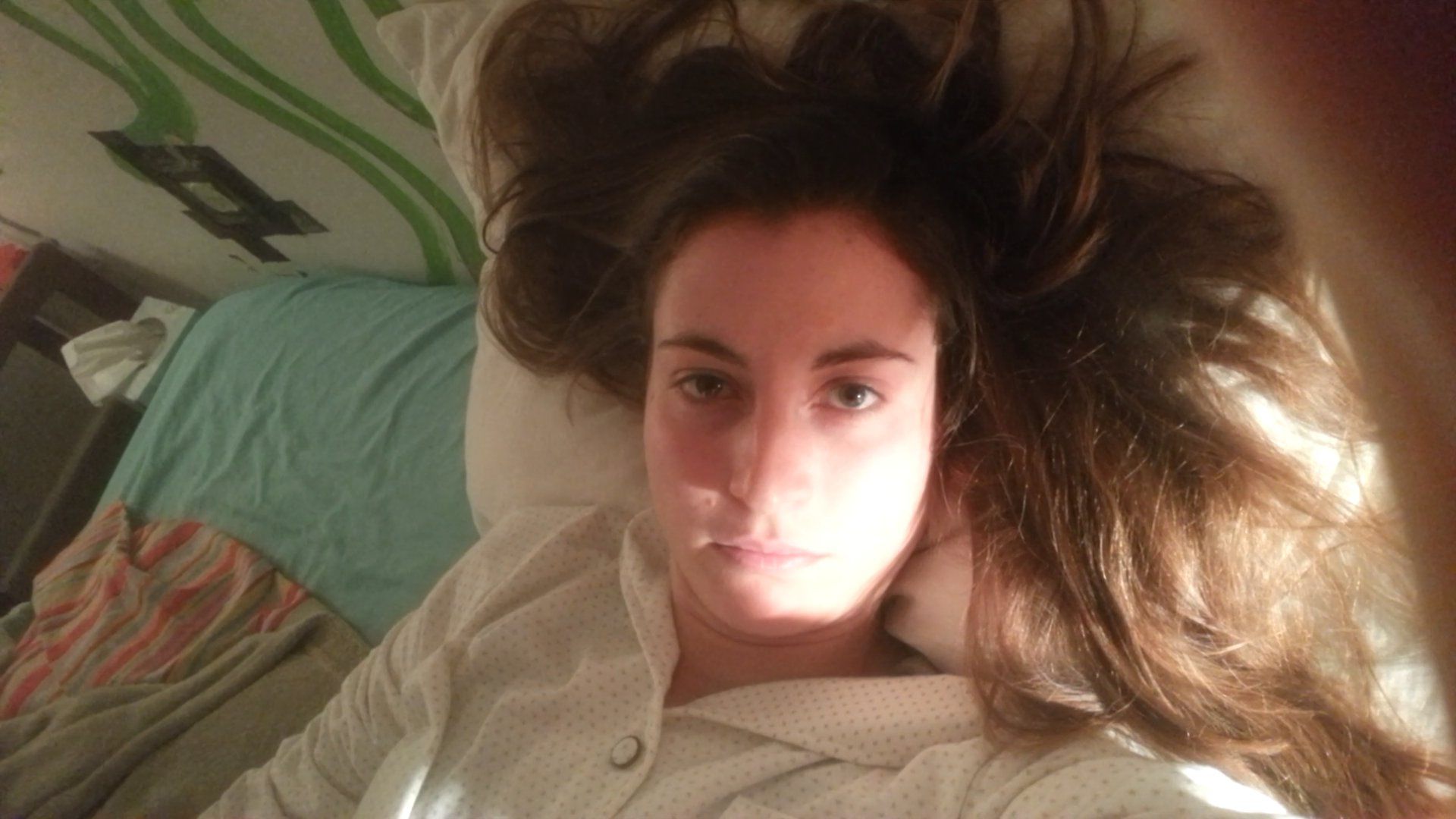}};
                \spy [every spy on node/.append style={line width=0.06cm}, spy connection path={\draw[line width=0.06cm] (tikzspyonnode) -- (tikzspyinnode);}] on (-0.79,0.55) in node at (0.0,-1.85);
            \end{tikzpicture}
        &
            \begin{tikzpicture}[spy using outlines={3787CF, magnification=18, width={\itemwidth - 0.06cm}, height=1.8cm, connect spies,
    every spy in node/.append style={line width=0.06cm}}]
                \node [inner sep=0.0cm] {\includegraphics[width=\itemwidth]{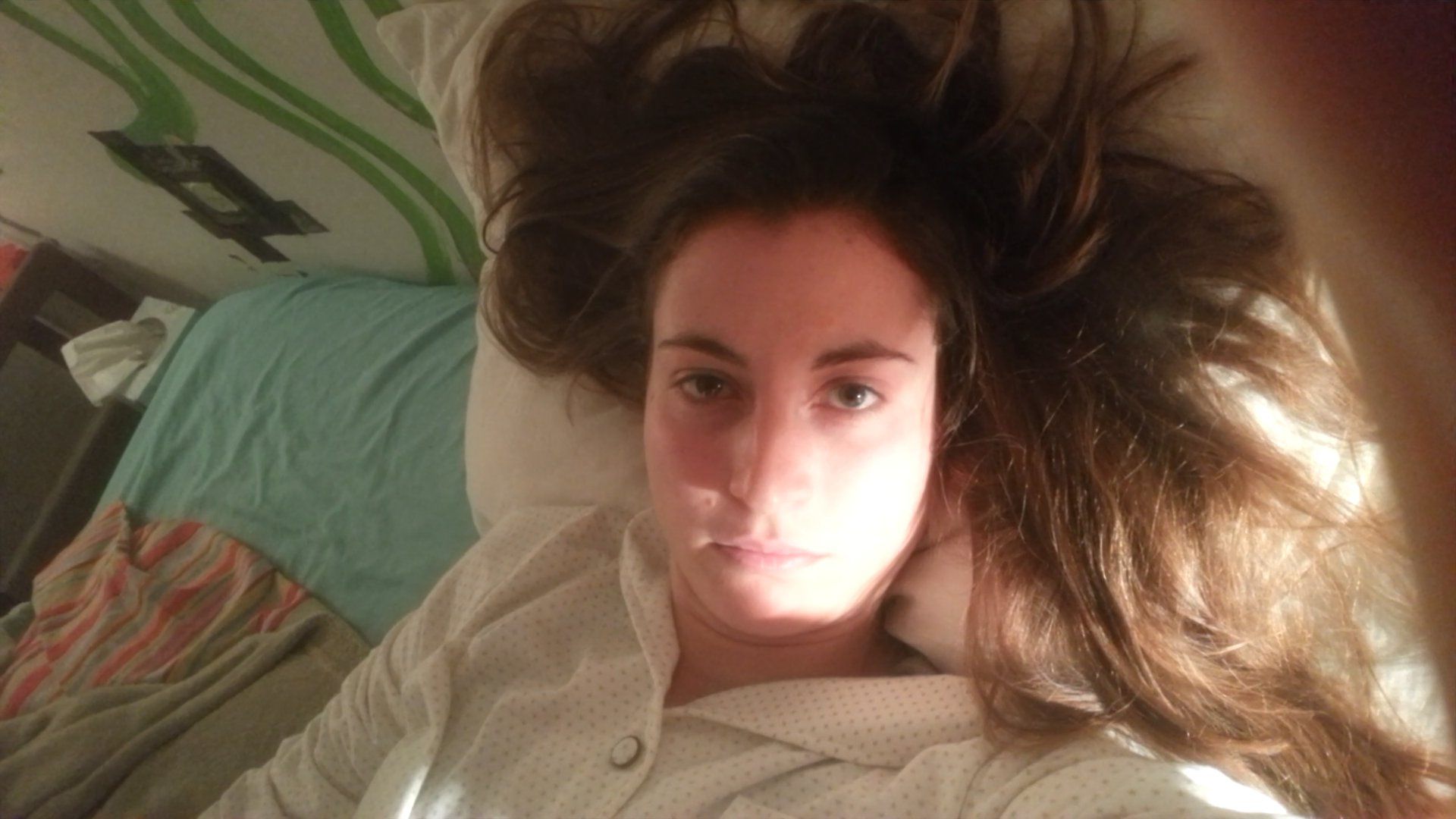}};
                \spy [every spy on node/.append style={line width=0.06cm}, spy connection path={\draw[line width=0.06cm] (tikzspyonnode) -- (tikzspyinnode);}] on (-0.79,0.55) in node at (0.0,-1.85);
            \end{tikzpicture}
        &
            \begin{tikzpicture}[spy using outlines={3787CF, magnification=18, width={\itemwidth - 0.06cm}, height=1.8cm, connect spies,
    every spy in node/.append style={line width=0.06cm}}]
                \node [inner sep=0.0cm] {\includegraphics[width=\itemwidth]{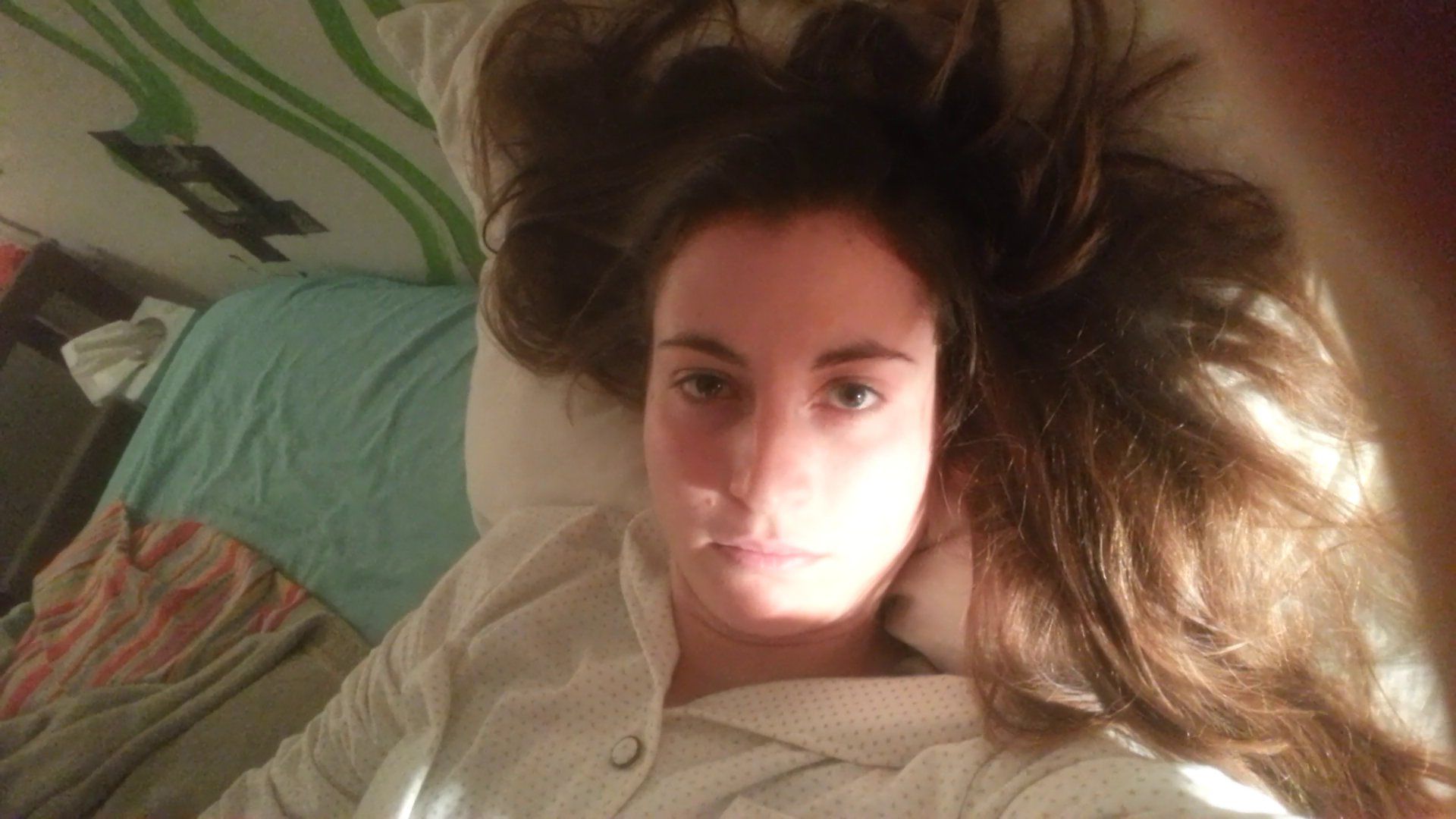}};
                \spy [every spy on node/.append style={line width=0.06cm}, spy connection path={\draw[line width=0.06cm] (tikzspyonnode) -- (tikzspyinnode);}] on (-0.79,0.55) in node at (0.0,-1.85);
            \end{tikzpicture}
        &
            \begin{tikzpicture}[spy using outlines={3787CF, magnification=18, width={\itemwidth - 0.06cm}, height=1.8cm, connect spies,
    every spy in node/.append style={line width=0.06cm}}]
                \node [inner sep=0.0cm] {\includegraphics[width=\itemwidth]{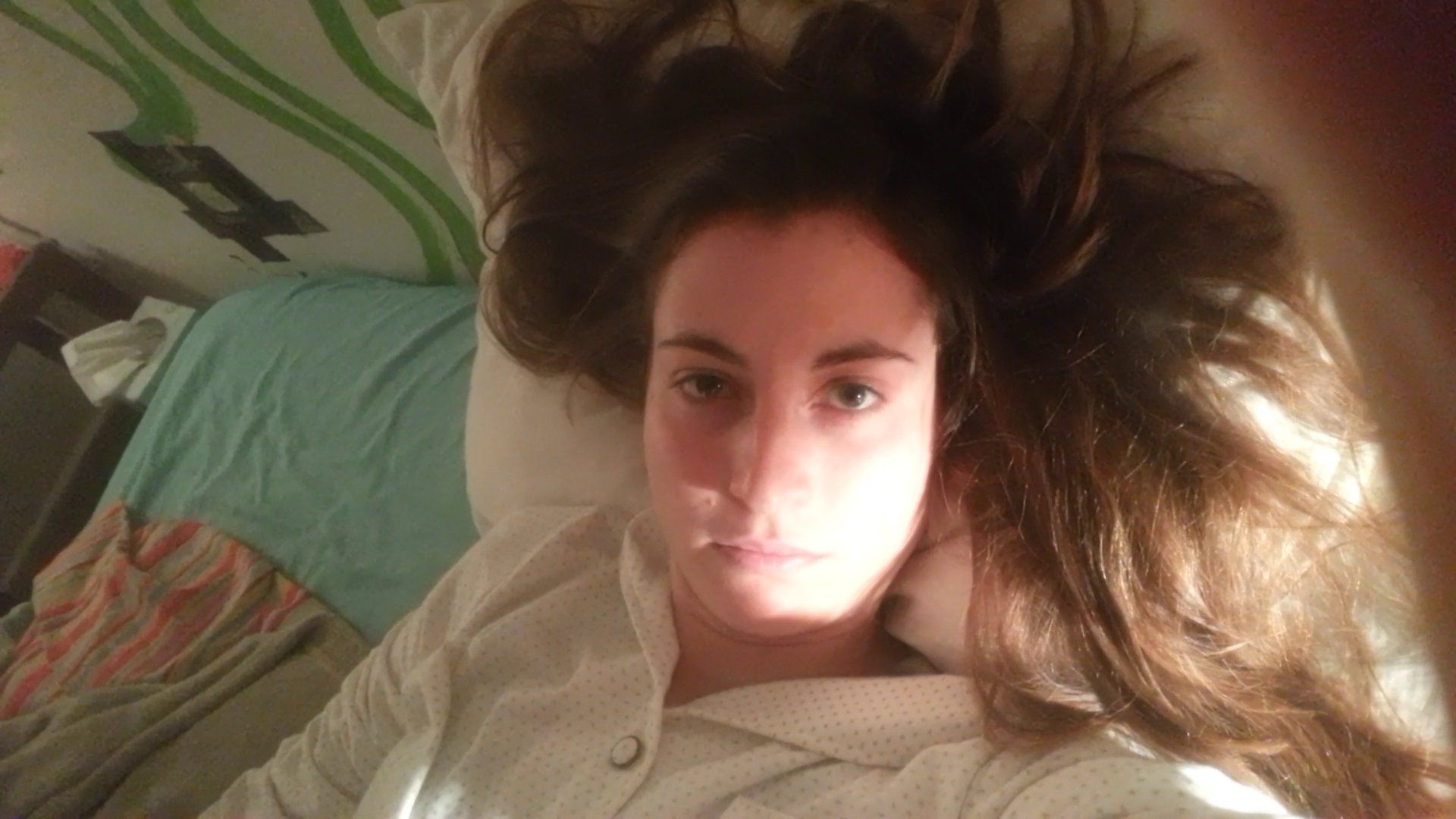}};
                \spy [every spy on node/.append style={line width=0.06cm}, spy connection path={\draw[line width=0.06cm] (tikzspyonnode) -- (tikzspyinnode);}] on (-0.79,0.55) in node at (0.0,-1.85);
            \end{tikzpicture}
        &
            \begin{tikzpicture}[spy using outlines={3787CF, magnification=18, width={\itemwidth - 0.06cm}, height=1.8cm, connect spies,
    every spy in node/.append style={line width=0.06cm}}]
                \node [inner sep=0.0cm] {\includegraphics[width=\itemwidth]{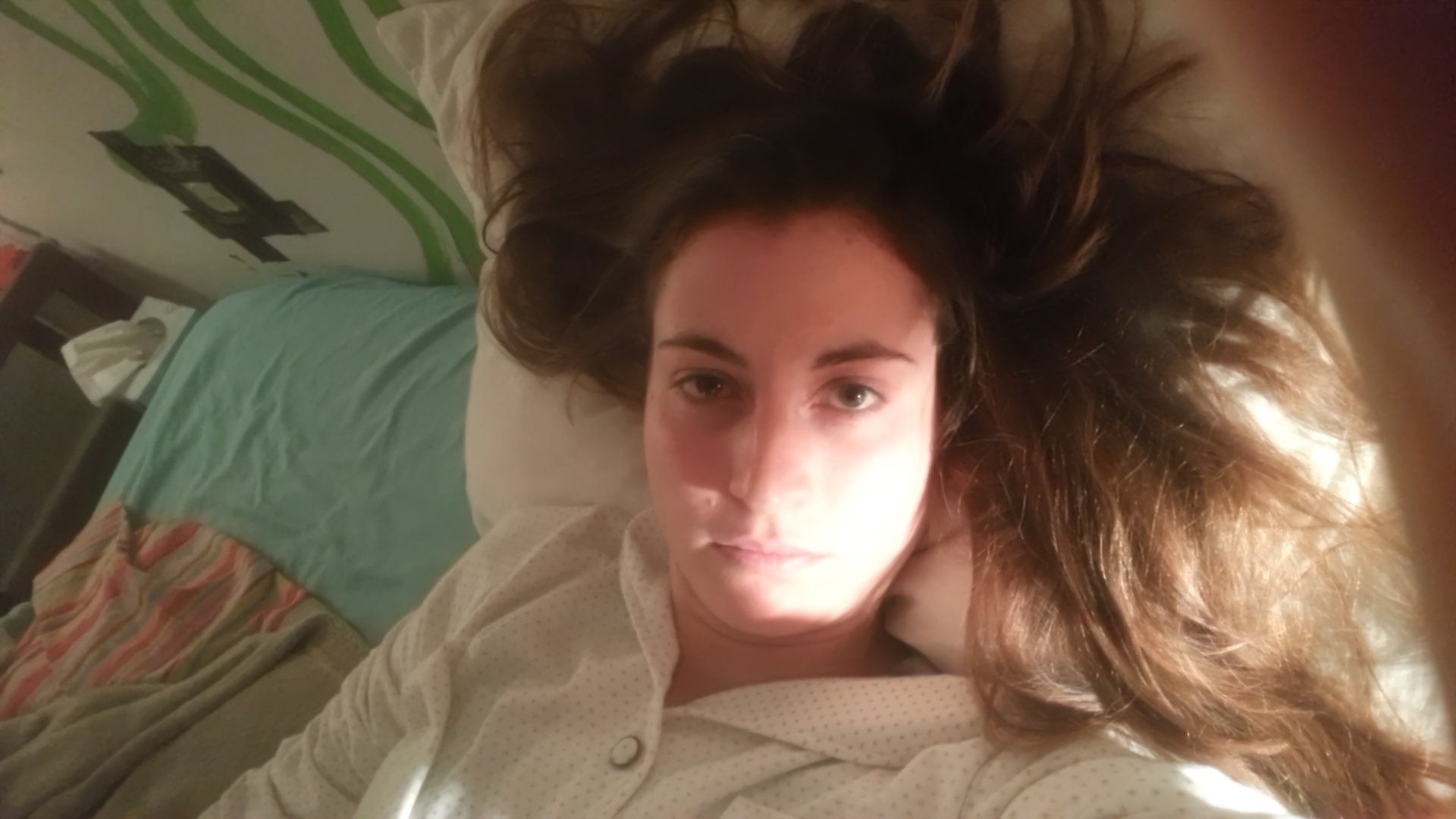}};
                \spy [every spy on node/.append style={line width=0.06cm}, spy connection path={\draw[line width=0.06cm] (tikzspyonnode) -- (tikzspyinnode);}] on (-0.79,0.55) in node at (0.0,-1.85);
            \end{tikzpicture}
        \\
            \footnotesize Input
        &
            \footnotesize reduced temporal only
        &
            \footnotesize temporal only
        &
            \footnotesize reduced spatial only
        &
            \footnotesize spatial only
        &
            \footnotesize default settings
        \vspace{-0.1cm}\\
            
        &
            \scriptsize \scalebox{0.83}[1.0]{( $0.2 \cdot \sigma^2$, $0 \cdot \sigma_d$, $0 \cdot \sigma_r$ )}
        &
            \scriptsize \scalebox{0.83}[1.0]{( $1 \cdot \sigma^2$, $0 \cdot \sigma_d$, $0 \cdot \sigma_r$ )}
        &
            \scriptsize \scalebox{0.83}[1.0]{( $0 \cdot \sigma^2$, $0.2 \cdot \sigma_d$, $0.2 \cdot \sigma_r$ )}
        &
            \scriptsize \scalebox{0.83}[1.0]{( $0 \cdot \sigma^2$, $1 \cdot \sigma_d$, $1 \cdot \sigma_r$ )}
        &
            \scriptsize \scalebox{0.83}[1.0]{( $1 \cdot \sigma^2$, $1 \cdot \sigma_d$, $1 \cdot \sigma_r$ )}
        \\
    \end{tabular}\vspace{-0.38cm}
    \captionof{figure}{By scaling the denoising parameters estimated by $\mathcal{P}(\cdot;\theta)$, our framework makes it easy to control the amount of temporal denoising (through $\sigma^2$) as well as the amount of spatial denoising (through $\sigma_d$ and $\sigma_r$). We scale the denoising parameters equivalently for the luma and chroma channels in this sample, but they can be scaled independently for more control. Please see the supplementary for a video demo that demonstrates how users can control the denoising in an interactive manner. Footage provided by Amit Zinman.}\vspace{-0.48cm}
    \label{fig:control}
\end{figure*}

\subsection{Noise Profiling}

Since noise is not necessarily spatially uniform but often dependent on the signal, we argue that a noise profile should not be just a fixed set of values for the entire input but rather be a descriptor that makes it possible to derive spatially-varying denoising parameters. We see this as an ideal use case for hypernetworks~\cite{ha2016hypernets}, where a noise profiler estimates the weights $\theta$ of a small neural network $\mathcal{P}(\cdot;\theta)$ that predicts spatially-varying denoising parameters. However, instead of directly predicting $\theta$, we stabilize the training through NPA~\cite{ortiz2023npa} where $\theta = \theta^0 + \Delta\theta$ with $\theta^0$ being learnable parameters and $\Delta\theta$ being the hypernet prediction.

Intuitively, analyzing the noise profile of a given image is a mixture of low-level image processing and high-level semantics. Specifically, one first needs to understand what an image depicts before being able to examine the details to discern between unintended noise and actual texture. For this reason, we leverage a pre-trained backbone $\mathcal{B}$ in the form of a ConvNext~\cite{liu2022convnext} with a random MLP head to serve as the hypernet. We have found this backbone to be crucial for the quality of the predicted noise profile.

Lastly, which anchor frame should we choose to analyze the noise of a given video? We would like this choice not to matter, such that the denoising result is independent of having to choose a ``good'' anchor. To achieve this, we always pick the first frame of a video as the anchor frame because it is as good as any, and then utilize a consistency loss
\begin{equation}
    \mathcal{L}_{cstsy} = \left\| \Delta\theta - \Delta\theta_{i} \right\|_2,
\label{eq:consis}
\end{equation}
where we encourage the hypernet prediction of our anchor frame $\Delta\theta$ and any random neighbor $\Delta\theta_{i}$ to be equal.

\subsection{Denoising the Video}

Once we have the noise profile $\theta$, we process the input video one output frame at a tim where we obtain spatially-varying denoising parameters through a small neural network $\mathcal{P}(\cdot;\theta)$ with three convolutional layers. We use these parameters to do temporal and then spatial denoising.

For the temporal denoising, to denoise a given target frame, we first align the two preceding and the two subsequent frames. Specifically, we estimate the optical flow between the target frame and the neighbors using an off-the-shelf SpyNet~\cite{ranjan2017spynet} at a quarter of the target resolution for improved computational efficiency as well as robustness to noise, and warp the neighbors to the target frame using the estimated optical flow. We then merge the aligned frames using a Wiener filter as in Hasinoff~\etal~\cite{hasinoff2016burst} but using a tiles size of $8 \times 8$ and the noise maps from $\mathcal{P}(\cdot;\theta)$ instead of approximating the noise as the RMS in each tile. As is typical and to facilitate more control, we perform Wiener filtering on the luma and the chroma channels independently.

We then apply spatial denoising to the temporally denoised target frame through a bilateral Laplacian pyramid filter with three levels. This borrows many ideas from multiresolution bilateral filtering~\cite{zhang2008mbf} but using a Laplacian instead of a Gaussian pyramid and using bilateral filtering throughout instead of wavelet thresholding for the first pyramid levels. More specifically, we have $\mathcal{P}(\cdot;\theta)$ estimate noise map pyramids and then use these as $\sigma_s$ and $\sigma_r$ for the bilateral filtering. Furthermore, just like with the Wiener filter, we perform the denoising separately on the luma and the chroma channels to facilitate user control.

\subsection{User Control}
\label{sec:control}

Through interviews with creative professionals, we have found that they want to assert their artistic expression when denoising a clip. That is, they often have to decide to leave some noise in the footage in favor of over-smoothing and vice versa. The key to facilitating user control in our method is the spatially-varying parameters from $\mathcal{P}(\cdot;\theta)$ that guide the temporal and spatial denoising. Specifically, we have two of these maps for $\sigma^2$ in the 
Wiener filter (one map for chroma and one for luma) and in the bilateral Laplacian pyramid filter we have two map pyramids for $\sigma_d$ as well as two map pyramids for $\sigma_r$ (one map pyramid for chroma and one for luma). This makes for a total of six sets of parameters where we can facilitate user control.

How do we attenuate these parameters though? As demonstrated in \figref{fig:control}, we have found that simply scaling the spatially-varying denoising parameters uniformly works reasonably well, so we can just provide six tunable knobs to end users. And since our denoising framework runs in real time, changing the knobs provides instantaneous feedback which makes it easy to attain the desired result.

\subsection{Augmentation Pipeline}
\label{sec:degradation}

To train our model, we need pairs of clean and noisy frame sequences just like any other denoiser that leverages machine learning without self-supervision. Specifically, we adopt the typical paradigm of taking a set of ``noise-free'' videos that serve as the ground truth, and augmenting them with various degradations to obtain the corresponding noisy inputs. For this to work well, we have found it to be crucial to include temporal compression in this degradation pipeline since it makes the noise temporally correlated.

But let's start with noise-free videos, which we obtain from the REDS dataset~\cite{nah2019reds}. It provides 240 videos and was shot at a relatively high 120 frames per second, allowing us to augment the frame rate via sub-sampling. We extract random frame sequences from these videos and first degrade them through additive white Gaussian noise with a variance sampled from $\mathcal{U}(1, 50)$. We then transcode them using an H.264 codec through libx264 with a CRF sampled from $\mathcal{U}(18, 30)$ to make the noise temporally correlated.

We have found this data pipeline to work surprisingly well. So well in fact that we retrained the models that we compare to in the evaluation since we were able to get much better results on the CRVD benchmark~\cite{yue2020supervised} with our retrained versions than the original checkpoints. However, we did not retrain Real-ESRGAN~\cite{wang2021real} since it brings its own much more complex augmentation pipeline, and we also did not retrain MF2F~\cite{dewil2021self} and UDVD~\cite{sheth2021unsupervised} since these leverage sophisticated ways of supervising the models which are difficult for us to replicate and properly do justice.

\subsection{Implementation Details}

In terms of architecture details, we utilize the base size of a ConvNext~\cite{liu2022convnext} for the backbone $\mathcal{B}$ in the hypernetwork. To translate the output from this backbone to the residual noise profile $\Delta\theta$, we utilize an MLP with five layers and PReLU activations~\cite{he2015prelu} in between each layer. And for the neural network $\mathcal{P}(\cdot;\theta)$ that predicts the spatially-varying denoising parameters, we leverage three convolution layers with a stride of two and PReLU activations in between. We additionally constrain the output of this network with a softplus~\cite{dugas2000softplus} to facilitate non-negative denoising parameters. Furthermore, we have found it beneficial to not only provide the (aligned) frames to  $\mathcal{P}(\cdot;\theta)$ but also their gradients, approximated by a Sobel filter, as well as masks that indicate whether or not a pixel in an aligned frame is valid (it may have been warped from outside the frame)~\cite{bhat2023burst}.

In our reconstruction loss, we minimize the difference between the temporally denoised image $I^\text{T}$ and the ground truth $I^\text{GT}$ as well as the difference between all levels of the spatial denoising pyramid $I^\text{ST}$ and $I^\text{GT}$ as
\begin{equation}
    \mathcal{L}_{rec} = \left\| I^\text{T} - I^\text{GT} \right\|_2 + \sum_{l = 1}^{3} \left\| I^\text{ST}_l - I^\text{GT}_l \right\|_2,
\end{equation}
where $l$ is the $l$-th level in spatial denoising pyramid. Recall that we also have a consistency loss that we outlined in Eq.~\eqref{eq:consis} in the section about noise profiling.

We train our model using Adam~\cite{kingma2015adam} with an initial learning rate of $2 \times 10^{-4}$ that decays to $1\times 10^{-7}$ using a cosine annealing schedule. In total, we train the model for 400 thousand iterations with a batch size of 24 consisting of patches with $512 \times 512$ pixels. Since our pipeline is computationally lightweight, this takes less than two days.
\section{Experiments}
\label{sec:exp}

\begin{figure}\centering
    \setlength{\tabcolsep}{0.0cm}
    \renewcommand{\arraystretch}{1.2}
    \newcommand{\quantTit}[1]{\multicolumn{2}{c}{\scriptsize #1}}
    \newcommand{\quantSec}[1]{\scriptsize #1}
    \newcommand{\quantInd}[1]{\multicolumn{2}{c}{\tiny #1}}
    \newcommand{\quantVal}[1]{\scalebox{0.83}[1.0]{$ #1 $}}
    \footnotesize
    \begin{tabularx}{\columnwidth}{@{\hspace{0.1cm}} X P{1.03cm} @{\hspace{-0.31cm}} P{1.31cm} P{1.03cm} @{\hspace{-0.31cm}} P{1.31cm} P{1.03cm} @{\hspace{-0.31cm}} P{1.31cm}}
        \toprule
            & \quantSec{PSNR} & delta & \quantSec{SSIM} & delta & \quantSec{LPIPS} & delta
        \\[-0.1cm]
            & \quantInd{\hspace{-0.1cm}(higher PSNR is better)} & \quantInd{\hspace{-0.1cm}(higher SSIM is better)} & \quantInd{\hspace{-0.15cm}(lower LPIPS is better)}
        \\ \midrule
            NAFNet~\cite{chen2022simple} & \quantVal{30.25} & \quantVal{-} & \quantVal{0.747} & \quantVal{-} & \quantVal{0.358} & \quantVal{-} \\
            NAFNet$^\dagger$~\cite{chen2022simple} & \quantVal{35.32} & \quantVal{\text{+ } 5.07} & \quantVal{0.937} & \quantVal{\text{+ } 0.190} & \quantVal{0.089} & \quantVal{\text{- } 0.269} \\
            FastDVDNet~\cite{tassano2020fastdvdnet} & \quantVal{27.89} & \quantVal{-} & \quantVal{0.565} & \quantVal{-} & \quantVal{0.502} & \quantVal{-} \\
            FastDVDNet$^\dagger$~\cite{tassano2020fastdvdnet} & \quantVal{34.63} & \quantVal{\text{+ } 6.75} & \quantVal{0.914} & \quantVal{\text{+ } 0.349} & \quantVal{0.124} & \quantVal{\text{- } 0.378} \\
            TOFlow~\cite{xue2019video} & \quantVal{25.96} & \quantVal{-} & \quantVal{0.673} & \quantVal{-} & \quantVal{0.251} & \quantVal{-} \\
            TOFlow$^\dagger$~\cite{xue2019video} & \quantVal{34.08} & \quantVal{\text{+ } 8.11} & \quantVal{0.903} & \quantVal{\text{+ } 0.231} & \quantVal{0.147} & \quantVal{\text{- } 0.104} \\
            BasicVSR++~\cite{chan2022basicvsr++} & \quantVal{31.98} & \quantVal{-} & \quantVal{0.769} & \quantVal{-} & \quantVal{0.326} & \quantVal{-} \\
            BasicVSR++$^\dagger$~\cite{chan2022basicvsr++} & \quantVal{34.34} & \quantVal{\text{+ } 2.35} & \quantVal{0.873} & \quantVal{\text{+ } 0.104} & \quantVal{0.167} & \quantVal{\text{- } 0.158} \\
            VRT~\cite{liang2024vrt} & \quantVal{31.99} & \quantVal{-} & \quantVal{0.784} & \quantVal{-} & \quantVal{0.296} & \quantVal{-} \\
            VRT$^\dagger$~\cite{liang2024vrt} & \quantVal{33.86} & \quantVal{\text{+ } 1.86} & \quantVal{0.848} & \quantVal{\text{+ } 0.064} & \quantVal{0.192} & \quantVal{\text{- } 0.104} \\
        \bottomrule
    \end{tabularx}\vspace{-0.2cm}
    \captionof{table}{Denoising results on the CRVD (sRGB) benchmark for various methods that we compare to, both for their original version as well as our retrained one which we denote with a $\dagger$.}\vspace{-0.3cm}
    \label{tbl:retrained}
\end{figure}

\begin{figure*}\centering
    \setlength{\tabcolsep}{0.0cm}
    \renewcommand{\arraystretch}{1.2}
    \newcommand{\quantTit}[1]{\multicolumn{2}{c}{\scriptsize #1}}
    \newcommand{\quantSec}[1]{\scriptsize #1}
    \newcommand{\quantInd}[1]{\multicolumn{2}{c}{\tiny #1}}
    \newcommand{\quantVal}[1]{\scalebox{0.83}[1.0]{$ #1 $}}
    \newcommand{\quantFirst}[1]{\usolid{\scalebox{0.83}[1.0]{$ #1 $}}}
    \footnotesize
    \begin{tabularx}{\textwidth}{@{\hspace{0.1cm}} X P{1.07cm} @{\hspace{-0.31cm}} P{1.35cm} P{1.07cm} @{\hspace{-0.31cm}} P{1.35cm} P{1.07cm} @{\hspace{-0.31cm}} P{1.35cm} P{1.07cm} @{\hspace{-0.31cm}} P{1.35cm} P{1.07cm} @{\hspace{-0.31cm}} P{1.35cm} P{1.07cm} @{\hspace{-0.31cm}} P{1.35cm} P{1.07cm} @{\hspace{-0.31cm}} P{1.35cm}}
        \toprule
            & \quantTit{ISO 1600} & \quantTit{ISO 3200} & \quantTit{ISO 6400} & \quantTit{ISO 12800} & \quantTit{ISO 25600} & \quantTit{Overall} & \quantTit{Speed}
        \\ \cmidrule(l{2pt}r{2pt}){2-3} \cmidrule(l{2pt}r{2pt}){4-5} \cmidrule(l{2pt}r{2pt}){6-7} \cmidrule(l{2pt}r{2pt}){8-9} \cmidrule(l{2pt}r{2pt}){10-11} \cmidrule(l{2pt}r{2pt}){12-13} \cmidrule(l{2pt}r{2pt}){14-15}
            & \quantSec{PSNR} & rank & \quantSec{PSNR} & rank & \quantSec{PSNR} & rank & \quantSec{PSNR} & rank & \quantSec{PSNR} & rank & \quantSec{PSNR} & rank & \quantSec{FPS} & rank
        \\[-0.1cm]
            & \quantInd{\hspace{-0.15cm}(higher PSNR is better)} & \quantInd{\hspace{-0.15cm}(higher PSNR is better)} & \quantInd{\hspace{-0.15cm}(higher PSNR is better)} & \quantInd{\hspace{-0.15cm}(higher PSNR is better)} & \quantInd{\hspace{-0.15cm}(higher PSNR is better)} & \quantInd{\hspace{-0.15cm}(higher PSNR is better)} & \quantInd{\hspace{-0.1cm}(higher FPS is better)}
        \\ \midrule
            SID$^\dagger$~\cite{chen2018learning} & \quantVal{38.85} & \quantVal{\hphantom{0}7^\text{\parbox{0.15cm}{th}}\text{ of }10} & \quantVal{37.68} & \quantVal{\hphantom{0}7^\text{\parbox{0.15cm}{th}}\text{ of }10} & \quantVal{35.82} & \quantVal{\hphantom{0}4^\text{\parbox{0.15cm}{th}}\text{ of }10} & \quantVal{33.51} & \quantVal{\hphantom{0}4^\text{\parbox{0.15cm}{th}}\text{ of }10} & \quantVal{29.18} & \quantVal{\hphantom{0}3^\text{\parbox{0.15cm}{rd}}\text{ of }10} & \quantVal{35.01} & \quantVal{\hphantom{0}4^\text{\parbox{0.15cm}{th}}\text{ of }10} & \quantVal{6.95} & \quantVal{\hphantom{0}3^\text{\parbox{0.15cm}{rd}}\text{ of }10} \\
            NAFNet$^\dagger$~\cite{chen2022simple} & \quantVal{39.48} & \quantVal{\hphantom{0}3^\text{\parbox{0.15cm}{rd}}\text{ of }10} & \quantVal{38.12} & \quantVal{\hphantom{0}5^\text{\parbox{0.15cm}{th}}\text{ of }10} & \quantVal{35.94} & \quantVal{\hphantom{0}3^\text{\parbox{0.15cm}{rd}}\text{ of }10} & \quantVal{33.53} & \quantVal{\hphantom{0}3^\text{\parbox{0.15cm}{rd}}\text{ of }10} & \quantVal{29.55} & \quantVal{\hphantom{0}2^\text{\parbox{0.15cm}{nd}}\text{ of }10} & \quantVal{35.32} & \quantVal{\hphantom{0}2^\text{\parbox{0.15cm}{nd}}\text{ of }10} & \quantVal{1.69} & \quantVal{\hphantom{0}7^\text{\parbox{0.15cm}{th}}\text{ of }10} \\
            FastDVDNet$^\dagger$~\cite{tassano2020fastdvdnet} & \quantVal{39.16} & \quantVal{\hphantom{0}5^\text{\parbox{0.15cm}{th}}\text{ of }10} & \quantVal{37.92} & \quantVal{\hphantom{0}6^\text{\parbox{0.15cm}{th}}\text{ of }10} & \quantVal{35.60} & \quantVal{\hphantom{0}5^\text{\parbox{0.15cm}{th}}\text{ of }10} & \quantVal{32.56} & \quantVal{\hphantom{0}5^\text{\parbox{0.15cm}{th}}\text{ of }10} & \quantVal{27.93} & \quantVal{\hphantom{0}6^\text{\parbox{0.15cm}{th}}\text{ of }10} & \quantVal{34.63} & \quantVal{\hphantom{0}5^\text{\parbox{0.15cm}{th}}\text{ of }10} & \quantVal{5.72} & \quantVal{\hphantom{0}4^\text{\parbox{0.15cm}{th}}\text{ of }10} \\
            TOFlow$^\dagger$~\cite{xue2019video} & \quantVal{38.25} & \quantVal{\hphantom{0}8^\text{\parbox{0.15cm}{th}}\text{ of }10} & \quantVal{36.97} & \quantVal{\hphantom{0}8^\text{\parbox{0.15cm}{th}}\text{ of }10} & \quantVal{34.90} & \quantVal{\hphantom{0}7^\text{\parbox{0.15cm}{th}}\text{ of }10} & \quantVal{32.21} & \quantVal{\hphantom{0}6^\text{\parbox{0.15cm}{th}}\text{ of }10} & \quantVal{28.07} & \quantVal{\hphantom{0}5^\text{\parbox{0.15cm}{th}}\text{ of }10} & \quantVal{34.08} & \quantVal{\hphantom{0}7^\text{\parbox{0.15cm}{th}}\text{ of }10} & \quantVal{2.84} & \quantVal{\hphantom{0}6^\text{\parbox{0.15cm}{th}}\text{ of }10} \\
            BasicVSR++$^\dagger$~\cite{chan2022basicvsr++} & \quantVal{39.40} & \quantVal{\hphantom{0}4^\text{\parbox{0.15cm}{th}}\text{ of }10} & \quantVal{38.24} & \quantVal{\hphantom{0}2^\text{\parbox{0.15cm}{nd}}\text{ of }10} & \quantVal{35.55} & \quantVal{\hphantom{0}6^\text{\parbox{0.15cm}{th}}\text{ of }10} & \quantVal{31.72} & \quantVal{\hphantom{0}7^\text{\parbox{0.15cm}{th}}\text{ of }10} & \quantVal{26.78} & \quantVal{\hphantom{0}9^\text{\parbox{0.15cm}{th}}\text{ of }10} & \quantVal{34.34} & \quantVal{\hphantom{0}6^\text{\parbox{0.15cm}{th}}\text{ of }10} & \quantVal{7.41} & \quantVal{\hphantom{0}2^\text{\parbox{0.15cm}{nd}}\text{ of }10} \\
            VRT$^\dagger$~\cite{liang2024vrt} & \quantVal{39.55} & \quantVal{\hphantom{0}2^\text{\parbox{0.15cm}{nd}}\text{ of }10} & \quantVal{38.12} & \quantVal{\hphantom{0}4^\text{\parbox{0.15cm}{th}}\text{ of }10} & \quantVal{34.82} & \quantVal{\hphantom{0}8^\text{\parbox{0.15cm}{th}}\text{ of }10} & \quantVal{30.77} & \quantVal{\hphantom{0}8^\text{\parbox{0.15cm}{th}}\text{ of }10} & \quantVal{26.01} & \quantVal{10^\text{\parbox{0.15cm}{th}}\text{ of }10} & \quantVal{33.86} & \quantVal{\hphantom{0}8^\text{\parbox{0.15cm}{th}}\text{ of }10} & \quantVal{0.05} & \quantVal{10^\text{\parbox{0.15cm}{th}}\text{ of }10} \\
            Real-ESRGAN~\cite{wang2021real} & \quantVal{29.98} & \quantVal{10^\text{\parbox{0.15cm}{th}}\text{ of }10} & \quantVal{28.27} & \quantVal{10^\text{\parbox{0.15cm}{th}}\text{ of }10} & \quantVal{28.04} & \quantVal{10^\text{\parbox{0.15cm}{th}}\text{ of }10} & \quantVal{27.52} & \quantVal{10^\text{\parbox{0.15cm}{th}}\text{ of }10} & \quantVal{27.63} & \quantVal{\hphantom{0}8^\text{\parbox{0.15cm}{th}}\text{ of }10} & \quantVal{28.29} & \quantVal{10^\text{\parbox{0.15cm}{th}}\text{ of }10} & \quantVal{0.24} & \quantVal{\hphantom{0}8^\text{\parbox{0.15cm}{th}}\text{ of }10} \\
            UDVD~\cite{sheth2021unsupervised} & \quantVal{31.15} & \quantVal{\hphantom{0}9^\text{\parbox{0.15cm}{th}}\text{ of }10} & \quantVal{30.72} & \quantVal{\hphantom{0}9^\text{\parbox{0.15cm}{th}}\text{ of }10} & \quantVal{30.23} & \quantVal{\hphantom{0}9^\text{\parbox{0.15cm}{th}}\text{ of }10} & \quantVal{29.10} & \quantVal{\hphantom{0}9^\text{\parbox{0.15cm}{th}}\text{ of }10} & \quantVal{27.63} & \quantVal{\hphantom{0}7^\text{\parbox{0.15cm}{th}}\text{ of }10} & \quantVal{29.77} & \quantVal{\hphantom{0}9^\text{\parbox{0.15cm}{th}}\text{ of }10} & \quantVal{0.16} & \quantVal{\hphantom{0}9^\text{\parbox{0.15cm}{th}}\text{ of }10} \\
            MF2F~\cite{dewil2021self} & \quantVal{39.09} & \quantVal{\hphantom{0}6^\text{\parbox{0.15cm}{th}}\text{ of }10} & \quantVal{38.20} & \quantVal{\hphantom{0}3^\text{\parbox{0.15cm}{rd}}\text{ of }10} & \quantFirst{36.36} & \quantVal{\hphantom{0}1^\text{\parbox{0.15cm}{st}}\text{ of }10} & \quantVal{33.57} & \quantVal{\hphantom{0}2^\text{\parbox{0.15cm}{nd}}\text{ of }10} & \quantVal{29.04} & \quantVal{\hphantom{0}4^\text{\parbox{0.15cm}{th}}\text{ of }10} & \quantVal{35.25} & \quantVal{\hphantom{0}3^\text{\parbox{0.15cm}{rd}}\text{ of }10} & \quantVal{4.62} & \quantVal{\hphantom{0}5^\text{\parbox{0.15cm}{th}}\text{ of }10} \\
            Ours - RFCVD & \quantFirst{40.35} & \quantVal{\hphantom{0}1^\text{\parbox{0.15cm}{st}}\text{ of }10} & \quantFirst{38.60} & \quantVal{\hphantom{0}1^\text{\parbox{0.15cm}{st}}\text{ of }10} & \quantVal{36.28} & \quantVal{\hphantom{0}2^\text{\parbox{0.15cm}{nd}}\text{ of }10} & \quantFirst{33.86} & \quantVal{\hphantom{0}1^\text{\parbox{0.15cm}{st}}\text{ of }10} & \quantFirst{31.12} & \quantVal{\hphantom{0}1^\text{\parbox{0.15cm}{st}}\text{ of }10} & \quantFirst{36.04} & \quantVal{\hphantom{0}1^\text{\parbox{0.15cm}{st}}\text{ of }10} & \quantFirst{31.66} & \quantVal{\hphantom{0}1^\text{\parbox{0.15cm}{st}}\text{ of }10} \\        
        \bottomrule
    \end{tabularx}\vspace{-0.2cm}
    \captionof{table}{Video denoising results on the CRVD (sRGB) benchmark. Not only does our approach perform best overall, it is also four times faster than the second-fastest place. Please kindly see the supplementary for SSIM and LPIPS where our approach ranks first as well.}\vspace{-0.3cm}
    \label{tbl:crvd}
\end{figure*}

\begin{figure*}\centering
    \setlength{\tabcolsep}{0.0cm}
    \renewcommand{\arraystretch}{1.2}
    \newcommand{\quantTit}[1]{\multicolumn{6}{c}{\scriptsize #1}}
    \newcommand{\quantFps}[1]{\multicolumn{2}{c}{\scriptsize #1}}
    \newcommand{\quantSec}[1]{\scriptsize #1}
    \newcommand{\quantInd}[1]{\multicolumn{2}{c}{\tiny #1}}
    \newcommand{\quantVal}[1]{\scalebox{0.83}[1.0]{$ #1 $}}
    \newcommand{\quantFirst}[1]{\usolid{\scalebox{0.83}[1.0]{$ #1 $}}}
    \footnotesize
    \begin{tabularx}{\textwidth}{@{\hspace{0.1cm}} X P{1.07cm} @{\hspace{-0.31cm}} P{1.35cm} P{1.07cm} @{\hspace{-0.31cm}} P{1.35cm} P{1.07cm} @{\hspace{-0.31cm}} P{1.35cm} P{1.07cm} @{\hspace{-0.31cm}} P{1.35cm} P{1.07cm} @{\hspace{-0.31cm}} P{1.35cm} P{1.07cm} @{\hspace{-0.31cm}} P{1.35cm} P{1.07cm} @{\hspace{-0.31cm}} P{1.35cm}}
        \toprule
            & \quantTit{film grain noise w/ AV1 compression} & \quantTit{spatially correlated noise w/ H.265 compression} & \quantFps{Speed}
        \\ \cmidrule(l{2pt}r{2pt}){2-7} \cmidrule(l{2pt}r{2pt}){8-13} \cmidrule(l{2pt}r{2pt}){14-15}
            & \quantSec{PSNR} & rank & \quantSec{SSIM} & rank & \quantSec{LPIPS} & rank & \quantSec{PSNR} & rank & \quantSec{SSIM} & rank & \quantSec{LPIPS} & rank & \quantSec{FPS} & rank
        \\[-0.1cm]
            & \quantInd{\hspace{-0.15cm}(higher PSNR is better)} & \quantInd{\hspace{-0.1cm}(higher SSIM is better)} & \quantInd{\hspace{-0.2cm}(lower LPIPS is better)} & \quantInd{\hspace{-0.15cm}(higher PSNR is better)} & \quantInd{\hspace{-0.1cm}(higher SSIM is better)} & \quantInd{\hspace{-0.2cm}(lower LPIPS is better)} & \quantInd{\hspace{-0.1cm}(higher FPS is better)}
        \\ \midrule
            SID$^\dagger$~\cite{chen2018learning} & \quantVal{27.05} & \quantVal{5^\text{\parbox{0.15cm}{th}}\text{ of }7} & \quantVal{0.664} & \quantVal{5^\text{\parbox{0.15cm}{th}}\text{ of }7} & \quantVal{0.293} & \quantVal{4^\text{\parbox{0.15cm}{th}}\text{ of }7} & \quantVal{28.44} & \quantVal{3^\text{\parbox{0.15cm}{rd}}\text{ of }7} & \quantVal{0.771} & \quantVal{4^\text{\parbox{0.15cm}{th}}\text{ of }7} & \quantVal{0.282} & \quantVal{5^\text{\parbox{0.15cm}{th}}\text{ of }7} & \quantVal{15.00} & \quantVal{3^\text{\parbox{0.15cm}{rd}}\text{ of }7} \\
            NAFNet$^\dagger$~\cite{chen2022simple} & \quantVal{27.16} & \quantVal{4^\text{\parbox{0.15cm}{th}}\text{ of }7} & \quantVal{0.672} & \quantVal{4^\text{\parbox{0.15cm}{th}}\text{ of }7} & \quantVal{0.294} & \quantVal{5^\text{\parbox{0.15cm}{th}}\text{ of }7} & \quantVal{28.40} & \quantVal{4^\text{\parbox{0.15cm}{th}}\text{ of }7} & \quantVal{0.764} & \quantVal{6^\text{\parbox{0.15cm}{th}}\text{ of }7} & \quantVal{0.257} & \quantVal{3^\text{\parbox{0.15cm}{rd}}\text{ of }7} & \quantVal{3.812} & \quantVal{6^\text{\parbox{0.15cm}{th}}\text{ of }7} \\
            FastDVDNet$^\dagger$~\cite{tassano2020fastdvdnet} & \quantVal{27.39} & \quantVal{3^\text{\parbox{0.15cm}{rd}}\text{ of }7} & \quantVal{0.686} & \quantVal{3^\text{\parbox{0.15cm}{rd}}\text{ of }7} & \quantVal{0.272} & \quantVal{3^\text{\parbox{0.15cm}{rd}}\text{ of }7} & \quantVal{27.92} & \quantVal{6^\text{\parbox{0.15cm}{th}}\text{ of }7} & \quantVal{0.769} & \quantVal{5^\text{\parbox{0.15cm}{th}}\text{ of }7} & \quantVal{0.296} & \quantVal{6^\text{\parbox{0.15cm}{th}}\text{ of }7} & \quantVal{12.09} & \quantVal{4^\text{\parbox{0.15cm}{th}}\text{ of }7} \\
            TOFlow$^\dagger$~\cite{xue2019video} & \quantVal{28.15} & \quantVal{2^\text{\parbox{0.15cm}{nd}}\text{ of }7} & \quantVal{0.750} & \quantVal{2^\text{\parbox{0.15cm}{nd}}\text{ of }7} & \quantFirst{0.220} & \quantVal{1^\text{\parbox{0.15cm}{st}}\text{ of }7} & \quantVal{28.39} & \quantVal{5^\text{\parbox{0.15cm}{th}}\text{ of }7} & \quantVal{0.788} & \quantVal{3^\text{\parbox{0.15cm}{rd}}\text{ of }7} & \quantVal{0.258} & \quantVal{4^\text{\parbox{0.15cm}{th}}\text{ of }7} & \quantVal{5.665} & \quantVal{5^\text{\parbox{0.15cm}{th}}\text{ of }7} \\
            BasicVSR++$^\dagger$~\cite{chan2022basicvsr++} & \quantVal{26.90} & \quantVal{6^\text{\parbox{0.15cm}{th}}\text{ of }7} & \quantVal{0.651} & \quantVal{6^\text{\parbox{0.15cm}{th}}\text{ of }7} & \quantVal{0.313} & \quantVal{6^\text{\parbox{0.15cm}{th}}\text{ of }7} & \quantVal{27.48} & \quantVal{7^\text{\parbox{0.15cm}{th}}\text{ of }7} & \quantVal{0.728} & \quantVal{7^\text{\parbox{0.15cm}{th}}\text{ of }7} & \quantVal{0.358} & \quantVal{7^\text{\parbox{0.15cm}{th}}\text{ of }7} & \quantVal{15.32} & \quantVal{2^\text{\parbox{0.15cm}{nd}}\text{ of }7} \\
            VRT$^\dagger$~\cite{liang2024vrt} & \quantVal{26.55} & \quantVal{7^\text{\parbox{0.15cm}{th}}\text{ of }7} & \quantVal{0.629} & \quantVal{7^\text{\parbox{0.15cm}{th}}\text{ of }7} & \quantVal{0.331} & \quantVal{7^\text{\parbox{0.15cm}{th}}\text{ of }7} & \quantVal{28.78} & \quantVal{2^\text{\parbox{0.15cm}{nd}}\text{ of }7} & \quantVal{0.803} & \quantVal{2^\text{\parbox{0.15cm}{nd}}\text{ of }7} & \quantFirst{0.206} & \quantVal{1^\text{\parbox{0.15cm}{st}}\text{ of }7} & \quantVal{0.105} & \quantVal{7^\text{\parbox{0.15cm}{th}}\text{ of }7} \\
            Ours - RFCVD & \quantFirst{28.59} & \quantVal{1^\text{\parbox{0.15cm}{st}}\text{ of }7} & \quantFirst{0.774} & \quantVal{1^\text{\parbox{0.15cm}{st}}\text{ of }7} & \quantVal{0.247} & \quantVal{2^\text{\parbox{0.15cm}{nd}}\text{ of }7} & \quantFirst{28.93} & \quantVal{1^\text{\parbox{0.15cm}{st}}\text{ of }7} & \quantFirst{0.808} & \quantVal{1^\text{\parbox{0.15cm}{st}}\text{ of }7} & \quantVal{0.239} & \quantVal{2^\text{\parbox{0.15cm}{nd}}\text{ of }7} & \quantFirst{69.73} & \quantVal{1^\text{\parbox{0.15cm}{st}}\text{ of }7} \\              
        \bottomrule
    \end{tabularx}\vspace{-0.2cm}
    \captionof{table}{Testing the generalizability when training on AWGN with H.264 transcoding but testing with two different degradation schemes. Since our approach is based on classic methods, the denoising itself only has a few parameters which naturally reduces domain gaps.}\vspace{-0.0cm}
    \label{tbl:generalizability}
\end{figure*}

\begin{figure*}
    \centering
    \setlength{\tabcolsep}{0.05cm}
    \setlength{\itemwidth}{2.85cm}
    \hspace*{-\tabcolsep}\begin{tabular}{cccccc}
            \begin{tikzpicture}[spy using outlines={3787CF, magnification=8, width={\itemwidth - 0.06cm}, height=2.55cm, connect spies,
        every spy in node/.append style={line width=0.06cm}}]
                \node [inner sep=0.0cm] {\includegraphics[width=\itemwidth]{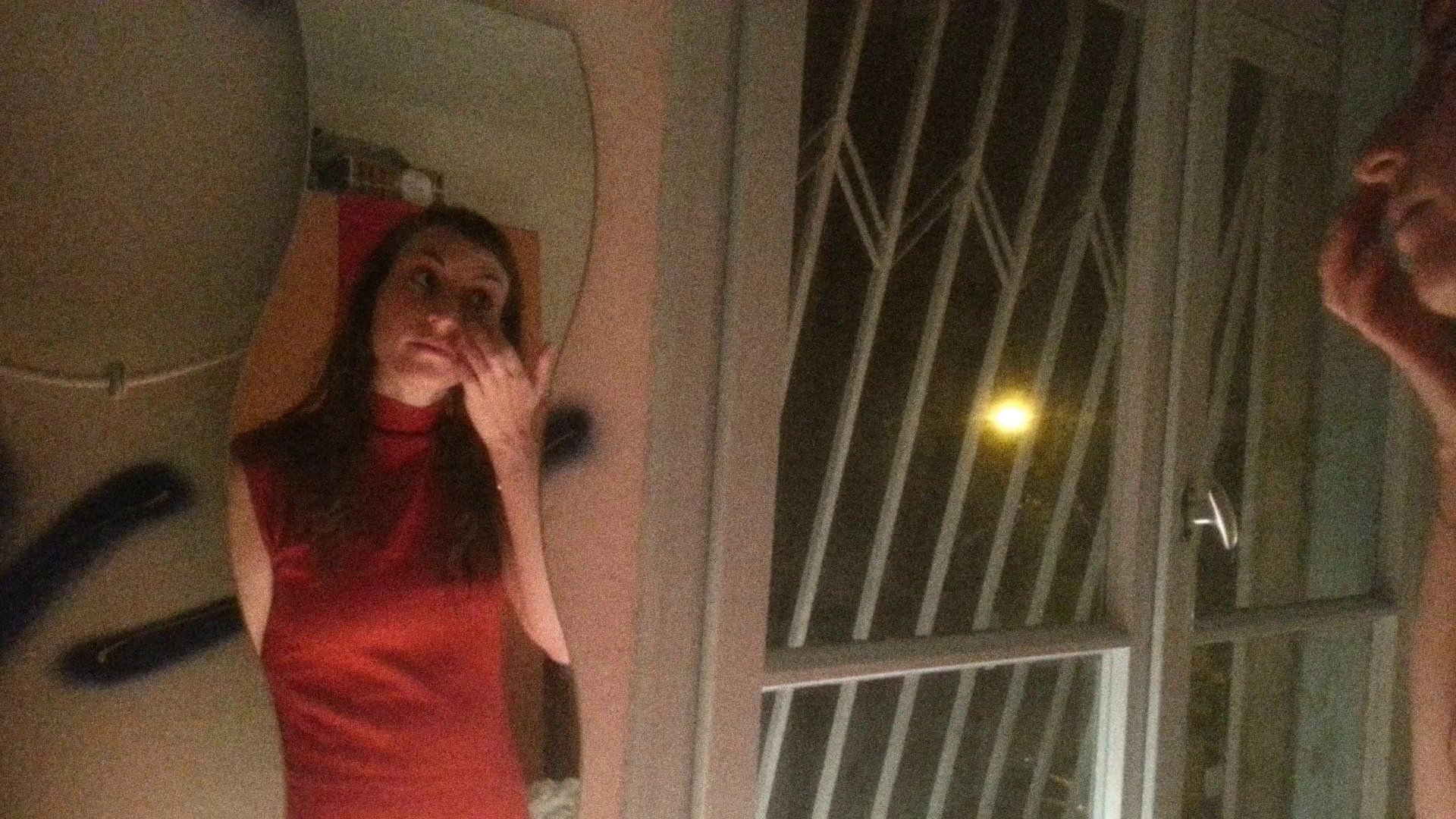}};
                \spy [every spy on node/.append style={line width=0.06cm}, spy connection path={\draw[line width=0.06cm] (tikzspyonnode) -- (tikzspyinnode);}] on (-0.02,0.2) in node at (0.0,-2.2);
            \end{tikzpicture}
        &
            \begin{tikzpicture}[spy using outlines={3787CF, magnification=8, width={\itemwidth - 0.06cm}, height=2.55cm, connect spies,
        every spy in node/.append style={line width=0.06cm}}]
                \node [inner sep=0.0cm] {\includegraphics[width=\itemwidth]{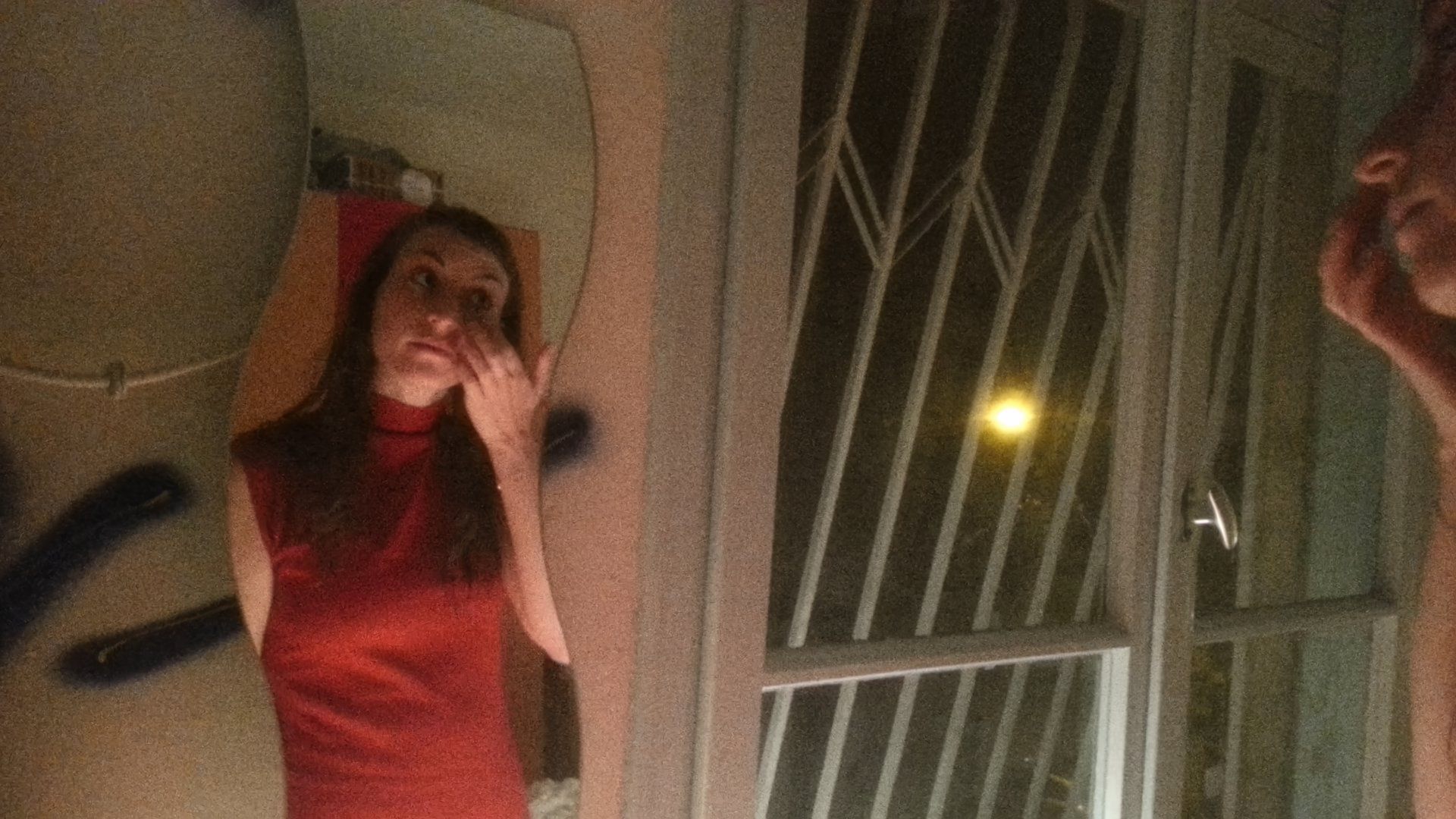}};
                \spy [every spy on node/.append style={line width=0.06cm}, spy connection path={\draw[line width=0.06cm] (tikzspyonnode) -- (tikzspyinnode);}] on (-0.02,0.2) in node at (0.0,-2.2);
            \end{tikzpicture}
        &
            \begin{tikzpicture}[spy using outlines={3787CF, magnification=8, width={\itemwidth - 0.06cm}, height=2.55cm, connect spies,
        every spy in node/.append style={line width=0.06cm}}]
                \node [inner sep=0.0cm] {\includegraphics[width=\itemwidth]{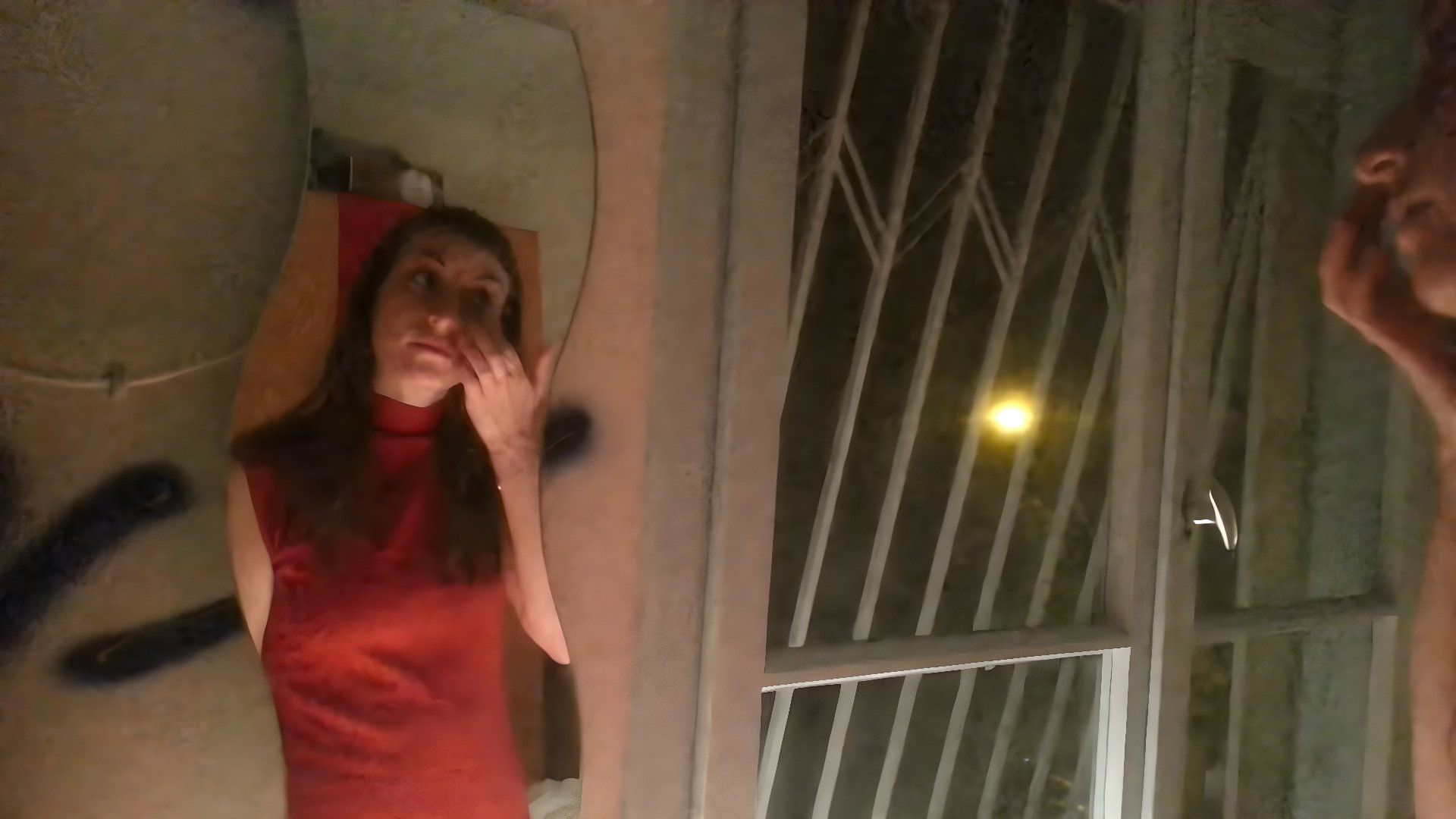}};
                \spy [every spy on node/.append style={line width=0.06cm}, spy connection path={\draw[line width=0.06cm] (tikzspyonnode) -- (tikzspyinnode);}] on (-0.02,0.2) in node at (0.0,-2.2);
            \end{tikzpicture}
        &
            \begin{tikzpicture}[spy using outlines={3787CF, magnification=8, width={\itemwidth - 0.06cm}, height=2.55cm, connect spies,
        every spy in node/.append style={line width=0.06cm}}]
                \node [inner sep=0.0cm] {\includegraphics[width=\itemwidth]{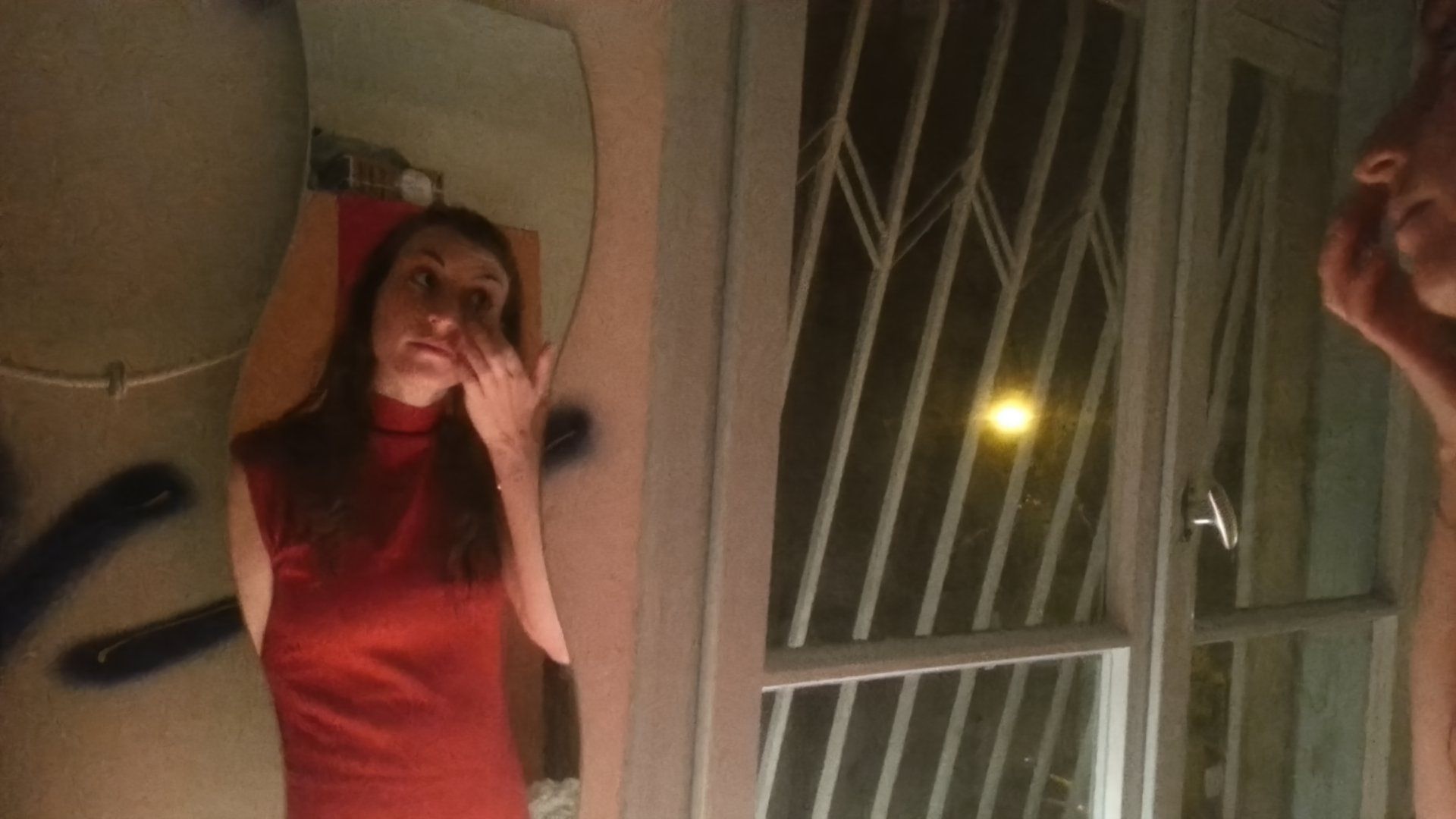}};
                \spy [every spy on node/.append style={line width=0.06cm}, spy connection path={\draw[line width=0.06cm] (tikzspyonnode) -- (tikzspyinnode);}] on (-0.02,0.2) in node at (0.0,-2.2);
            \end{tikzpicture}
        &
            \begin{tikzpicture}[spy using outlines={3787CF, magnification=8, width={\itemwidth - 0.06cm}, height=2.55cm, connect spies,
        every spy in node/.append style={line width=0.06cm}}]
                \node [inner sep=0.0cm] {\includegraphics[width=\itemwidth]{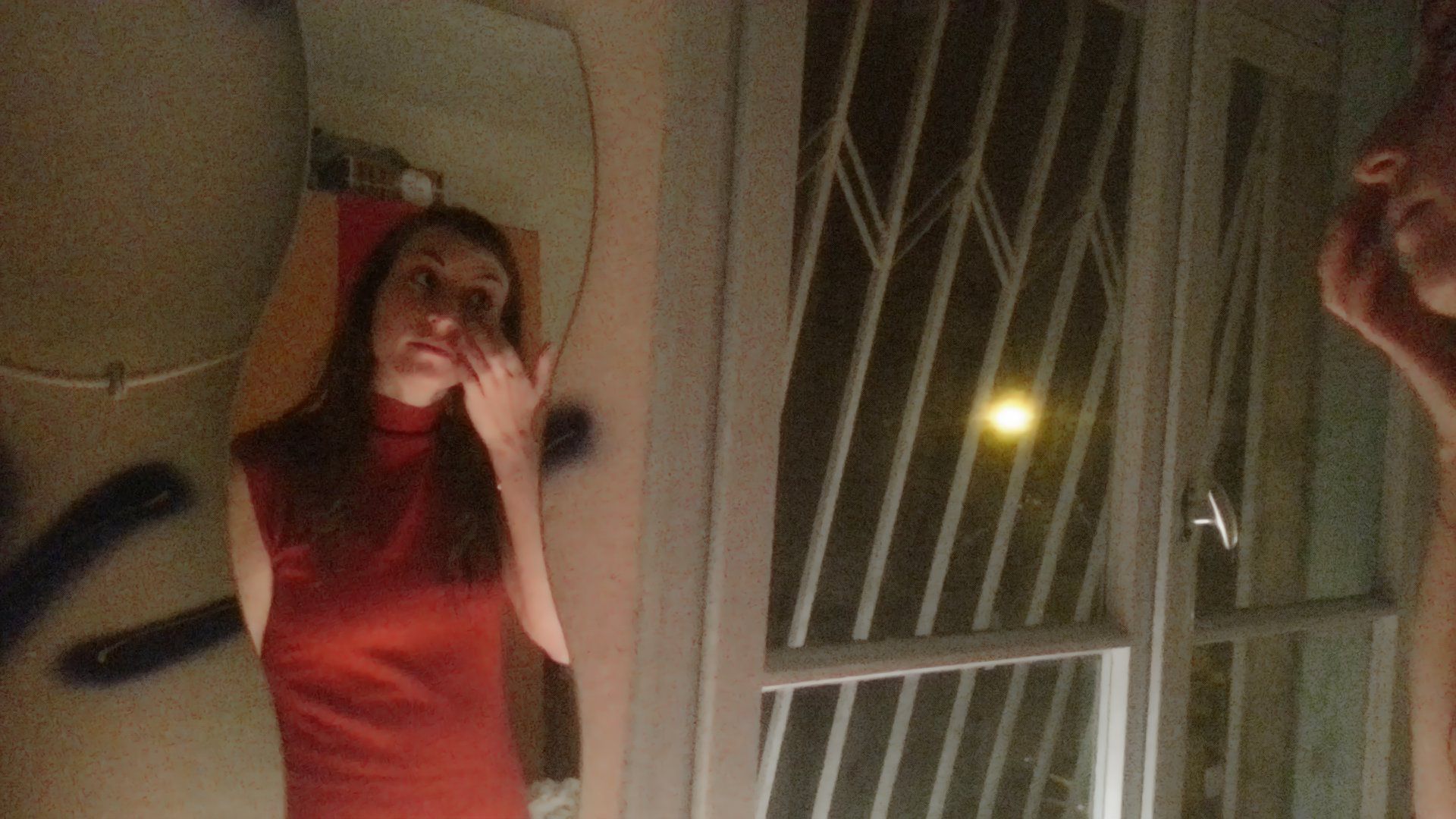}};
                \spy [every spy on node/.append style={line width=0.06cm}, spy connection path={\draw[line width=0.06cm] (tikzspyonnode) -- (tikzspyinnode);}] on (-0.02,0.2) in node at (0.0,-2.2);
            \end{tikzpicture}
        &
            \begin{tikzpicture}[spy using outlines={3787CF, magnification=8, width={\itemwidth - 0.06cm}, height=2.55cm, connect spies,
        every spy in node/.append style={line width=0.06cm}}]
                \node [inner sep=0.0cm] {\includegraphics[width=\itemwidth]{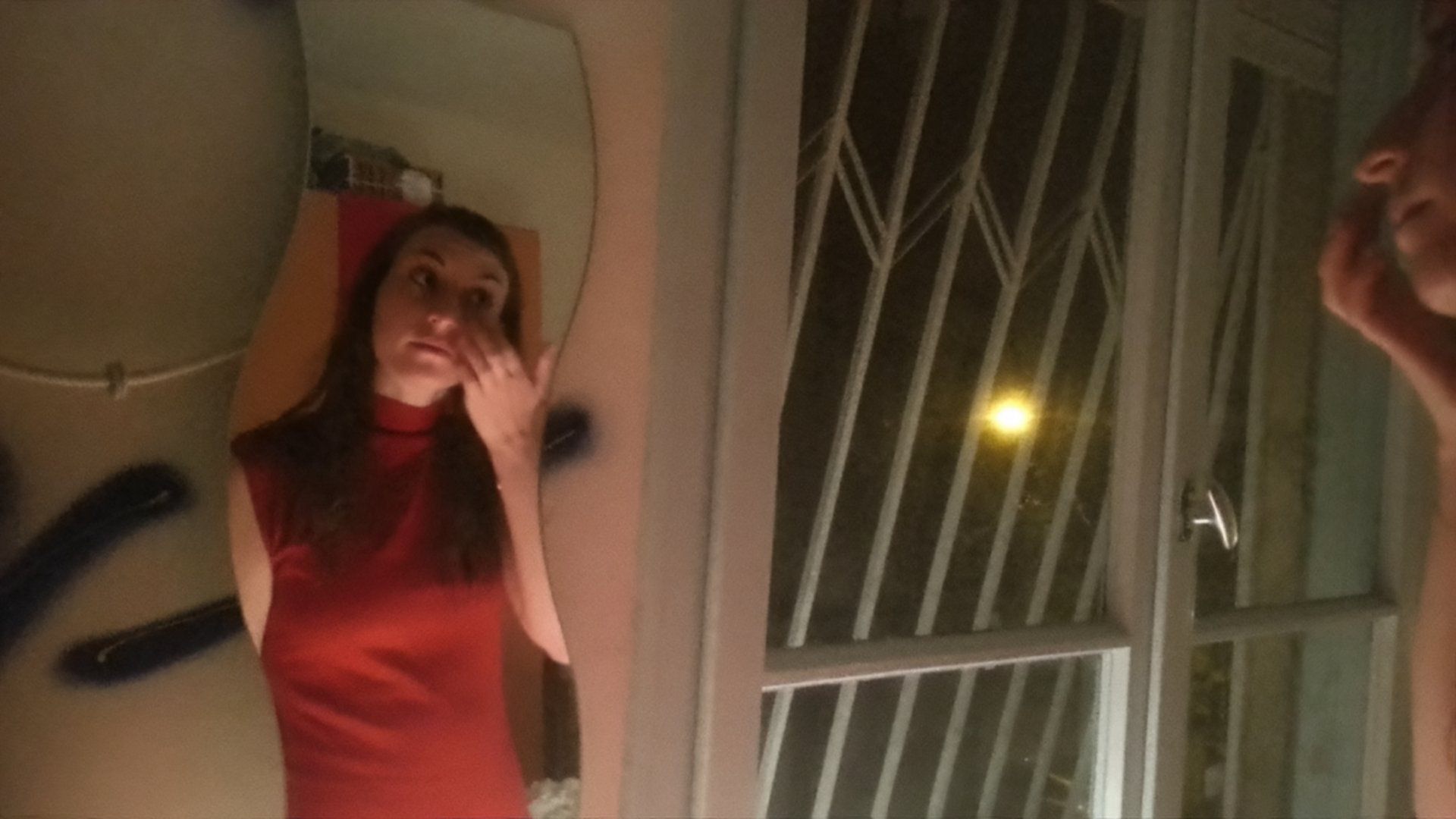}};
                \spy [every spy on node/.append style={line width=0.06cm}, spy connection path={\draw[line width=0.06cm] (tikzspyonnode) -- (tikzspyinnode);}] on (-0.02,0.2) in node at (0.0,-2.2);
            \end{tikzpicture}        
        \\
            \begin{tikzpicture}[spy using outlines={3787CF, magnification=6, width={\itemwidth - 0.06cm}, height=2.55cm, connect spies,
        every spy in node/.append style={line width=0.06cm}}]
                \node [inner sep=0.0cm] {\includegraphics[width=\itemwidth]{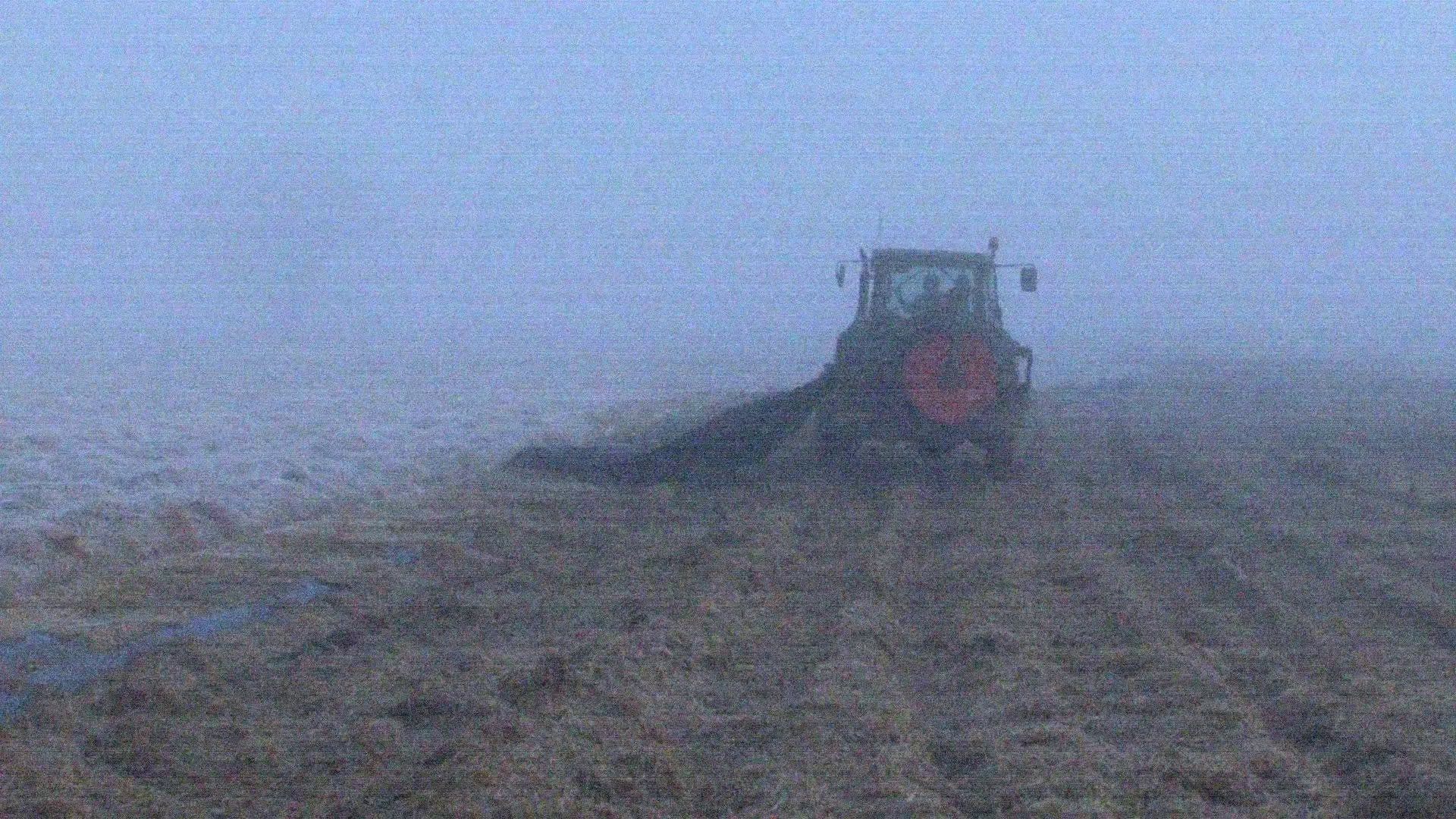}};
                \spy [every spy on node/.append style={line width=0.06cm}, spy connection path={\draw[line width=0.06cm] (tikzspyonnode) -- (tikzspyinnode);}] on (0.3,0.15) in node at (0.0,-2.2);
            \end{tikzpicture}
        &
            \begin{tikzpicture}[spy using outlines={3787CF, magnification=6, width={\itemwidth - 0.06cm}, height=2.55cm, connect spies,
        every spy in node/.append style={line width=0.06cm}}]
                \node [inner sep=0.0cm] {\includegraphics[width=\itemwidth]{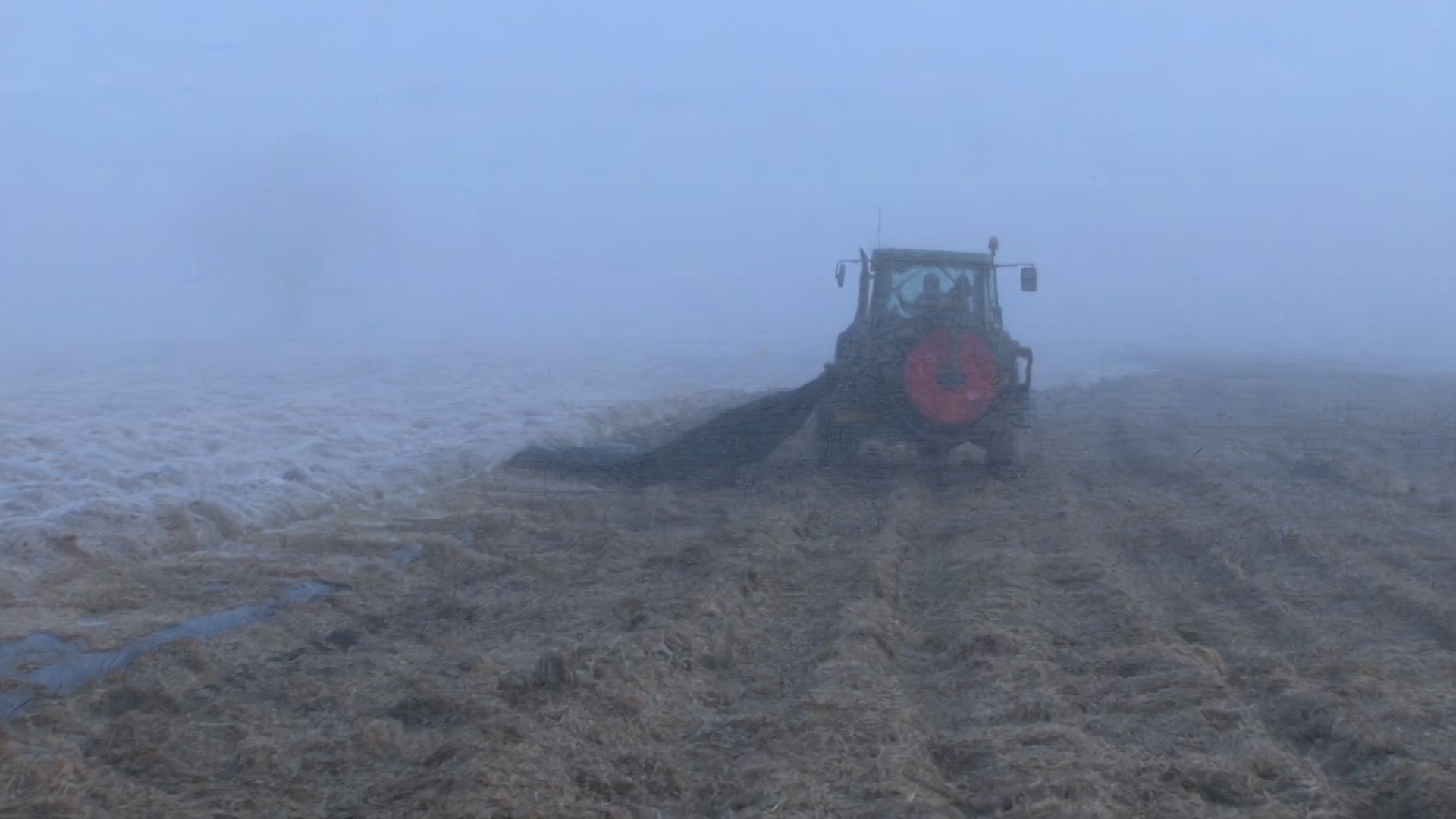}};
                \spy [every spy on node/.append style={line width=0.06cm}, spy connection path={\draw[line width=0.06cm] (tikzspyonnode) -- (tikzspyinnode);}] on (0.3,0.15) in node at (0.0,-2.2);
            \end{tikzpicture}
        &
            \begin{tikzpicture}[spy using outlines={3787CF, magnification=6, width={\itemwidth - 0.06cm}, height=2.55cm, connect spies,
        every spy in node/.append style={line width=0.06cm}}]
                \node [inner sep=0.0cm] {\includegraphics[width=\itemwidth]{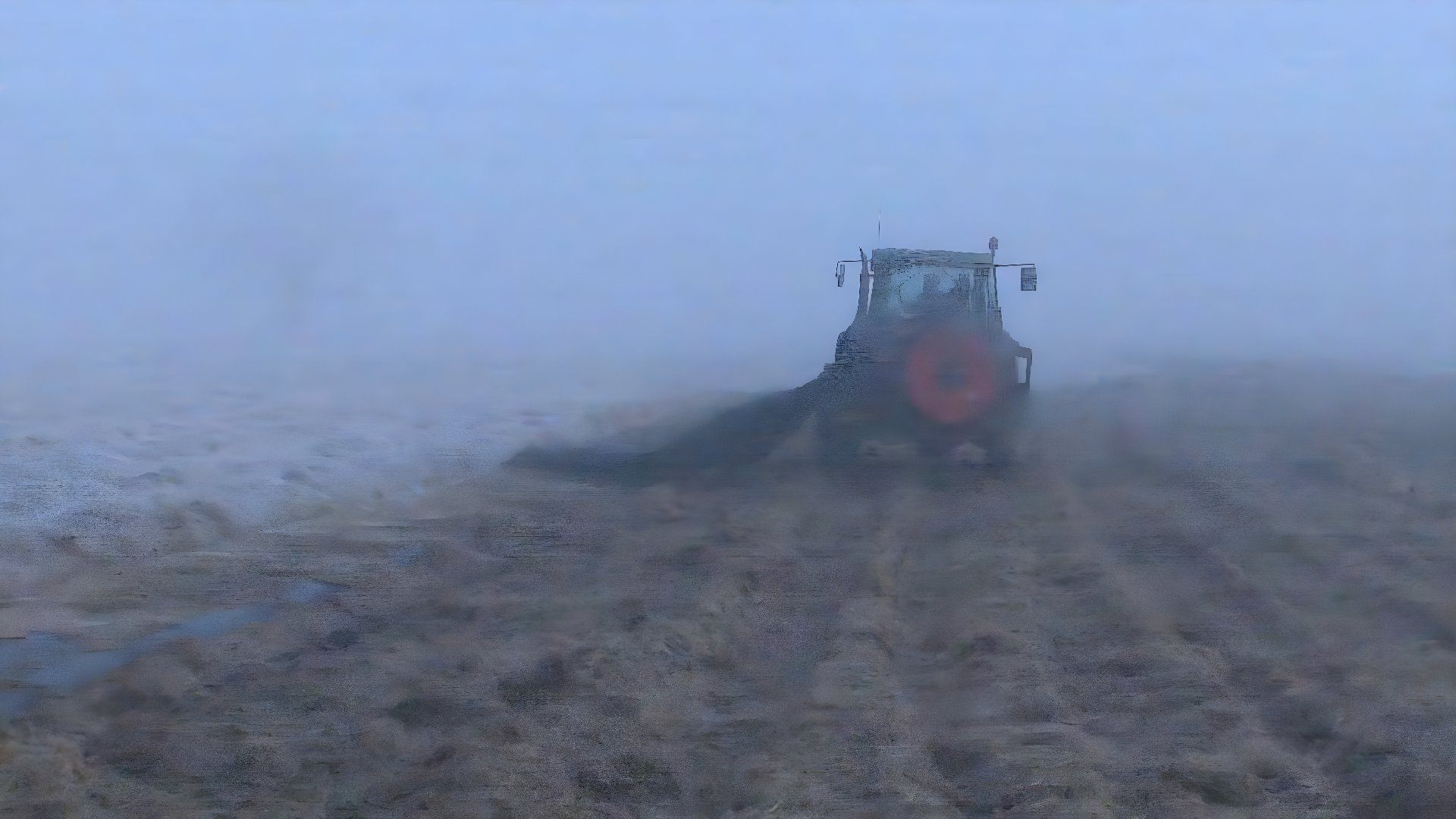}};
                \spy [every spy on node/.append style={line width=0.06cm}, spy connection path={\draw[line width=0.06cm] (tikzspyonnode) -- (tikzspyinnode);}] on (0.3,0.15) in node at (0.0,-2.2);
            \end{tikzpicture}
        &
            \begin{tikzpicture}[spy using outlines={3787CF, magnification=6, width={\itemwidth - 0.06cm}, height=2.55cm, connect spies,
        every spy in node/.append style={line width=0.06cm}}]
                \node [inner sep=0.0cm] {\includegraphics[width=\itemwidth]{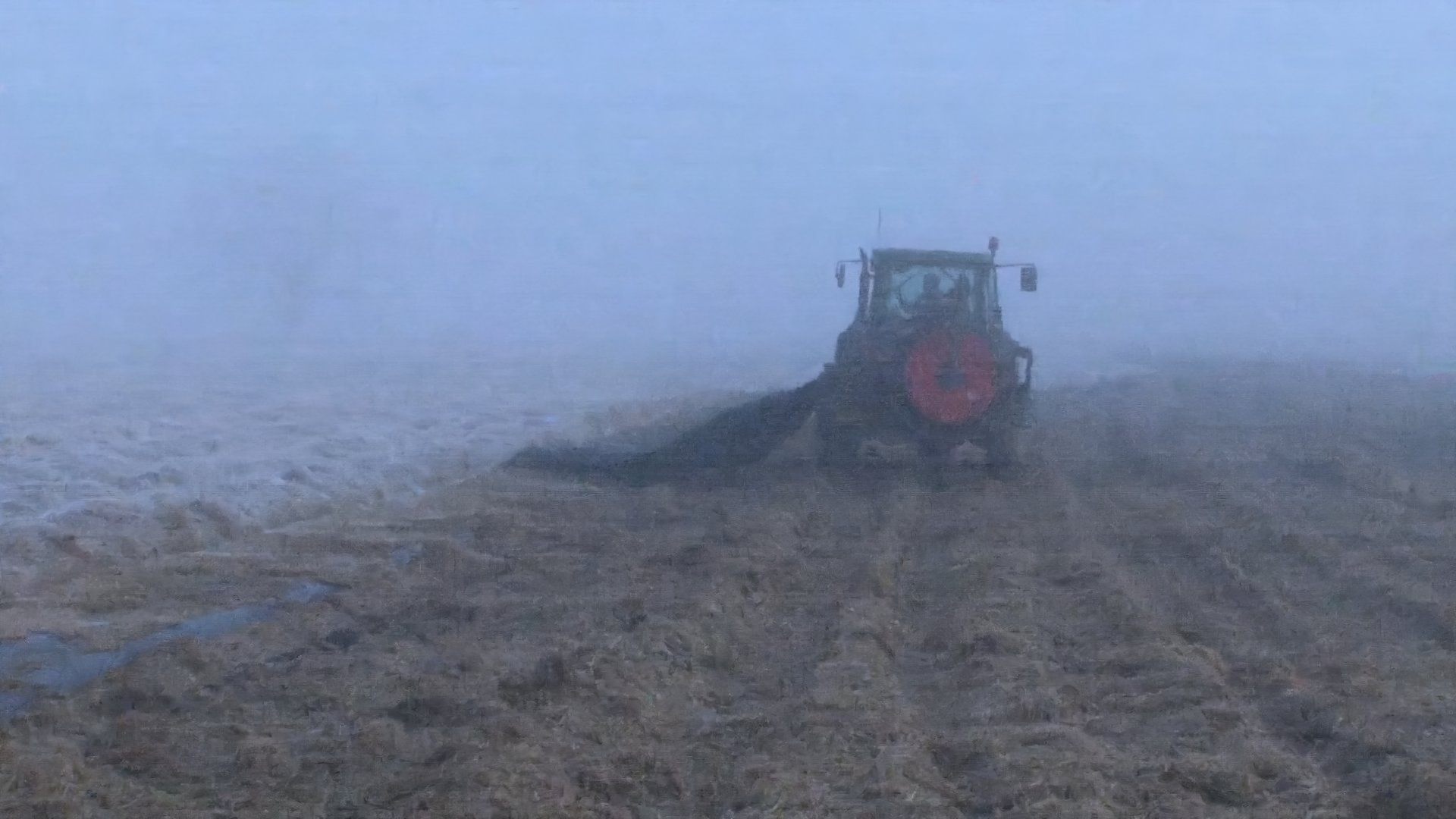}};
                \spy [every spy on node/.append style={line width=0.06cm}, spy connection path={\draw[line width=0.06cm] (tikzspyonnode) -- (tikzspyinnode);}] on (0.3,0.15) in node at (0.0,-2.2);
            \end{tikzpicture}
        &
            \begin{tikzpicture}[spy using outlines={3787CF, magnification=6, width={\itemwidth - 0.06cm}, height=2.55cm, connect spies,
        every spy in node/.append style={line width=0.06cm}}]
                \node [inner sep=0.0cm] {\includegraphics[width=\itemwidth]{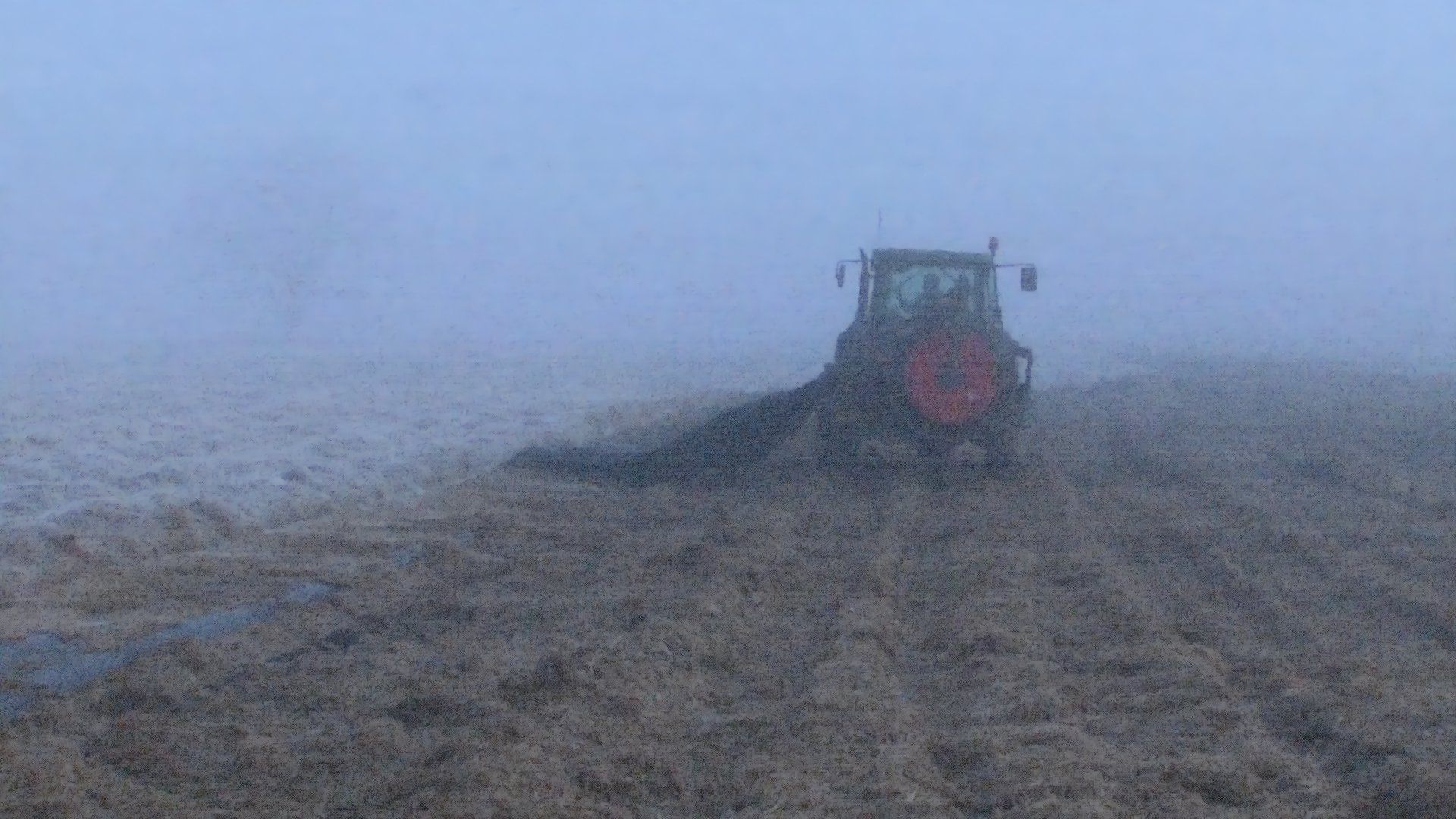}};
                \spy [every spy on node/.append style={line width=0.06cm}, spy connection path={\draw[line width=0.06cm] (tikzspyonnode) -- (tikzspyinnode);}] on (0.3,0.15) in node at (0.0,-2.2);
            \end{tikzpicture}
        &
            \begin{tikzpicture}[spy using outlines={3787CF, magnification=6, width={\itemwidth - 0.06cm}, height=2.55cm, connect spies,
        every spy in node/.append style={line width=0.06cm}}]
                \node [inner sep=0.0cm] {\includegraphics[width=\itemwidth]{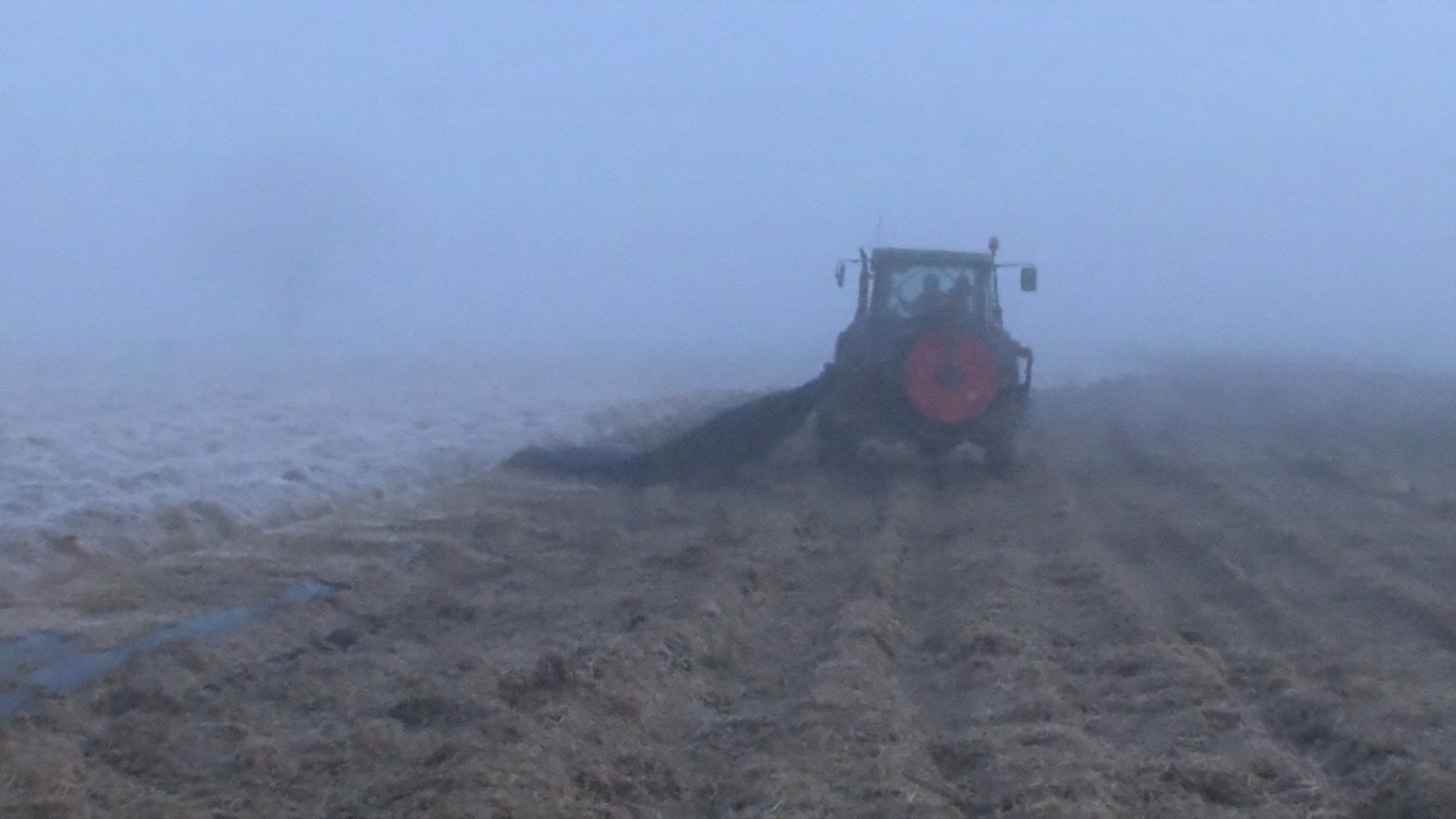}};
                \spy [every spy on node/.append style={line width=0.06cm}, spy connection path={\draw[line width=0.06cm] (tikzspyonnode) -- (tikzspyinnode);}] on (0.3,0.15) in node at (0.0,-2.2);
            \end{tikzpicture}
        \\
            \begin{tikzpicture}[spy using outlines={3787CF, magnification=8, width={\itemwidth - 0.06cm}, height=2.55cm, connect spies,
        every spy in node/.append style={line width=0.06cm}}]
                \node [inner sep=0.0cm] {\includegraphics[width=\itemwidth]{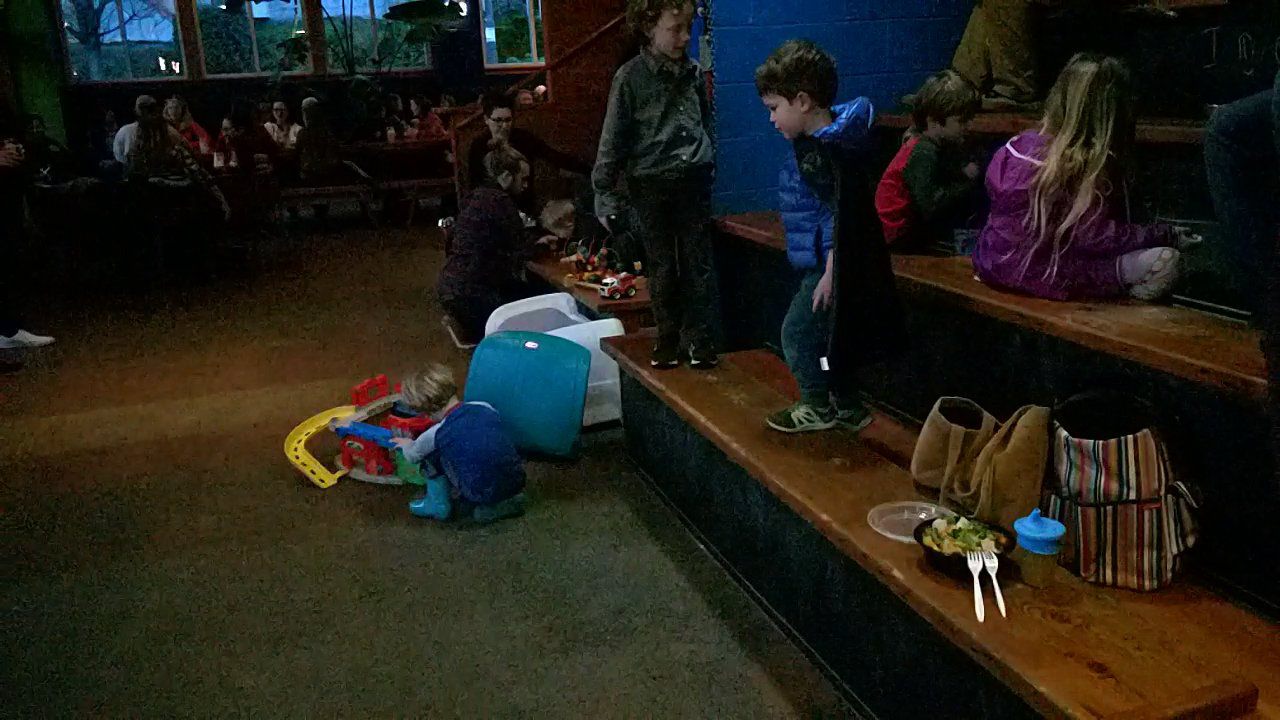}};
                \spy [every spy on node/.append style={line width=0.06cm}, spy connection path={\draw[line width=0.06cm] (tikzspyonnode) -- (tikzspyinnode);}] on (-0.35,-0.4) in node at (0.0,-2.2);
            \end{tikzpicture}
        &
            \begin{tikzpicture}[spy using outlines={3787CF, magnification=8, width={\itemwidth - 0.06cm}, height=2.55cm, connect spies,
        every spy in node/.append style={line width=0.06cm}}]
                \node [inner sep=0.0cm] {\includegraphics[width=\itemwidth]{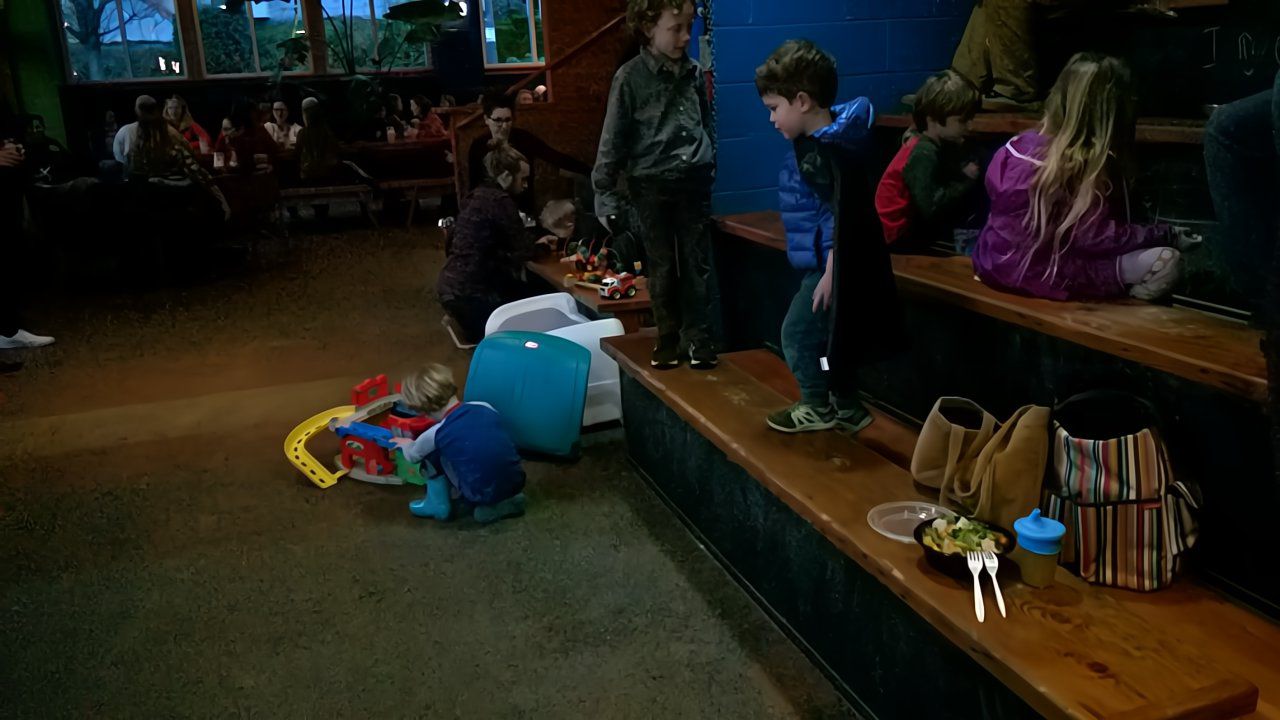}};
                \spy [every spy on node/.append style={line width=0.06cm}, spy connection path={\draw[line width=0.06cm] (tikzspyonnode) -- (tikzspyinnode);}] on (-0.35,-0.4) in node at (0.0,-2.2);
            \end{tikzpicture}
        &
            \begin{tikzpicture}[spy using outlines={3787CF, magnification=8, width={\itemwidth - 0.06cm}, height=2.55cm, connect spies,
        every spy in node/.append style={line width=0.06cm}}]
                \node [inner sep=0.0cm] {\includegraphics[width=\itemwidth]{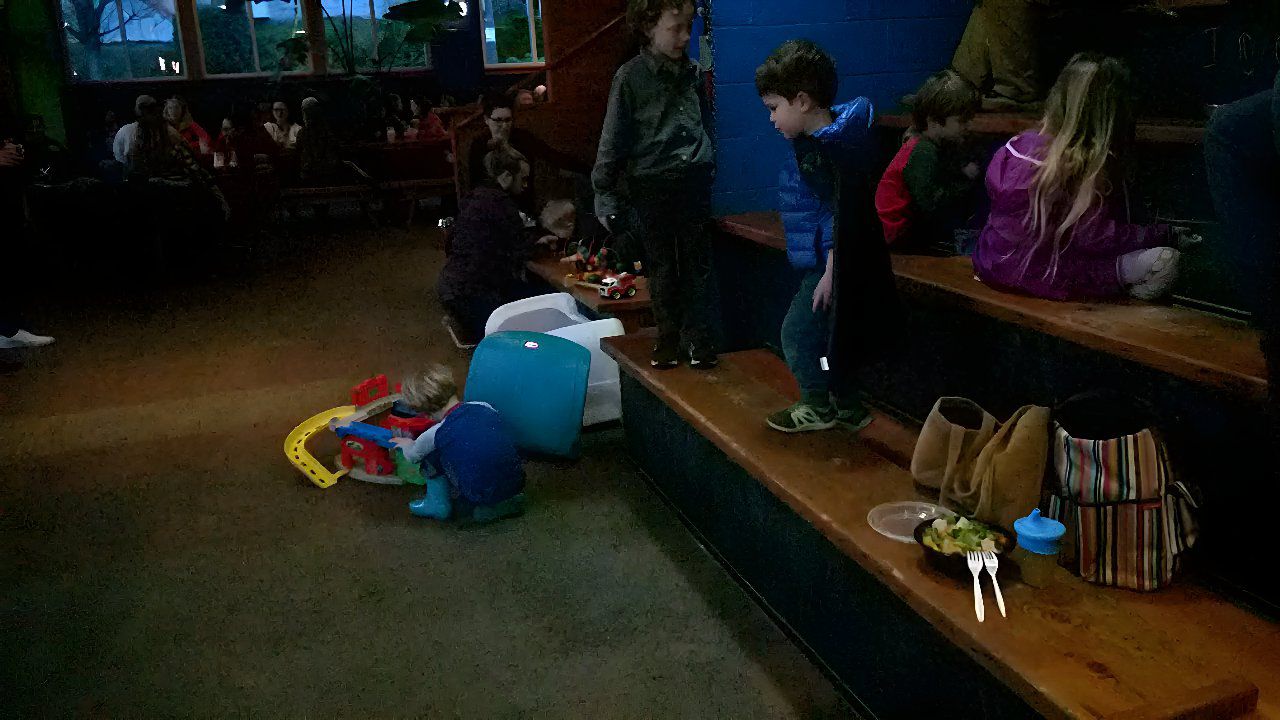}};
                \spy [every spy on node/.append style={line width=0.06cm}, spy connection path={\draw[line width=0.06cm] (tikzspyonnode) -- (tikzspyinnode);}] on (-0.35,-0.4) in node at (0.0,-2.2);
            \end{tikzpicture}
        &
            \begin{tikzpicture}[spy using outlines={3787CF, magnification=8, width={\itemwidth - 0.06cm}, height=2.55cm, connect spies,
        every spy in node/.append style={line width=0.06cm}}]
                \node [inner sep=0.0cm] {\includegraphics[width=\itemwidth]{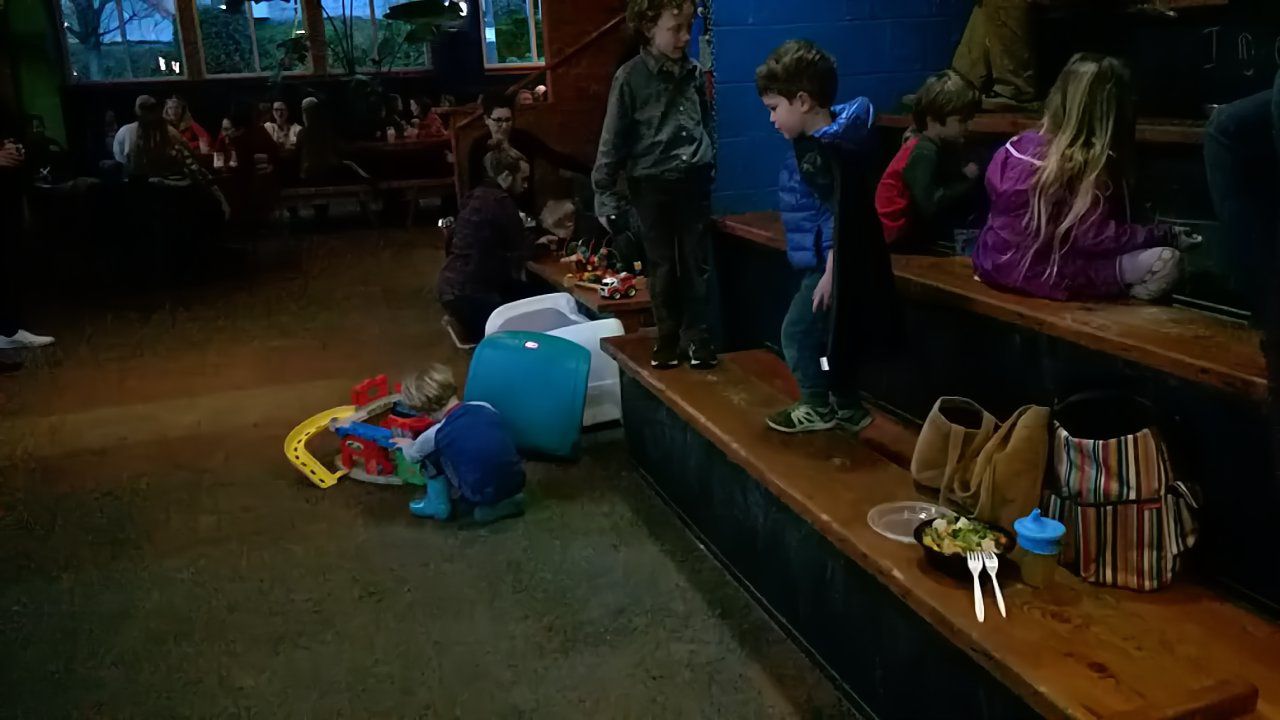}};
                \spy [every spy on node/.append style={line width=0.06cm}, spy connection path={\draw[line width=0.06cm] (tikzspyonnode) -- (tikzspyinnode);}] on (-0.35,-0.4) in node at (0.0,-2.2);
            \end{tikzpicture}
        &
            \begin{tikzpicture}[spy using outlines={3787CF, magnification=8, width={\itemwidth - 0.06cm}, height=2.55cm, connect spies,
        every spy in node/.append style={line width=0.06cm}}]
                \node [inner sep=0.0cm] {\includegraphics[width=\itemwidth]{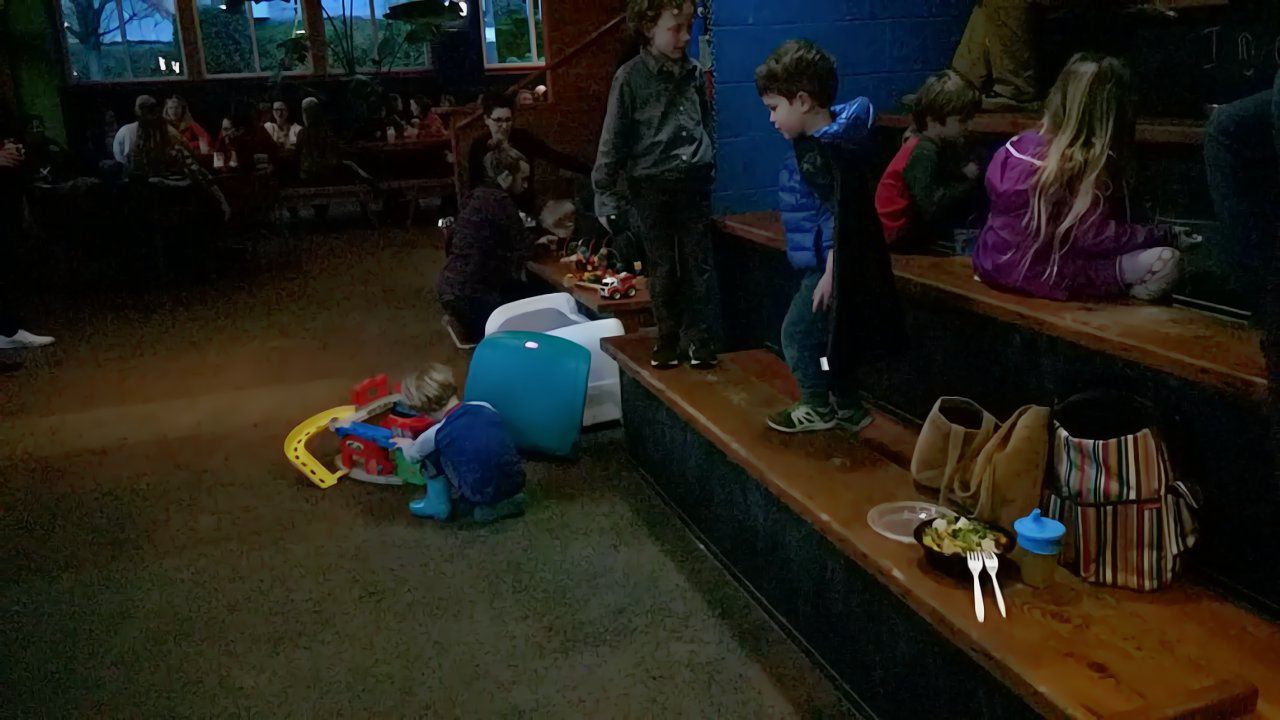}};
                \spy [every spy on node/.append style={line width=0.06cm}, spy connection path={\draw[line width=0.06cm] (tikzspyonnode) -- (tikzspyinnode);}] on (-0.35,-0.4) in node at (0.0,-2.2);
            \end{tikzpicture}
        &
            \begin{tikzpicture}[spy using outlines={3787CF, magnification=8, width={\itemwidth - 0.06cm}, height=2.55cm, connect spies,
        every spy in node/.append style={line width=0.06cm}}]
                \node [inner sep=0.0cm] {\includegraphics[width=\itemwidth]{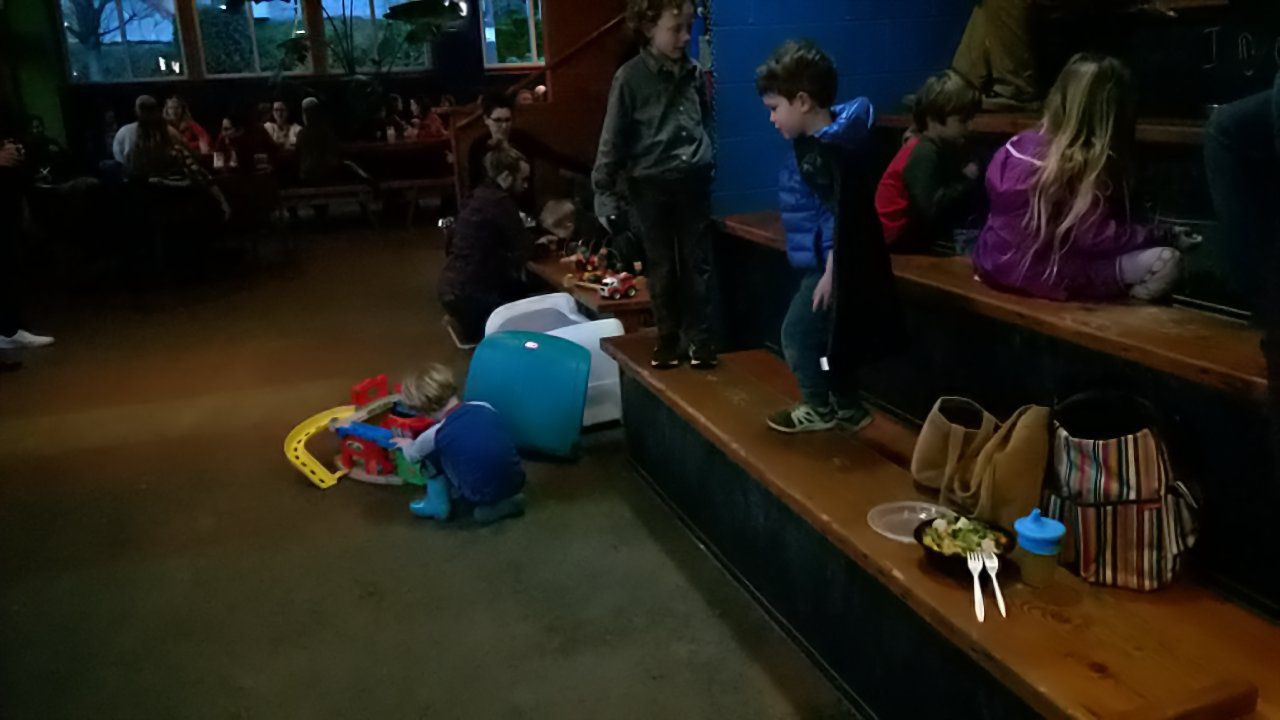}};
                \spy [every spy on node/.append style={line width=0.06cm}, spy connection path={\draw[line width=0.06cm] (tikzspyonnode) -- (tikzspyinnode);}] on (-0.35,-0.4) in node at (0.0,-2.2);
            \end{tikzpicture}
        \\
            \footnotesize Input
        &
            \footnotesize VRT$^\dag$~\cite{liang2024vrt}
        &
            \footnotesize Real-ESRGAN~\cite{wang2021real}
        &
            \footnotesize MF2F~\cite{dewil2021self}
        &
            \footnotesize UDVD~\cite{sheth2021unsupervised}
        &
            \footnotesize Ours - RFCVD
        \\
    \end{tabular}\vspace{-0.2cm}
    \captionof{figure}{Video denoising results from exemplary still frames. Due to space constraints, we only share the results from a representative subset of the methods that we compare to. Please kindly refer to the supplementary to find video results of all methods. We are grateful that Amit Zinman (top row), Robert Kjettrup (middle row), and an anonymous artist (last row) were willing to provide test footage.}\vspace{-0.0cm}
    \label{fig:qualitative}
\end{figure*}

We quantitatively evaluate our approach on two types of videos, real and synthetic. For the former, we leverage the CRVD benchmark~\cite{yue2020supervised} consisting of RAW videos which we convert to sRGB using the provided deep ISP~\cite{ronneberger2015u}. For the latter, we use the validation samples from the REDS dataset~\cite{nah2019reds} which we augmented in two ways that are notably different from the AWGN subject to H.264 transcoding during training. Specifically, one way is using film grain noise~\cite{newson2017realistic} with AV1 transcoding, and the other way is using spatially correlated Gaussian noise with H.265 transcoding. As for evaluation metrics, we follow the typical paradigm of reporting PSNR, SSIM~\cite{wang2004image}, and LPIPS~\cite{zhang2018unreasonable}. Since we argue that our method is computationally efficient, we also report the frames per second (FPS) which we measured on an RTX 3090 GPU with proper CUDA synchronization.

In terms of data for our qualitative evaluation, we reached out to industry professionals who sent us footage where they had trouble removing the noise. Since these professionals did not have a release form for all the actors, we sometimes had to anonymize faces post inference.

Lastly, we compare our approach to various kinds of denoisers. This includes image denoisers~\cite{chen2018learning, chen2022simple}, video denoisers~\cite{tassano2020fastdvdnet, xue2019video}, recurrent methods~\cite{chan2022basicvsr++}, transformers~\cite{liang2024vrt}, models with a sophisticated augmentation pipeline~\cite{wang2021real}, as well as methods that leverage self-supervision~\cite{dewil2021self, sheth2021unsupervised}. Since we have found that our augmentation pipeline works surprisingly well, we have retrained all methods that neither leverage a sophisticated augmentation pipeline already nor utilize self-supervision. This makes for a more fair comparison since, as shown in \tabref{tbl:retrained}, we have improved their performance on the CRVD benchmark with our retrained version. Throughout this paper, we denote all retrained models with a $\dagger$ to prevent any potential confusions.

\subsection{Quantitative Comparison}

Please see \tabref{tbl:crvd} for a summary of the quantitative evaluation on the CRVD benchmark. In short, not only does our approach perform best overall, it is also four times faster than the second-fastest method. We attribute these favorable results to our utilization of classic denoising techniques which greatly helps with the ability to generalize to unseen noise patterns. Specifically, the denoising itself is driven by only a few parameters which naturally reduces overfitting and helps with bridging domain gaps. To test this hypothesis, we conducted an experiment with synthetic data where all models are trained on AWGN with H.264 transcoding but tested on two different degradation pipelines. As shown in \tabref{tbl:generalizability}, our approach performs favorably in this evaluation as well which supports our initial hypothesis.

\subsection{Qualitative Comparison}

To test video denoising in the real world, we reached out to various creative professionals who kindly shared footage where they had trouble removing the noise. We show a representative excerpt of the denoising results on this footage in \figref{fig:qualitative} but kindly refer to the supplementary which includes video results of all methods. Like with the quantitative evaluation, we have found that our approach performs favorably, and once again we hypothesize that this is due to the underlying utilization of classic denoising techniques.

\subsection{Ablative Experiments}
\label{sec:ablations}

One of the main things we wondered ourselves was the performance with respect to the choice of anchor frame. After all, the noise profile of I-frames can be quite different from P- and B-frames. Nevertheless as shown in \tabref{tbl:anchor} (top), we have always used the first frame as the anchor throughout the experiments section and would have drawn the same conclusions with the middle or the last frame. And as shown in \tabref{tbl:anchor} (bottom), we have also found that $\mathcal{L}_{cstsy}$ helps with providing a slightly more consistent experience.

\begin{figure}
    \centering
    \setlength{\tabcolsep}{0.05cm}
    \setlength{\itemwidth}{2.01cm}
    \hspace*{-\tabcolsep}\begin{tabular}{cccc}
            \begin{tikzpicture}[spy using outlines={3787CF, magnification=16, width={\itemwidth - 0.05cm}, height=2.01cm, connect spies,
    every spy in node/.append style={line width=0.05cm}}]
                \node [inner sep=0.0cm] {\includegraphics[width=\itemwidth]{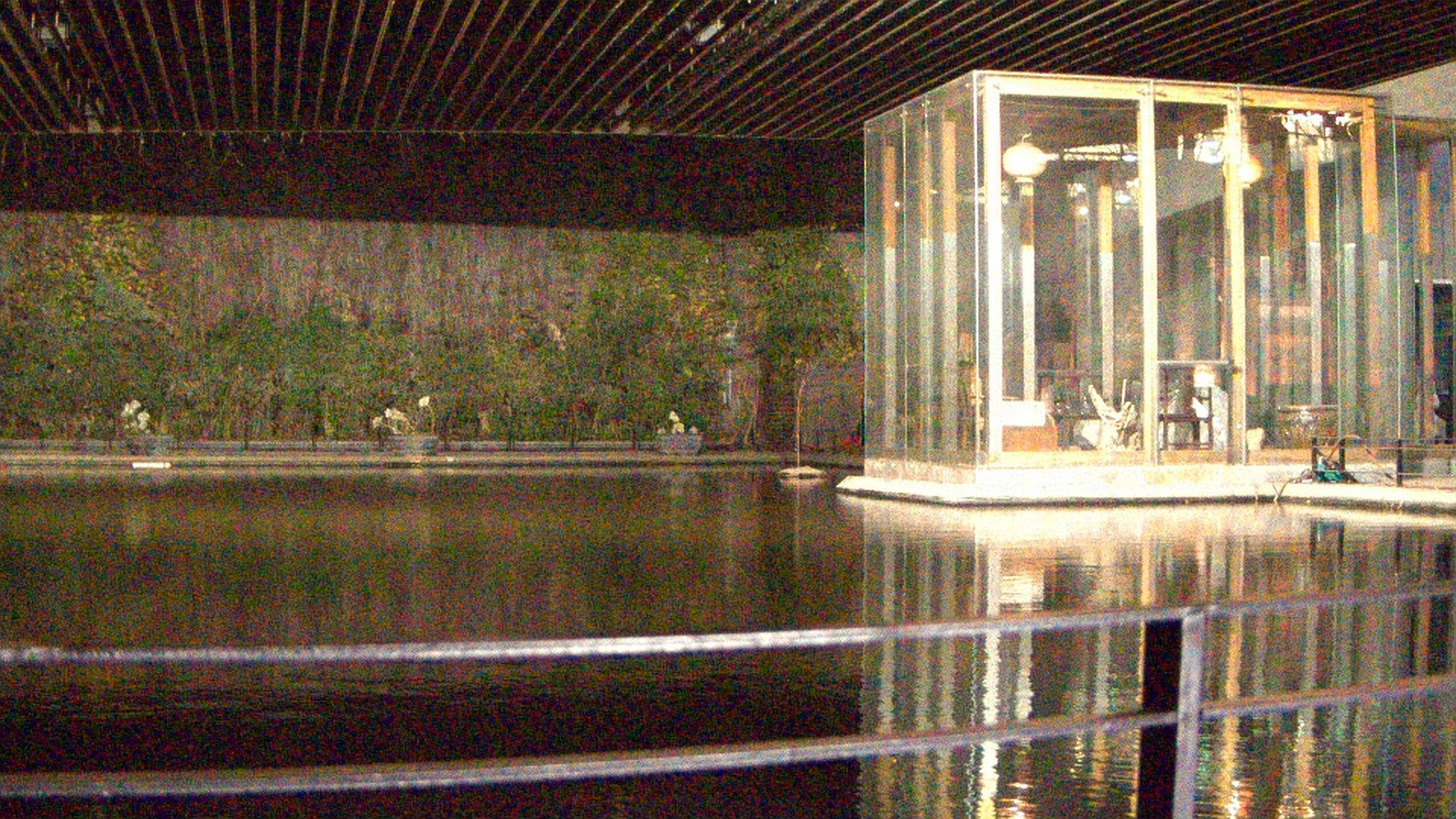}};
                \spy [every spy on node/.append style={line width=0.05cm}, spy connection path={\draw[line width=0.05cm] (tikzspyonnode) -- (tikzspyinnode);}] on (0.5,0.1) in node at (0.0,-1.71);
                \spy [every spy on node/.append style={line width=0.05cm}, spy connection path={\draw[line width=0.05cm] (tikzspyonnode) -- (tikzspyinnode);}] on (-0.2,-0.1) in node at (0.0,1.71);
            \end{tikzpicture}
        &
            \begin{tikzpicture}[spy using outlines={3787CF, magnification=16, width={\itemwidth - 0.05cm}, height=2.01cm, connect spies,
    every spy in node/.append style={line width=0.05cm}}]
                \node [inner sep=0.0cm] {\includegraphics[width=\itemwidth]{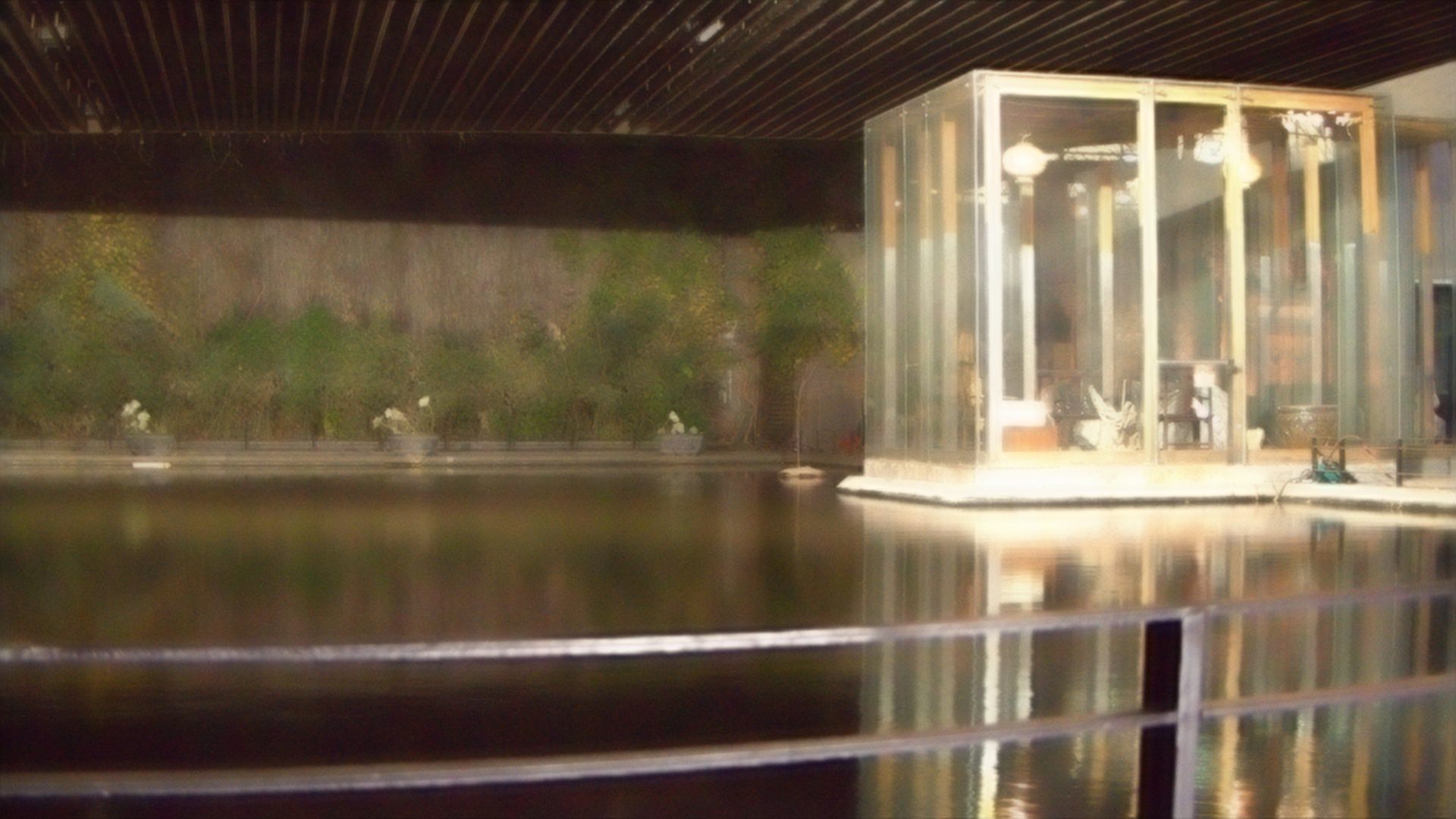}};
                \spy [every spy on node/.append style={line width=0.05cm}, spy connection path={\draw[line width=0.05cm] (tikzspyonnode) -- (tikzspyinnode);}] on (0.5,0.1) in node at (0.0,-1.71);
                \spy [every spy on node/.append style={line width=0.05cm}, spy connection path={\draw[line width=0.05cm] (tikzspyonnode) -- (tikzspyinnode);}] on (-0.2,-0.1) in node at (0.0,1.71);
            \end{tikzpicture}
        &
            \begin{tikzpicture}[spy using outlines={3787CF, magnification=16, width={\itemwidth - 0.05cm}, height=2.01cm, connect spies,
    every spy in node/.append style={line width=0.05cm}}]
                \node [inner sep=0.0cm] {\includegraphics[width=\itemwidth]{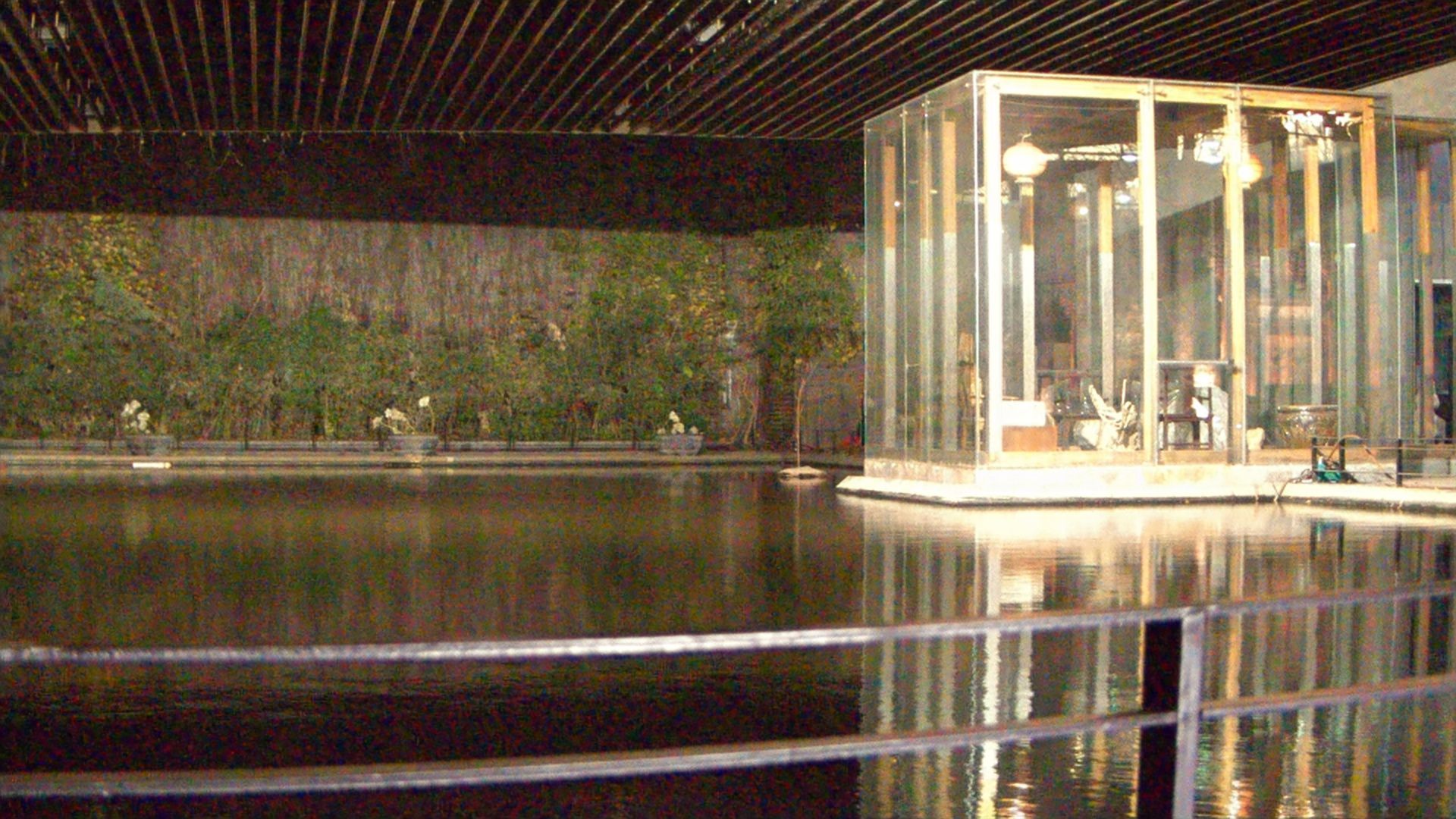}};
                \spy [every spy on node/.append style={line width=0.05cm}, spy connection path={\draw[line width=0.05cm] (tikzspyonnode) -- (tikzspyinnode);}] on (0.5,0.1) in node at (0.0,-1.71);
                \spy [every spy on node/.append style={line width=0.05cm}, spy connection path={\draw[line width=0.05cm] (tikzspyonnode) -- (tikzspyinnode);}] on (-0.2,-0.1) in node at (0.0,1.71);
            \end{tikzpicture}
        &
            \begin{tikzpicture}[spy using outlines={3787CF, magnification=16, width={\itemwidth - 0.05cm}, height=2.01cm, connect spies,
    every spy in node/.append style={line width=0.05cm}}]
                \node [inner sep=0.0cm] {\includegraphics[width=\itemwidth]{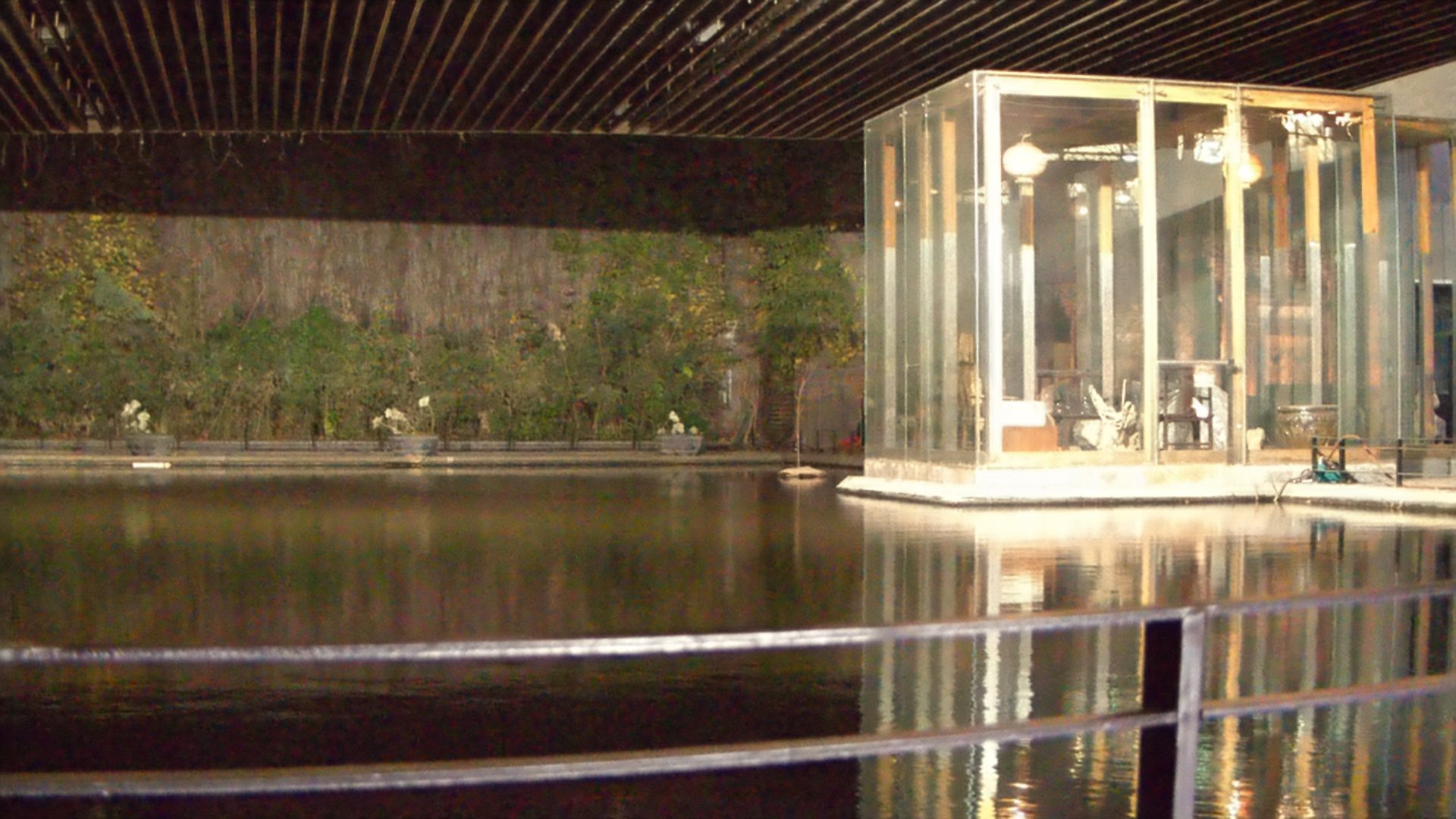}};
                \spy [every spy on node/.append style={line width=0.05cm}, spy connection path={\draw[line width=0.05cm] (tikzspyonnode) -- (tikzspyinnode);}] on (0.5,0.1) in node at (0.0,-1.71);
                \spy [every spy on node/.append style={line width=0.05cm}, spy connection path={\draw[line width=0.05cm] (tikzspyonnode) -- (tikzspyinnode);}] on (-0.2,-0.1) in node at (0.0,1.71);
            \end{tikzpicture}
        \\
            \footnotesize Input
        &
            \footnotesize w/o spat. varying
        &
            \footnotesize w/o backbone $\mathcal{B}$
        &
            \footnotesize Ours
        \\
    \end{tabular}\vspace{-0.2cm}
    \captionof{figure}{Visual comparison of the two most significant ablations on a representative sample from the CRVD dataset.}\vspace{-0.2cm}
    \label{fig:ablations}
\end{figure}

A key part of our integration of classic denoising techniques into a machine learning pipeline is that we account for spatially-varying noise. To assess the importance of this design decision, we trained an ablation that only predicts a set of scalar denoising parameters. As expected and as shown in \tabref{tbl:ablations}, this ``w/o spat. varying'' experiment exhibits a significant drop in quality. As shown in \figref{fig:ablations}, this ablation produces much more blurry denoising results since it cannot easily account for signal-dependent noise.

To best support $\mathcal{P}(\cdot;\theta)$ in predicting the denoising parameters, we not only provide it access to the (merged) input frames but also their gradients as well as a mask indicating whether or not a pixel in an aligned frame is valid. To test the effectiveness of this design, we trained two ablations where we refrained from providing one (``w/o img. gradients'') or the other (``w/o align. mask '') and found that, as shown in \tabref{tbl:ablations}, this context is indeed beneficial.

We argue that analyzing the noise profile of a given image is a mixture of both low-level image processing and high-level semantics, because one first needs an understanding of what an image depicts before being able to go into the low-level details to discern between unintended noise and actual texture. For this reason, we leverage a classification backbone $\mathcal{B}$ in the hypernetwork that is predicting the noise profile $\theta$, and we trained an ablation without this backbone to evaluate this hypothesis. As intuitively expected and as shown in \tabref{tbl:ablations}, this ``w/o backbone $\mathcal{B}$'' experiment indeed exhibits a significant drop in quality. This drop is also visually exemplified in \figref{fig:ablations}, where the lack of this backbone leads to the noise not fully being removed.

Lastly, we not only fine-tune the aforementioned backbone $\mathcal{B}$ but also stabilize the training of the hypernetwork through NPA~\cite{ortiz2023npa}. To analyze the effect of these measures, we trained an ablation for the former (``w/o fine-tuning $\mathcal{B}$'') as well as the latter (``w/o NPA'') and found that, as shown in \tabref{tbl:ablations}, both contribute to the denoising quality.

\begin{figure}\centering
    \setlength{\tabcolsep}{0.0cm}
    \renewcommand{\arraystretch}{1.2}
    \newcommand{\quantTit}[1]{\multicolumn{2}{c}{\scriptsize #1}}
    \newcommand{\quantSec}[1]{\scriptsize #1}
    \newcommand{\quantInd}[1]{\multicolumn{2}{c}{\tiny #1}}
    \newcommand{\quantVal}[1]{\scalebox{0.83}[1.0]{$ #1 $}}
    \footnotesize
    \begin{tabularx}{\columnwidth}{@{\hspace{0.1cm}} X P{1.03cm} @{\hspace{-0.31cm}} P{1.31cm} P{1.03cm} @{\hspace{-0.31cm}} P{1.31cm} P{1.03cm} @{\hspace{-0.31cm}} P{1.31cm}}
        \toprule
            & \quantSec{PSNR} & delta & \quantSec{SSIM} & delta & \quantSec{LPIPS} & delta
        \\[-0.1cm]
            & \quantInd{\hspace{-0.1cm}(higher PSNR is better)} & \quantInd{\hspace{-0.1cm}(higher SSIM is better)} & \quantInd{\hspace{-0.15cm}(lower LPIPS is better)}
        \\ \midrule
            Avg. w/ $\mathcal{L}_{cstsy}$ & \quantVal{35.95} & \quantVal{-} & \quantVal{0.9477} & \quantVal{-} & \quantVal{0.0757} & \quantVal{-} \\
            first frame $\leftarrow$ Ours & \quantVal{36.04} & \quantVal{\text{+ } 0.09} & \quantVal{0.9472} & \quantVal{\text{- } 0.0005} & \quantVal{0.0763} & \quantVal{\text{+ } 0.0006} \\
            middle frame & \quantVal{35.92} & \quantVal{\text{- } 0.03} & \quantVal{0.9480} & \quantVal{\text{+ } 0.0003} & \quantVal{0.0753} & \quantVal{\text{- } 0.0003} \\
            last frame & \quantVal{35.90} & \quantVal{\text{- } 0.06} & \quantVal{0.9480} & \quantVal{\text{+ } 0.0003} & \quantVal{0.0754} & \quantVal{\text{- } 0.0003} \\
            \midrule
            Avg. w/o $\mathcal{L}_{cstsy}$ & \quantVal{35.86} & \quantVal{-} & \quantVal{0.9460} & \quantVal{-} & \quantVal{0.0754} & \quantVal{-} \\
            first frame & \quantVal{35.80} & \quantVal{\text{- } 0.05} & \quantVal{0.9454} & \quantVal{\text{- } 0.0005} & \quantVal{0.0750} & \quantVal{\text{- } 0.0004} \\
            middle frame & \quantVal{36.02} & \quantVal{\text{+ } 0.16} & \quantVal{0.9473} & \quantVal{\text{+ } 0.0014} & \quantVal{0.0753} & \quantVal{\text{- } 0.0001} \\
            last frame & \quantVal{35.75} & \quantVal{\text{- } 0.11} & \quantVal{0.9451} & \quantVal{\text{- } 0.0008} & \quantVal{0.0759} & \quantVal{\text{+ } 0.0005} \\ 
        \bottomrule
    \end{tabularx}\vspace{-0.2cm}
    \captionof{table}{Denoising results on the CRVD benchmark with respect to the anchor frame choice. We find that our approach is relatively robust to this choice (top), partly thanks to $\mathcal{L}_{cstsy}$ (bottom).}\vspace{-0.0cm}
    \label{tbl:anchor}
\end{figure}

\begin{figure}\centering
    \setlength{\tabcolsep}{0.0cm}
    \renewcommand{\arraystretch}{1.2}
    \newcommand{\quantTit}[1]{\multicolumn{2}{c}{\scriptsize #1}}
    \newcommand{\quantSec}[1]{\scriptsize #1}
    \newcommand{\quantInd}[1]{\multicolumn{2}{c}{\tiny #1}}
    \newcommand{\quantVal}[1]{\scalebox{0.83}[1.0]{$ #1 $}}
    \footnotesize
    \begin{tabularx}{\columnwidth}{@{\hspace{0.1cm}} X P{1.03cm} @{\hspace{-0.31cm}} P{1.31cm} P{1.03cm} @{\hspace{-0.31cm}} P{1.31cm} P{1.03cm} @{\hspace{-0.31cm}} P{1.31cm}}
        \toprule
            & \quantSec{PSNR} & delta & \quantSec{SSIM} & delta & \quantSec{LPIPS} & delta
        \\[-0.1cm]
            & \quantInd{\hspace{-0.1cm}(higher PSNR is better)} & \quantInd{\hspace{-0.1cm}(higher SSIM is better)} & \quantInd{(lower LPIPS is better)}
        \\ \midrule
            Ours & \quantVal{36.04} & \quantVal{-} & \quantVal{0.9472} & \quantVal{-} & \quantVal{0.0763} & \quantVal{-} \\
            fine-tuned on GT & \quantVal{36.75} & \quantVal{\text{+ } 0.70} & \quantVal{0.9494} & \quantVal{\text{+ } 0.0022} & \quantVal{0.0591} & \quantVal{\text{- } 0.0172} \\
        \bottomrule
    \end{tabularx}\vspace{-0.2cm}
    \captionof{table}{Fine-tuning our estimated denoising parameters on the ground truth of the CRVD benchmark indicates that our estimates are good but not perfect if we had access to an oracle.}\vspace{-0.2cm}
    \label{tbl:oracle}
\end{figure}

\subsection{Limitations}

While our noise profile $\theta$ with the subsequent denoising parameter prediction from $\mathcal{P}(\cdot;\theta)$ works reasonably well as demonstrated throughout the experiments section, it is far from perfect. Specifically and as shown in \tabref{tbl:oracle}, we fine-tuned the predicted denoising parameters on the ground truth and found that this hypothetical oracle is able to improve our results by a rather significant margin. Nevertheless, we leave bridging this gap to future research.

In the case where the noise profile is entirely known a priori, such as when developing a denoiser for a specific camera model subject to a constrained encoder, any reasonable deep learning method would provide better results than our approach. And with a baked-in noise profile it would perhaps even be faster. However, such a method would in turn be limited in terms of robustness and control.

Lastly, and on a more practical note, deploying hypernetworks in end-user applications is not well supported by the common inference ecosystems yet. Specifically, popular libraries like CoreML, WinML, or OpenVINO expect the model weights to be fixed which is obviously not the case for hypernetworks. As such, while our approach is easy to implement within frameworks like PyTorch, it is not straightforward to deploy in production environments.

\begin{figure}\centering
    \setlength{\tabcolsep}{0.0cm}
    \renewcommand{\arraystretch}{1.2}
    \newcommand{\quantTit}[1]{\multicolumn{2}{c}{\scriptsize #1}}
    \newcommand{\quantSec}[1]{\scriptsize #1}
    \newcommand{\quantInd}[1]{\multicolumn{2}{c}{\tiny #1}}
    \newcommand{\quantVal}[1]{\scalebox{0.83}[1.0]{$ #1 $}}
    \footnotesize
    \begin{tabularx}{\columnwidth}{@{\hspace{0.1cm}} X P{1.03cm} @{\hspace{-0.31cm}} P{1.31cm} P{1.03cm} @{\hspace{-0.31cm}} P{1.31cm} P{1.03cm} @{\hspace{-0.31cm}} P{1.31cm}}
        \toprule
            & \quantSec{PSNR} & delta & \quantSec{SSIM} & delta & \quantSec{LPIPS} & delta
        \\[-0.1cm]
            & \quantInd{\hspace{-0.1cm}(higher PSNR is better)} & \quantInd{\hspace{-0.1cm}(higher SSIM is better)} & \quantInd{\hspace{-0.15cm}(lower LPIPS is better)}
        \\ \midrule
            Ours & \quantVal{36.04} & \quantVal{-} & \quantVal{0.9472} & \quantVal{-} & \quantVal{0.0763} & \quantVal{-} \\
            w/o spat. varying & \quantVal{33.62} & \quantVal{\text{- } 2.42} & \quantVal{0.9359} & \quantVal{\text{- } 0.0112} & \quantVal{0.1184} & \quantVal{\text{+ } 0.0421} \\
            w/o img. gradients & \quantVal{35.61} & \quantVal{\text{- } 0.44} & \quantVal{0.9469} & \quantVal{\text{- } 0.0003} & \quantVal{0.0756} & \quantVal{\text{- } 0.0007} \\
            w/o align. mask & \quantVal{35.26} & \quantVal{\text{- } 0.78} & \quantVal{0.9468} & \quantVal{\text{- } 0.0003} & \quantVal{0.0740} & \quantVal{\text{- } 0.0023} \\
            w/o backbone $\mathcal{B}$ & \quantVal{34.38} & \quantVal{\text{- } 1.67} & \quantVal{0.9059} & \quantVal{\text{- } 0.0412} & \quantVal{0.1444} & \quantVal{\text{+ } 0.0680} \\
            w/o fine-tuning $\mathcal{B}$ & \quantVal{35.71} & \quantVal{\text{- } 0.34} & \quantVal{0.9464} & \quantVal{\text{- } 0.0008} & \quantVal{0.0732} & \quantVal{\text{- } 0.0032} \\
            w/o NPA~\cite{ortiz2023npa} & \quantVal{35.83} & \quantVal{\text{- } 0.21} & \quantVal{0.9453} & \quantVal{\text{- } 0.0018} & \quantVal{0.0814} & \quantVal{\text{+ } 0.0051} \\
         \bottomrule
    \end{tabularx}\vspace{-0.2cm}
    \captionof{table}{Various ablative experiments, please see Section~\ref{sec:ablations} for more context and details of each individual ablation.}\vspace{-0.2cm}
    \label{tbl:ablations}
\end{figure}
\section{Conclusion}
\label{sec:conclusion}

We show that traditional denoising techniques are still very much relevant in the modern machine learning world. Not only have we found them to be robust and fast, but also controllable which is an important property in many video denoising applications where creative professionals want to assert their artistic expression. To bridge the gap between classic denoising and modern approaches, we train a model to estimate the various parameters of a traditional denoising approach for a given input video, which is otherwise not only tedious but also requires a certain amount of skill.

Furthermore, in an effort to make best use of the computational efficiency of classic denoising techniques, our method separates the analysis of the noise from the denoising in order to avoid having to estimate the noise profile over and over again as is typical for machine learning approaches to video denoising. Lastly, we have found a simple yet effective degradation pipeline for training video denoisers based on AWGN subject to H.264 video transcoding. This pipeline ensures that the noise is temporally correlated just like it often is the case in the real world, and we have found it not only to work well for our approach but we have also been able to retrain and improve existing models.

\smallsec{Acknowledgement.}
This work was supported in part by the National Natural Science Foundation of China (62306153, 62225604), the Fundamental Research Funds for the Central Universities (Nankai University, 070-63243143), the Natural Science Foundation of Tianjin, China (24JCJQJC00020), and Shenzhen Science and Technology Program (JCYJ20240813114237048). The computational devices is partly supported by the Supercomputing Center of Nankai University (NKSC).

{
    \small
    \bibliographystyle{ieeenat_fullname}
    \bibliography{main}
}


\end{document}